\documentclass[11pt]{article}

\usepackage[final]{acl}

\usepackage{times}
\usepackage{latexsym}

\usepackage[T1]{fontenc}

\usepackage[utf8]{inputenc}

\usepackage{microtype}

\usepackage{inconsolata}

\usepackage{graphicx}

\usepackage{booktabs}
\usepackage{hyperref}
\usepackage{amsmath}
\usepackage{amssymb}
\usepackage{multirow}
\usepackage{algorithm}
\usepackage{algpseudocode}

\usepackage{multirow}
\usepackage{booktabs}
\usepackage[table]{xcolor}
\usepackage{graphicx}
  
\usepackage[most]{tcolorbox}
\usepackage{fvextra}
\tcbuselibrary{breakable}
\usepackage{listings}

\usepackage{pifont}
\newcommand{\cmark}{\ding{51}}
\newcommand{\xmark}{\ding{55}}

\title{Cross-lingual Biography Enrichment via Claim Extraction and Alignment}

\author{
  \textbf{Yifei Song\textsuperscript{1}},
  \textbf{Ziyang Chen\textsuperscript{2}\thanks{This work was conducted during a six-month internship at CNRS/LORIA.}},
  \textbf{Emil Sayilov\textsuperscript{3}},
  \textbf{Claire Gardent\textsuperscript{1}} \\
  \textsuperscript{1}CNRS/LORIA and Université de Lorraine \quad
  \textsuperscript{2}Université Paris Dauphine - PSL \\
  \textsuperscript{3}ICube Laboratory, Université de Strasbourg \\
  \texttt{\{yifei.song,ziyang.chen,claire.gardent\}@loria.fr} \quad
  \texttt{sayilov@unistra.fr}
}

\begin{document}
\maketitle
\begin{abstract}
English Wikipedia is often treated as the default encyclopedic source, yet non-English Wikipedia editions can contain richer locally grounded information for long-tail figures. We study cross-lingual biography enrichment: enriching an existing English biography with facts supported by a non-English biography about the same person. Focusing on women from non-English-speaking contexts, we introduce \textsc{CLAW-4L}, a benchmark consisting of 300 Wikipedia biography pairs linking an English biography with its French, Chinese or  Azerbaijani counterpart, along with claim annotations and a fine-grained claim-pair relation corpus. We propose a claim-based enrichment framework that extracts English claims from both biographies, aligns them to identify enrichment evidence from the non-English biography, and rewrites the English biography using the selected claims. 
Our results show that non-English Wikipedia biographies provide valuable evidence for improving English biography coverage, while lower-resource settings remain challenging. Code and data are available at \url{https://github.com/MeloS7/cross_lingual_biography_enrichment}.
\end{abstract}

\section{Introduction}
\label{sec:intro}

Wikipedia is a foundational knowledge resource for NLP, supporting language-model pretraining, multilingual modeling, retrieval-augmented generation, and fact verification~\citep{pires-etal-2019-multilingual, guu2020realmretrievalaugmentedlanguagemodel, lewis2021retrievalaugmentedgenerationknowledgeintensivenlp}. Yet Wikipedia coverage is uneven across languages: the same entity may be described with substantially different levels of detail across language editions~\citep{roy-etal-2020-topic}. While English Wikipedia is often treated as the default encyclopedic source,  it is not always the richest source for long-tail entities with stronger local relevance in non-English communities.

\begin{figure}[t]
    \centering
    \includegraphics[width=1.0\linewidth]{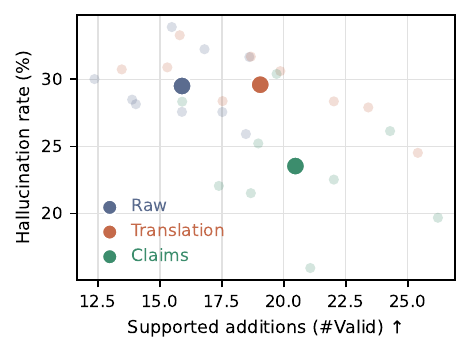}
    \caption{Generation trade-off across three representations of non-English evidence. \textbf{Raw}: the generator receives the original English biography and the raw non-English biography. \textbf{Translation}: the non-English biography is first translated into English, and the generator receives the original English biography and the translation. \textbf{Claims}: the generator receives the original English biography and English enrichment claims selected through cross-lingual claim extraction and alignment. Faint points show individual language--generator settings; large points average over three non-English languages and three open-source generators. Translation and Claims increase supported additions over Raw, while the Claims setting substantially reduces hallucination.}
    \label{fig:halluc_supported_add_plot}
\end{figure}

\begin{figure*}[t]
    \centering
    \includegraphics[width=0.95\linewidth]{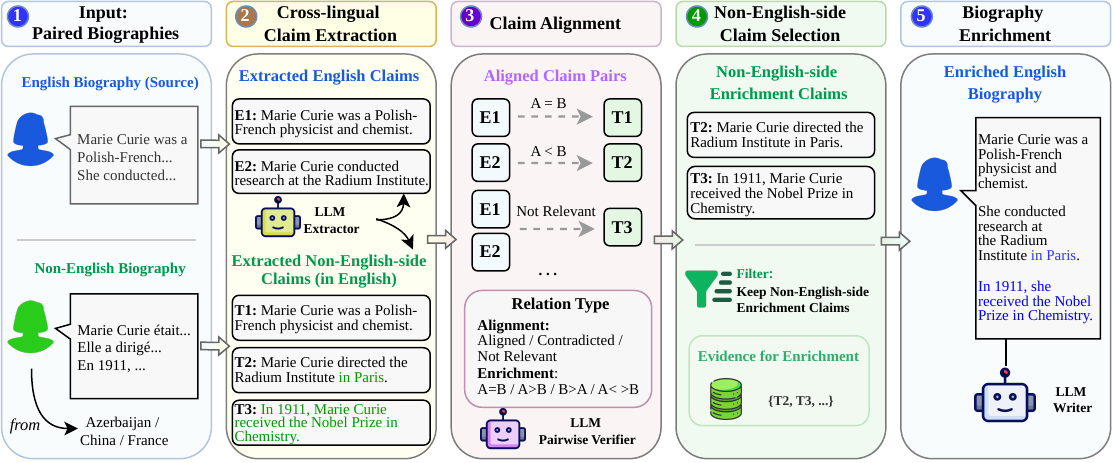}
    \caption{
    \textbf{Overview of our claim-based framework for cross-lingual biography enrichment.}
    The framework first extracts claims from paired English and non-English biographies into a shared English space, then performs claim-pair alignment to identify non-English-side enrichment candidates not already covered by the English side.
    The selected enrichment claims are used as structured evidence for controlled English biography enrichment.
    }
    \label{fig:main_pipeline}
\end{figure*}

This asymmetry motivates our focus on English-language biography enrichment for women from non-English-speaking contexts, where figures may be well documented in local Wikipedia editions while remaining underspecified in English~\cite{Wagner_2016}. Prior work either generates short biographies from structured data~\citep{Vougiouklis_2018, gardent-etal-2017-webnlg} or full biographies with retrieval-augmented web evidence~\citep{fan-gardent-2022-generating}; neither directly studies enriching existing English biographies with curated non-English Wikipedia evidence.

We study \textit{cross-lingual biography enrichment}: given an English biography and a non-English biography of the same woman, the goal is to enrich the English version by incorporating additional information that appears in the non-English biography but is absent from the English one. Unlike generating a biography from scratch, this setting starts from an existing English article and uses a curated non-English encyclopedic article as additional evidence to enrich it. 


We investigate three text generation settings -  they all use the same rewriting generators but differ in how the non-English biography is represented. In \textit{raw cross-lingual generation}, the generator receives the original English biography together with the non-English biography, and must directly interpret, align, and integrate the non-English content into the generated English biography. In \textit{translation-based generation}, the non-English biography is first translated into English, after which the generator rewrites the original English biography using both the original English biography and the translated text as evidence. In our \textit{claim-based setting}, both biographies are mapped into a shared English claim space; claim alignment filters out information already covered by the English biography, and the generator receives the original English biography together with the remaining enrichment claims. This decomposition makes evidence selection explicit before generation, reduces the burden of both long-context and implicit cross-lingual alignment, and enables intermediate evaluation.

Figure~\ref{fig:halluc_supported_add_plot} previews the resulting trade-off. Both translation-based and claim-based generation yield more supported additions than raw cross-lingual generation, whereas the claim-based setting achieves a substantially lower hallucination rate. This suggests that English normalization facilitates information transfer, while explicit selection of novel evidence is important for reliable enrichment.
Our contributions are as follows:
\begin{itemize}
    \item We study cross-lingual biography enrichment, a task that uses non-English Wikipedia biographies as evidence for enriching existing English biographies.
    \item We provide benchmarks for evaluating cross-lingual biography enrichment from either French, Chinese or Azerbaijani into English (\textsc{CLAW-4L}), cross-lingual claim extraction (\textsc{CLAW-4L-CX}) and English claim-pair relation classification  (\textsc{CLAW-4L-RC}). 
    \item We develop and evaluate a claim-based enrichment framework that explicitly selects non-English-supported information missing from the English biography before rewriting, improving the supported-addition versus hallucination trade-off over raw cross-lingual and translation-based generation.
\end{itemize}

\section{Related Work}
\label{sec:related_work}

\paragraph{Wikipedia Biography Generation and Enrichment.}
Early biography-generation work framed the task as data-to-text generation from infoboxes or Wikidata, usually producing a first sentence or short summary~\citep{lebret-etal-2016-neural,chisholm-etal-2017-learning,Vougiouklis_2018,kaffee-etal-2018-learning}. Recent systems also generate editable Wikipedia drafts for underrepresented groups from structured data~\citep{mille-etal-2024-filling-gaps}. These approaches improve coverage, but remain structured-data-driven and short-form compared with enriching existing full biographies.

Retrieval-grounded systems instead synthesize Wikipedia articles from similar pages and web evidence~\citep{10.5555/3060832.3061004}, with recent benchmarks emphasizing full-length structure, grounding, and citations~\citep{zhang-etal-2025-wikigenbench}. In the biography domain, \citet{fan-gardent-2022-generating} generate full-length English biographies using retrieved web evidence and show that women biographies are challenging due to limited evidence availability. Closest to our setting, \citet{adak-etal-2025-reversum} enrich tail biographies with personal narratives. Our work is complementary: we use human-written non-English Wikipedia biographies as cross-lingual encyclopedic evidence and select non-English-side facts before generation.


\paragraph{Claim Extraction, Decomposition, and Alignment.} Claim extraction has been studied through open information extraction and fact extraction for verification~\citep{stanovsky-etal-2018-supervised, thorne2018feverlargescaledatasetfact}, but such methods often rely on task-specific schemas or language-specific NLP tools. Recent LLM-based factuality work decomposes long-form text into atomic or verifiable claims, including FactScore~\citep{min-etal-2023-factscore}, FactCheck-GPT~\citep{wang-etal-2024-factcheck}, VeriScore~\citep{song-etal-2024-veriscore}, DnDScore~\citep{wanner-etal-2025-dndscore} and Claimify~\citep{metropolitansky-larson-2025-towards}. These methods differ in how they decontextualize, decompose, and verify claims, but they are primarily developed for English inputs rather than cross-lingual claim extraction into a shared English space. In this work, we use these methods as candidate extraction frameworks and evaluate their cross-lingual adaptations empirically.

Cross-lingual fact-checking studies multilingual claim detection, retrieval, and verification~\citep{gupta-srikumar-2021-x,chang-etal-2023-xfever}. Our setting differs in requiring English claims to be extracted from non-English biographies, and aligned to identify enrichment evidence. For alignment, NLI and fact-verification models provide useful baselines~\citep{bowman-etal-2015-large,tang-etal-2024-minicheck}, but their binary or three-way label spaces cannot distinguish whether a non-English-side claim is exactly covered, more specific, less specific, complementary, contradictory, or unrelated. We use this finer-grained relation space to support enrichment-oriented claim selection in \textsc{CLAW-4L-RC}.

\section{Task and Method}
\label{sec:task}

Given $(B^{en}, B^{X})$, two Wikipedia biographies about the same person where $B^{en}$ is in English and $B^{X}$ is in either French, Chinese or Azerbaijani,  the \textit{cross-lingual biography enrichment} task consists in generating an English biography $\tilde{B}^{en}$ that preserves the content of $B^{en}$ while incorporating additional information supported by $B^{X}$. 

We frame this task as an LLM rewriting task where the input is the original English biography $B^{en}$ augmented with $\mathcal{C}^{add}$, the set of English claims that can be extracted from $B^{X}$ but not from  $B^{en}$:
\[
\tilde{B}^{en} = G(B^{en}, \mathcal{C}^{add}),
\]

We define a claim as a predicate together with its arguments and modifiers and, when present, the corresponding hedge. However, since biographies are highly factual in nature,  claims in our settings largely correspond to facts and hedged contexts are rare. 


\section{Datasets}
\label{sec:claw-4l}

We create three benchmarks for evaluation.

\textbf{CLAW-4L} (\textbf{C}ross-\textbf{L}ingual \textbf{A}lignment on \textbf{W}omen biographies in \textbf{4} \textbf{L}anguages) comprises 300 cross-lingual pairs of  women biographies $(B^{en}, B^X)$, automatically labelled with claims $(C^{en}, C^X)$. Here, $B^{en}$ is in English while  $B^X$ is in one of our three non-English languages:  French, Chinese or Azerbaijani. We use this benchmark to evaluate biography enrichment i.e., the difference between the informational content of a generated biography and the informational content of both the original English and the non English biography (Section~\ref{sec:generation}). 

\textbf{CLAW-4L-CX} \textit{(Claim Extraction)} consists of 600 sentences manually annotated with English claims, of which 300 are English sentences and 100 for each non-English language. We use this benchmark to evaluate different claim extraction models (cf. Section~\ref{subsec:claim_extract_method}). 


Finally, \textbf{CLAW-4L-RC} \textit{(Relation Classification)} contains 600 claim pairs annotated with a label indicating the semantic relation between the two claims: is the relation between the two claims one of exact alignment, partial alignment, contradiction or irrelevance? We use this benchmark to evaluate the relation classifiers (Cf. Section~\ref{sec:claim_classifier}).

\subsection{CLAW-4L}

\textsc{CLAW-4L} consists of 300 pairs of English and non-English Wikipedia biographies automatically labelled with claims.

We construct the biography pairs in two stages. First, we build oversized country-specific candidate pools from Wikidata by requiring that each entity be a woman, have the corresponding country of citizenship, and include both English and non-English-language Wikipedia sitelinks. We then retrieve and clean the paired biographies, removing markup and non-biographical trailing sections such as references and external links.

\begin{table*}[!htbp]
\centering
\small
\resizebox{\textwidth}{!}{
\begin{tabular}{lrrrrrrrrr}
\toprule
\textbf{Pair} & \textbf{N} & \textbf{Tok (EN/X)} & $\mathbf{r_t^{norm}}$ & \textbf{Sent (EN/X)} & $\mathbf{r_s}$ & \textbf{Claim (EN/X)} & $\mathbf{r_c}$ & \textbf{Max T$_{X}$} & \textbf{Max C$_{X}$} \\
\midrule
EN$\rightarrow$FR & 100 & 465 / 1867 & 4.19 & 16.1 / 48.4 & 4.64 & 29.8 / 101.8 & 3.42 & 12852 & 705 \\
EN$\rightarrow$ZH & 100 & 489 / 1041 & 2.83 & 15.8 / 27.2 & 2.01 & 29.3 / 72.4 & 2.47 & 8772  & 662 \\
EN$\rightarrow$AZ & 100 & 544 / 1967 & 3.20 & 19.4 / 55.2 & 3.48 & 33.6 / 93.8 & 2.79 & 28420 & 1363 \\
\midrule
\textbf{EN$\rightarrow$X}  & \textbf{300} & \textbf{500 / 1625} & \textbf{3.41} & \textbf{17.1 / 43.6} & \textbf{3.38} & \textbf{30.9 / 89.3} & \textbf{2.89} & \textbf{28420} & \textbf{1363} \\
\bottomrule
\end{tabular}
}
\caption{
\textbf{Summary statistics of \textsc{CLAW-4L}}. 
We report average token (Tok), sentence (Sent), and claim (Claim) counts for English (EN) and non-English (X) biographies, along with normalized token ratios ($r_t^{norm}$), sentence ratios ($r_s$), and claim ratios ($r_c$). 
Non-English biographies are substantially richer across all levels (e.g., 3.4$\times$ tokens and 2.9$\times$ claims on average). 
The maximum non-English biography length reaches 28,420 tokens, and the number of extracted claims reaches 1,363 per instance, highlighting the context length challenges of claim-level alignment and biography generation.
}
\label{tab:claw4-main-stats}
\end{table*}

To favor biography pairs where the non-English biography is richer (more informative) than the corresponding English biography, we rank candidates by a language-calibrated non-English-to-English token ratio computed on cleaned biographies. The ratio normalizes raw token counts using language-specific inflation factors estimated from parallel multilingual data, reducing tokenizer-induced cross-lingual bias. We use it only as a heuristic proxy for non-English-side content richness, rather than as a direct measure of factual superiority. We then select 100 biographies per non-English language, while applying coarse occupation-aware balancing over artists, scientists, athletes, and politicians. We then annotate each biography with a set of silver claims using the claim extraction model described in Section~\ref{sec:claim_extract}.

Table~\ref{tab:claw4-main-stats} summarizes the resulting benchmark. Here $r_t^{norm}$ denotes the language-calibrated non-English-to-English token ratio, $r_s$ the non-English-to-English sentence ratio, and $r_c$ the non-English-to-English claim ratio. Claim counts are computed from automatically derived reference claim sets using the best claim extraction method identified in Section~\ref{subsec:claim_extract_method}. As the table shows, non-English biographies are substantially richer across all three views, with 3.4 more tokens, 3.4 more sentences, and 2.9 more claims on average. Construction details are provided in Appendix~\ref{app:claw4-construction}.

\subsection{CLAW-4L-CX}
\label{sec:claw4-sent}

\textsc{CLAW-4L-CX} consists of sentences from our four languages which are manually annotated with English claims. We create this benchmark by first selecting sentences with various levels of complexity and then crowdsourcing the corresponding claims.

\paragraph{Sentence Selection from CLAW-4L.}
Our goal is to sample sentences spanning different levels of claim complexity, to avoid evaluating only simple one-claim cases. We use automatically extracted claims as a bootstrapping signal for sentence selection, not as gold labels. Specifically, we run the GPT-5.1-assisted claim extraction pipeline adapted from Claimify~\citep{metropolitansky-larson-2025-towards} described in Section~\ref{subsec:claim_extract_method}; its selection--decontextualization--decomposition structure provides sentence-level silver claims suitable for estimating claim complexity. We then bucket candidate sentences by silver claim count to obtain a balanced range of sentence complexity.

For each biography, we independently select one sentence from the English version and one from the non-English version, while enforcing a balanced distribution across claim-count buckets within each country-language setting. This yields 600 sentences in total: 300 in English and 300 in non-English languages,  with 25 sentences per claim-count bucket, and an average of 2.63 and 2.72 claims per sentence, respectively. Details of the sentence selection procedure and bucket statistics are provided in Appendix~\ref{subsec:claw-4L-CX-strategy}.

\paragraph{Crowd-Sourced Annotation.} 
Annotators verify, revise, and supplement the automatically generated claim drafts for each selected sentence, producing the final human reference claim sets. We recruited 13 English-fluent annotators who were pursuing or had completed a Ph.D. in computer science or a closely related field; for each cross-lingual annotation task, annotators were native speakers of the corresponding non-English language. Four annotators participated in the French--English setting, and three  in each of the Chinese--English and Azerbaijani--English settings. Before annotation, annotators read the task instructions and completed the task through the provided interface, shown in Appendix~\ref{app:claw4-sent-annotation}. A post-hoc claim-level agreement analysis shows high consistency, with Aligned-F1 above 91\% for all languages; details are in Appendix~\ref{app:sent_iaa}.

\subsection{CLAW-4L-RC}
\label{subsec:claw-4l-rc}

\textbf{CLAW-4L-RC} comprises 600 pairs of English claims annotated with labels describing the relation between the two claims.  Each pair originates from paired English and non-English biographies.

Candidate claim pairs are drawn from the 
\textsc{CLAW-4L} benchmark and labelled using a two-level label schema for enrichment-oriented alignment. At the top level, each pair is labeled as \textit{Aligned}, \textit{Contradicted}, or \textit{Not Relevant}. For aligned pairs, we further distinguish between exact equivalence (\textit{A$=$B}), one-sided enrichment (\textit{A$>$B} or \textit{B$>$A}), and mutual enrichment (\textit{A$\leftrightarrow$B}).

  

We first identify 100 exact-alignment pairs by selecting pairs with high LaBSE similarity and manually validating them. 
Starting from these 100 validated exact-alignment pairs, we construct the remaining relation types through controlled structured perturbation and GPT-5.1 rewriting, followed by human validation. Partial-alignment pairs are created by adding or removing non-conflicting detail, contradicted pairs by changing incompatible field values, and not-relevant pairs by pairing claims about different factual aspects. All generated pairs are manually checked against their intended labels, and invalid cases are corrected or rewritten.
The final benchmark contains 600 claim pairs: 100 exact-alignment pairs, 300 partial-alignment/enrichment pairs, 100 contradicted pairs, and 100 not-relevant pairs. Table~\ref{tab:claw4l-rc-label-dist} presents the label distribution and semantic similarity statistics; additional examples and construction details are provided in Appendix~\ref{app:claw4l_rc}.

\section{Processing Claims}
\label{sec:claim_extract}

We first describe  how we classify the relation between two claims,  how we extract claims, and how we use the relation between two claims to identify "enrichment claims" i.e., claims that are present in the non-English biography but not in the English one. In the next section (Section~\ref{sec:generation}), we describe our cross-lingual biography generation methods.

\subsection{Classifying Claim Pairs}
\label{sec:claim_classifier}

To support claim alignment and enrichment selection, we use an LLM-classifier that assesses the relation between two English-written claims. The classifier follows the coarse-to-fine schema of \textsc{CLAW-4L-RC}: it first predicts whether the pair is aligned, contradicted, or not relevant, and for aligned pairs further distinguishes exact alignment, one-sided enrichment, and mutual enrichment.
We select this classifier by evaluating open-source and closed-source LLMs, together with strong NLI baselines, on \textsc{CLAW-4L-RC}. 
We also test whether adding GPT-5.1-parsed structured claim fields (e.g., subject, predicate, object, etc.) improves relation judgment over claim text alone. 
Performance is summarized using three aggregate metrics: ARC, a macro-average over the three top-level relations; Align-FG, the average accuracy over the four fine-grained aligned labels; and Overall, the macro-average over all six labels.
Full settings, NLI comparisons, and structured-field ablations 
are provided in Appendix~\ref{app:llm_classifier} and Table~\ref{tab:infobox-ablation}. In the text-only setting used downstream, Qwen3.5-9B is the strongest open-source classifier, achieving 95.5 ARC, 89.5 Align-FG, and 93.2 Overall, close to GPT-5.1 at 95.9, 90.8, and 93.3. Structured fields do not consistently improve performance and can introduce parsing noise, so we use the  Qwen3.5-9B  text-only classifier in subsequent experiments.


\subsection{Extracting Claims}
\label{subsec:claim_extract_method}

We adapt five LLM-based claim extraction frameworks to our cross-lingual setting: FactScore, FactCheck-GPT, DnDScore, VeriScore, and Claimify, denoting the adapted variants with the prefix \textit{X-}. Each adapted framework accepts either an English or a non-English biography and outputs English factual claims. To isolate extraction-framework differences from backbone model quality, we instantiate all five frameworks with GPT-5.1.  Detailed method descriptions are provided in Appendix~\ref{app:claim_extraction_method}.

We evaluate the adapted frameworks on \textsc{CLAW-4L-CX}, where each sentence is independently annotated by three annotators, yielding three acceptable human reference claim sets. We use these multiple references since sentence-level factual claims can vary in wording and in how contextual information is made explicit; Appendix~\ref{app:sent_iaa} shows high post-hoc agreement among annotators.

\begin{table*}[!htbp]
\centering
\small
\setlength{\tabcolsep}{4.0pt}
\renewcommand{\arraystretch}{1.08}
\begin{tabular}{lcccccccccc}
\toprule
\multirow{2}{*}{\textbf{Method}}
& \multicolumn{2}{c}{\textbf{EN}}
& \multicolumn{2}{c}{\textbf{FR}}
& \multicolumn{2}{c}{\textbf{ZH}}
& \multicolumn{2}{c}{\textbf{AZ}}
& \multicolumn{2}{c}{\textbf{Avg.}} \\
\cmidrule(lr){2-3} \cmidrule(lr){4-5} \cmidrule(lr){6-7} \cmidrule(lr){8-9} \cmidrule(lr){10-11}
& \textbf{A-F1} & \textbf{E-F1}
& \textbf{A-F1} & \textbf{E-F1}
& \textbf{A-F1} & \textbf{E-F1}
& \textbf{A-F1} & \textbf{E-F1}
& \textbf{A-F1} & \textbf{E-F1} \\
\midrule
X-FactScore     & 52.89 & 27.83 & 57.89 & 35.23 & 52.51 & 26.30 & 54.04 & 39.51 & 54.33 & 32.22 \\
X-VeriScore     & 68.82 & 46.92 & 70.01 & 52.91 & 68.12 & 42.94 & 66.30 & 55.11 & 68.31 & 49.47 \\
X-DnDScore      & 41.91 & 11.61 & 46.16 & 16.08 & 44.75 & 16.02 & 44.76 & 22.87 & 44.40 & 16.65 \\
X-FactCheck-GPT & 59.60 & 43.70 & 62.13 & 49.87 & 54.70 & 49.82 & 52.91 & 44.38 & 57.34 & 46.94 \\
X-Claimify      & \textbf{75.17} & \textbf{72.44} & \textbf{78.05} & \textbf{73.56} & \textbf{71.12} & \textbf{71.91} & \textbf{70.52} & \textbf{66.80} & \textbf{73.72} & \textbf{71.18} \\
\bottomrule
\end{tabular}
\caption{
Claim extraction framework comparison on \textsc{CLAW-4L-CX} using GPT-5.1 as the common backbone.
A-F1 denotes Aligned-F1, which counts exact and partial alignment as matches; E-F1 denotes Exact-Aligned-F1, which counts only exact alignment.
Avg. is the macro-average over EN, FR, ZH, and AZ.
}
\label{tab:claim_extraction_framework_main}
\end{table*}

Because exact string matching cannot account for paraphrase or decontextualization differences, we evaluate extraction quality 
based on the relation predicted by the classifier from Section~\ref{sec:claim_classifier}  between the extracted claim and the reference claim. 
We also compute degree-normalized claim-level credits: a generated claim that matches multiple reference claims receives reduced precision credit, and a reference claim matched by multiple generated claims receives reduced recall credit. Since each sentence has three human reference sets, we keep the reference set with the highest Exact-Aligned-F1 and report \textbf{Aligned-F1} and \textbf{Exact-Aligned-F1}; the former counts exact and partial alignment as matches, while the latter counts only exact claim matches. Full metric definitions are provided in Appendix~\ref{subsec:eval_claim_extract_quality}.

Table~\ref{tab:claim_extraction_framework_main} reports the F1 scores. X-Claimify performs best across languages, especially under Exact-Aligned-F1. We therefore use 
X-Claimify to extract English reference claim sets for the full \textsc{CLAW-4L} biography pairs. These reference claims are used to select enrichment claims (Section~\ref{subsec:non-English_only_claim}) and to support downstream generation evaluation (Section~\ref{subsec:generation_eval}). Open-source backbone comparisons are provided in Appendix~\ref{app:claim_extraction_method} and Table~\ref{tab:claim_model_alignment}.





\subsection{Identifying Enrichment Claims}
\label{subsec:non-English_only_claim}


Given $\mathcal{C}^{en}$ the set of claims extracted from an English biography by the GPT-5.1 based X-Claimify extractor and $\mathcal{C}^{X}$, the set of claims extracted from the corresponding non-English biography, we compute the set of enrichment claims $\mathcal{C}^{add}$ in two steps as follows. For efficiency at biography scale, we first retrieve candidate English claims for each non-English-side claim using semantic similarity computed by All-MPNet-Base-v2, keeping only pairs above a cosine threshold of 0.7 and within the top-5 nearest neighbors; non-retrieved pairs are treated as not relevant. We then apply the classifier from Section~\ref{sec:claim_classifier} to the retained pairs. The classifier predicts whether each pair is aligned, contradicted, or not relevant, with aligned pairs further categorized into exact alignment and one-sided or mutual enrichment.

Retrieval serves only as a permissive coarse filter before LLM-based relation verification, rather than as the final alignment decision. In raw cosine units, mean All-MPNet similarities range from 0.843 to 0.960 across aligned relation types and reach 0.885 for contradicted pairs, compared with only 0.359 for not-relevant pairs (shown on a 0--100 scale in Table~\ref{tab:claw4l-rc-label-dist}). We therefore use 0.7 to remove clearly unrelated pairs while retaining semantically related candidates; on the full benchmark, 96\% of Non English-side claims retain at least one candidate. We cap each candidate list at five to bound the number of subsequent verifier calls, with detailed candidate-list statistics provided in Appendix~\ref{app:non-English_enrichment_claims}.

We discard non-English-side claims already fully covered by English, including exact matches (\(A{=}B\)) and English-more-specific alignments (\(A{>}B\)), where \(A\) is an English-side claim and \(B\) is a non-English-side claim. We retain non-English-additive alignments (\(B{>}A\) and \(A{\leftrightarrow}B\)), classifier-labeled conflicts (\(A{\perp}B\)), and non-English-side claims with no relevant English counterpart (\(A{\dashv}B\)) as enrichment candidates. Conflicts are retained because both sides are Wikipedia-derived and may require contextual reconciliation rather than automatic filtering. The resulting set of selected claims, denoted $\mathcal{C}^{add}$, is used as evidence for downstream English biography rewriting.

\begin{table}[!htbp]
\centering
\small
\begin{tabular}{lcc}
\toprule
\textbf{Non-English-side status} & \textbf{Avg.} & \textbf{Share} \\
\midrule
Covered by EN & 12.4 & 13.9\% \\
Non-English-additive & 23.4 & 26.2\% \\
Conflict & 10.2 & 11.4\% \\
Unmatched & 43.4 & 48.6\% \\
\midrule
Selected & 77.0 & 86.2\% \\
\bottomrule
\end{tabular}
\caption{\textbf{Average count, ratio and status of non-English-side claims in Claw-4L 400 biography pairs. }
Covered by EN includes exact matches and English-more-specific alignments. Selected includes non-English-additive, conflict, and unmatched claims. Full language-level statistics are in Appendix~\ref{app:non-English_enrichment_claims}.}
\label{tab:non-English_enrichment_main}
\end{table}

Table~\ref{tab:non-English_enrichment_main} summarizes the enrichment statistics of \textsc{CLAW-4L}. Only 13.9\% of non-English-side claims are already covered by English, while 26.2\% add detail to partially aligned English claims, and 48.6\% have no relevant English-side counterpart. Overall, 86.2\% of non-English-side claims are selected as enrichment evidence.

\section{Enriching English Biographies}
\label{sec:generation}

We compare three methods for generating the target English biography and evaluate them on \textsc{CLAW-4L} using claim extraction, classification and alignment, asking which method most enriches the original English biographies in terms of claims.

In all settings, the system receives the original English biography together with some form of non-English-side evidence and is prompted to produce an English biography that preserves the original English content while incorporating the additional information supported by the non-English input.

For generation, we compare three strong open-weight multilingual instruction-tuned generators: Qwen3.6-27B~\citep{yang2025qwen3technicalreport}, Gemma-4-31B-it\footnote{\url{https://huggingface.co/google/gemma-4-31B-it}}, and Mistral-3.2-24B-it\footnote{\url{https://huggingface.co/mistralai/Mistral-Small-3.2-24B-Instruct-2506}}. We choose these models to cover high-performing model families developed in different linguistic ecosystems, allowing us to examine whether enrichment behavior varies across Chinese, French, and lower-resource Azerbaijani settings. 

The three methods differ only in
how information from the non-English input is represented and processed: (i) using the raw non-English biography as input and relying on the cross-lingual capacity of LLMs; (ii) using a machine-translated version of the non-English biography with an LLM rewriting conditioned on both the original English text and the translated content; and (iii) generating conditioned on the English biography together with the enrichment claims derived from the non-English biography.

\subsection{Methods}
\label{subsec:generation_methods}

\paragraph{Cross-Lingual Generation.}
This is the most direct baseline. We provide the original English biography together with the raw non-English biography and ask the generator to produce an enriched English rewrite based on these two inputs. Because biographies can be long (up to 28,420 tokens; Table~\ref{tab:claw4-main-stats}), we process the non-English biography incrementally at the section level. We cap the English biography input at 4,096 tokens and the non-English-side context at 2,048 tokens per step; when needed, long sections are further split. Each non-English-side chunk updates the current English biography, with later generations conditioned on the previously enriched version.

\paragraph{Machine Translation and Generation Pipeline.}
This setting decouples translation from enrichment. We first translate the non-English biography into English using a language-specific translation model, and then apply the same incremental enrichment procedure as in the \textit{Cross-lingual Generation} approach. To choose the translation model for each language, we compare candidate translators using GPT-5.1-extracted non-English-side claims as semantic references and select the model that best preserves non-English-side claims after translation. The selected translation models and evaluation details are provided in Appendix~\ref{app:mt_selection_bio}.

\paragraph{Claim-Based Enrichment.}
This setting exploits the enrichment claims $\mathcal{C}^{add}$ derived from the non-English biographies (cf. Section~\ref{subsec:non-English_only_claim}). We provide the original English biography, along with $\mathcal{C}^{add}$, and ask the generator to produce an enriched English rewrite. The selected claims are not mechanically appended to the original article; instead, they serve as structured evidence for rewriting the full biography while preserving and coherently integrating the original English content. Compared with the other two settings, this formulation filters out already-covered information before generation, reducing context length and focusing the model on incremental factual additions.

\subsection{Evaluation}
\label{subsec:generation_eval}

We evaluate generated biographies on \textsc{CLAW-4L} by comparing the claims extracted from the generated biographies ($\hat{\mathcal{C}}^{gen}$) with the claims extracted from both the original English (${\mathcal{C}}^{en}$) and the Non English Biography (${\mathcal{C}}^{X}$).
To reduce large-scale evaluation cost, claim extraction is performed using X-Claimify with Gemma-4-31B-it, the strongest open-source English extractor in our backbone comparison. We compare generated claims against the reference pool $\mathcal{C}^{ref}=\mathcal{C}^{en}\cup\mathcal{C}^{X}$. As in biography-level claim alignment (Section~\ref{subsec:non-English_only_claim}), for each generated claim, we retrieve candidate reference claims using All-MPNet-Base-v2 cosine similarity (threshold 0.7, top-5), then apply the claim-pair classifier to the retained pairs.

We report metrics for claim growth, factual support, hallucination, and an overall trade-off score.
For each generated claim \(B \in \hat{\mathcal{C}}^{gen}\), we categorize it against the reference pool with contradiction priority: \(B\) is \textit{contradicted} if any retained reference claim \(A\) is labeled \(A{\perp}B\); otherwise, \(B\) is \textit{valid} if some reference claim fully supports it (\(A{=}B\) or \(A{>}B\)); all remaining claims are \textit{unsupported}. Let \(\mathcal{V}\), \(\mathcal{K}\), and \(\mathcal{U}\) denote the valid, contradicted, and unsupported generated claims.
We define
\[
    \begin{aligned}
    \mathrm{Valid\%}
    &= 100\frac{|\mathcal{V}|}{|\hat{\mathcal{C}}^{gen}|},\\
    \mathrm{Halluc.\%}
    &= 100\frac{|\mathcal{K}|+|\mathcal{U}|}
    {|\hat{\mathcal{C}}^{gen}|}.
    \end{aligned}
\]
Claim growth is measured by \(\#\mathrm{Gain}=|\hat{\mathcal{C}}^{gen}|-|\mathcal{C}^{en}|\), with Coverage normalizing this gain by \(|\mathcal{C}^{add}|\), the enrichment claims from Section~\ref{subsec:non-English_only_claim}. We then define \(\#\mathrm{Valid}=\mathrm{Valid}\times\#\mathrm{Gain}\) as an estimated supported gain, where \(\mathrm{Valid}\) is \textbf{Valid\%} converted to a proportion.

Our primary overall metric is \textbf{Bal.}, which combines supported addition quantity and factual reliability. For each system \(s\), we define
\[
    \begin{gathered}
    g_s = \#\mathrm{Valid}_s, \qquad
    v_s = \mathrm{Valid\%}_s,\\
    \tilde{x}_s = \frac{x_s-\min_{s'}x_{s'}}
    {\max_{s'}x_{s'}-\min_{s'}x_{s'}},
    \quad x\in\{g,v\},\\
    \mathrm{Bal.}_s = 50(\tilde{g}_s+\tilde{v}_s).
    \end{gathered}
\]
where the minimum and maximum are computed over all systems in Table~\ref{tab:generation_full_results}. Bal. ranges from 0 to 100 and equally weights normalized supported claim growth and factual reliability. Since \(\mathrm{Valid\%}=100-\mathrm{Halluc.\%}\) under our categorization, a higher Bal. indicates more supported additions with less hallucination. Additional metric details and full model-level results are provided in Appendix~\ref{app:generation_eval_results}.

\paragraph{Human evaluation.}
We additionally conduct a writing-quality sanity check on 45 Mistral-generated biographies: three languages, five length-stratified biography pairs per language, and three enrichment methods. For each selected pair, we evaluate the Raw, Translation, and Claims outputs for the same entity. Three computer-science Ph.D. researchers who use English as a working language independently rate every output on a 1--5 Likert scale for Readability/Fluency, Coherence/Integration (including abrupt insertion of new information), and Wikipedia-Style Writing. Appendix~\ref{app:generation_human_eval} provides the sampling protocol, full length-stratified results, and agreement statistics.

\subsection{Results and Analysis}
\begin{figure}[!htbp]
    \centering
    \includegraphics[width=1.0\linewidth]{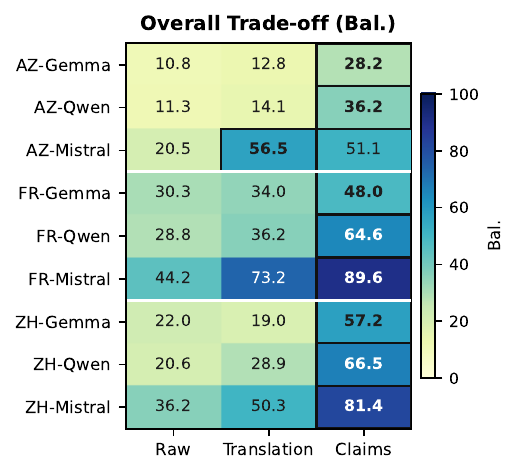}
    \caption{Overall generation trade-off across languages, generators, and non-English-evidence formats. Each cell reports Bal.; higher is better. Bolded cells mark the best evidence format for each language--generator pair.}
    \label{fig:generation_bal_heatmap}
\end{figure}

Figure~\ref{fig:generation_bal_heatmap} visualizes the primary Bal. score for all language, generator, and evidence-format combinations, while Figure~\ref{fig:halluc_supported_add_plot} summarizes the average \#Valid--hallucination trade-off across evidence formats. Full metric values are provided in Table~\ref{tab:generation_full_results}.

\paragraph{Evidence format.}
Table~\ref{tab:generation_evidence_avg} summarizes results across evidence formats by averaging across all languages and generators. Translation improves supported additions over raw non-English biographies, but does not reliably reduce hallucination; it increases \#Valid from 15.89 to 19.05, while hallucination remains almost unchanged (29.49 vs.\ 29.59). Claim-based evidence achieves the best trade-off, with the highest average \#Valid (20.47) and substantially lower hallucination (23.54). This suggests that our claim-based pipeline reduces three sources of difficulty: long non-English contexts, cross-lingual generation, and implicit alignment between English and non-English-side evidence.

\paragraph{Lower-resource exception.}
The only setting where translation outperforms claim evidence in Bal. is Azerbaijani with Mistral. We hypothesize that this reflects error propagation in the upstream claim pipeline for the lower-resource language: if claim extraction misses useful evidence, the downstream claim-based generator receives a cleaner but less complete evidence set. This is consistent with the cross-lingual claim extraction results in Table~\ref{tab:claim_extraction_framework_main}, where lower-resource inputs are more challenging than high-resource ones. In contrast, for Gemma and Qwen on Azerbaijani, claim evidence still improves Bal. over raw and translated biographies, suggesting that filtering and English normalization remain beneficial when the generator is more prone to hallucination.

\paragraph{Language and generator effects.}
Performance is substantially stronger for French and Chinese than for Azerbaijani, suggesting that enriching women from lower-resource language contexts remains challenging, even with evidence from the non-English Wikipedia. Across generators, Mistral achieves the highest Bal. in every language-evidence setting. The full results in Table~\ref{tab:generation_full_results} show that this is largely driven by lower hallucination rates while maintaining competitive supported additions, consistent with more conservative or better-grounded rewriting behavior.

\paragraph{Human writing quality.}
Claims obtains the highest Readability/Fluency (3.933) and Wikipedia-Style Writing (3.822) scores, while Raw is slightly higher in Coherence/Integration (3.911 vs.\ 3.867). Translation scores 3.756, 3.733, and 3.689 on the three respective dimensions. Thus, using selected claims as structured evidence does not incur an evident writing-quality penalty relative to the other enrichment pipelines. This result is a descriptive sanity check rather than a significance-tested comparison with the original English biographies; Appendix~\ref{app:generation_human_eval} provides the complete analysis.

\paragraph{Qualitative error analysis.}
We manually inspected 18 Mistral-generated biographies from six controlled cases. For each language, we selected one biography pair with the highest and one with the lowest number of verifier-supported output claims under its best-performing evidence format---Claims for French and Chinese, and Translation for Azerbaijani---and compared the Raw, Translation, and Claims outputs for the same entities. In these cases, claim-based enrichment generally added more targeted, fine-grained events and relations, whereas raw cross-lingual enrichment retained a more narrative style but provided less focused coverage. Translation recovered many details when reliable, but was more sensitive to source and translation noise. Missing claims mainly involved long-tail works, institutional roles, and event metadata. Remaining hallucinations included incorrect temporal, geographic, and educational details, unsupported over-specific enumerations, and, in one difficult translation-based case, repetitive generation. Appendix~\ref{app:generation_error_analysis} details the selection protocol and provides representative examples.

\section{Conclusion}

We introduced cross-lingual biography enrichment, in which an existing English biography is enriched with a non-English Wikipedia biography of the same person. We constructed \textsc{CLAW-4L}, a benchmark of English biographies paired with French, Chinese, and Azerbaijani biographies about the same women, together with resources for evaluating claim extraction and alignment. We also developed a claim-based framework that maps both biographies into a shared English claim space, selects enrichment evidence from non-English data, and uses it for controlled English rewriting.

Experiments show that claim-based evidence offers the strongest trade-off between supported additions and hallucination in most language-generator settings. Compared with raw non-English biographies and translations, selected claims reduce long-context burden and enhance interpretability.
A complementary human evaluation finds no evident writing-quality penalty from using selected claims as rewriting evidence relative to the other enrichment pipelines.
The Azerbaijani results also show that lower-resource settings remain difficult, suggesting that reliable cross-lingual enrichment depends on robust claim extraction and alignment.

\section*{Limitations}

\paragraph{Benchmark scope.}
Our benchmark focuses on women biographies across three non-English languages: French, Chinese, and Azerbaijani. This design lets us study enrichment in both higher-resource and lower-resource settings, but it does not cover the full diversity of Wikipedia language editions, scripts, regions, or biography types. Extending the benchmark to more languages and entity groups is an important direction for future work.

\paragraph{Source reliability.}
We use Wikipedia biographies as curated encyclopedic evidence, but we do not assume that every statement in every language edition is factually correct. Conflicts between English and non-English-language claims may reflect errors, temporal mismatch, or missing context. Our framework surfaces such cases for downstream reconciliation rather than resolving factual truth against external sources.

\paragraph{Factual verification and revision.}
While claim-based evidence achieves the lowest hallucination rate among the three settings, post-generation verification could further strengthen factual reliability. A natural extension is a generate--verify--revise loop. After producing an initial biography, the system could decompose it into claims and verify each claim against the input evidence using claim-level factuality and grounding methods such as FactScore, VeriScore, FactCheck-GPT, or MiniCheck~\citep{min-etal-2023-factscore,song-etal-2024-veriscore,wang-etal-2024-factcheck,tang-etal-2024-minicheck}. Unsupported or contradicted claims could then be removed or corrected before a final rewriting step. Verification could further retrieve independent evidence from structured knowledge bases, citations associated with the source articles, and corroborating evidence from additional Wikipedia language editions, building on retrieval-grounded and citation-aware generation~\citep{lewis2021retrievalaugmentedgenerationknowledgeintensivenlp,fan-gardent-2022-generating,zhang-etal-2025-wikigenbench}. This extension would complement our current contribution of explicitly selecting non-English enrichment evidence before generation with evidence-based factuality control after generation. For deployment-oriented settings, evidence provenance could additionally be exposed to human editors, with unresolved conflicts triggering abstention rather than automatic insertion.

\paragraph{Pipeline errors.}
Our pipeline also depends on automatic claim extraction and claim-pair relation judgment. Although we evaluate these components on \textsc{CLAW-4L-CX} and \textsc{CLAW-4L-RC}, extraction or alignment errors can still propagate to enrichment generation, especially for lower-resource languages. This limitation is reflected in the Azerbaijani results, where upstream claim processing can miss useful non-English-side evidence.

\paragraph{Evaluation and reproducibility.}
Our automatic generation evaluation emphasizes claim-level supported additions and hallucinations. We supplement it with a human evaluation of readability, coherence, and Wikipedia-style writing, but this remains a small-scale sanity check covering 15 biography pairs, one generator, and three expert annotators. It does not directly compare enriched outputs with the original English biographies, and method labels were visible to annotators. Accordingly, it is not a comprehensive assessment of editorial quality or deployment readiness. We also use GPT-5.1 for reference claim extraction and several prompt-assisted construction steps; while we release prompts, scripts, and open-weight alternatives, exact reproduction may be affected by closed-source model changes.


\paragraph{Beyond English.}
A broader direction is to use cross-lingual evidence not only to enrich English biographies, but also to improve biography coverage in lower-resource Wikipedia editions. Our current formulation treats English as the generation target, reflecting its role as a default encyclopedic source in NLP. Future work should study bidirectional and many-to-many enrichment settings, where claims from multiple language editions are aligned and used to support biography expansion in lower-resource languages.

\section*{Acknowledgments}
We thank the anonymous reviewers for their feedback. This work received government funding managed by the French National Research Agency under France 2030, reference number ``ANR-23-IACL-0004'' (AI Chair Gardent: ``Semantically Consistent LLM Based Text Generation''). Experiments presented in this paper were carried out using the Grid'5000 testbed, supported by a scientific interest group hosted by Inria and including CNRS, RENATER, and several universities as well as other organizations (see \url{https://www.grid5000.fr}). This work was also granted access to the HPC resources of IDRIS under the allocation AD011016561 made by GENCI.

\bibliography{custom}

\appendix
\label{sec:appendix}

\section{Implementation Details}
\label{app:implementation_details}

\subsection{Inference Setup}

All open-weight model inference is run with the MS-SWIFT~\citep{zhao2025swiftascalablelightweightinfrastructure} framework and the vLLM~\citep{kwon2023efficientmemorymanagementlarge} inference engine on a single NVIDIA A100 GPU with 80 GB of memory. We use this setup for open-weight generation, translation, claim extraction backbone comparisons, and classifier experiments unless otherwise specified. Closed-source GPT-5.1~\citep{singh2026openaigpt5card} calls are made through the OpenRouter API~\footnote{\url{https://openrouter.ai/}} and are used for reference claim extraction, prompt-assisted dataset construction, and closed-source comparison settings described in the main paper.

\subsection{Experimental Cost}
The open-weight model inference for enrichment generation requires only 2 hours with the help of vLLM, but the claim relation evaluation takes 4 hours per generated biography. 
The claim extraction pipeline with GPT-5.1 + X-Claimify on our CLAW-4L costs around \$ 300 via API calls, and the entire extraction procedure takes 10 hours. 

\subsection{Decoding Settings and Token Budgets}

Table~\ref{tab:hyperparameters} reports the token budgets and decoding settings used in our generation, translation, and LLM-based verification experiments. We use deterministic decoding for verification and model-specific generation settings for biography rewriting and translation. Qwen and Gemma generation runs use thinking disabled.

\begin{table}[!h]
\centering
\scriptsize
\setlength{\tabcolsep}{2.7pt}
\renewcommand{\arraystretch}{1.12}
\begin{tabular}{@{}lcccccc@{}}
\toprule
\multicolumn{7}{@{}l}{\textbf{Token Budgets}} \\
\midrule
\textbf{Task} & \textbf{Prompt} & \textbf{EN} & \textbf{non-English} & \textbf{Out.} & \textbf{Seed} & \\
\midrule
Translation & 4,096 & 4,096 & -- & 8,192 & 42 & \\
Enrichment & 2,048 & 4,096 & 2,048 & 8,192 & 42 & \\
Verification & 8,192 & -- & -- & 8,192 & 42 & \\
\midrule

\multicolumn{7}{@{}l}{\textbf{Translation / Enrichment}} \\
\midrule
\textbf{Model} & \textbf{Temp.} & \textbf{Top-$p$} & \textbf{Top-$k$} & \textbf{Pres.} & \textbf{Rep.} & \\
\midrule
Qwen3.6-27B & 0.7 & 0.80 & 20 & 1.5 & 1.0 & \\
Gemma-4-31B-IT & 1.0 & 0.95 & 64 & 0.0 & 1.0 & \\
Mistral-Small-3.2-24B-IT & 0.15 & 1.00 & 50 & 0.0 & 1.0 & \\
\midrule

\multicolumn{7}{@{}l}{\textbf{Verification}} \\
\midrule
\textbf{Model} & \textbf{Temp.} & \textbf{Top-$p$} & \textbf{Top-$k$} & \textbf{Pres.} & \textbf{Rep.} & \\
\midrule
Qwen3.5-9B & 0.0 & 1.00 & 0 & 0.0 & 1.0 & \\
\bottomrule
\end{tabular}
\caption{Hyperparameters for generation and LLM-based verification experiments. The upper panel reports token budgets, where EN denotes the English biography, non-English denotes the non-English biography or enrichment context, and Out. denotes the maximum number of generated tokens. The lower panels report decoding settings for generation and verification. Pres. denotes presence penalty and Rep. denotes repetition penalty. Qwen and Gemma generation runs use thinking disabled.}
\label{tab:hyperparameters}
\end{table}

\section{Datasets Construction Details}
\label{app:claw4-construction}
  
In this section, we present the details of the creation of our evaluation datasets: (i) CLAW-4L, a cross-lingual biography pairs benchmark, (ii) CLAW-4L-CX, a cross-lingual sentence-claim benchmark, and (iii) CLAW-4L-RC, an English claim-pair relation judgment benchmark.

\subsection{Data Collection from Wikidata}
We first collect country-specific female candidates from Wikidata using three eligibility constraints: each entity must be (1) a human, (2) marked as female, and (3) associated with the corresponding country of citizenship. In addition, we require that the entity have both an English Wikipedia sitelink and a sitelink in the non-English language. We instantiate this procedure for three country-language settings: France/French, China/Chinese, and Azerbaijan/Azerbaijani.

To support subsequent filtering and sampling, we intentionally build candidate pools that are substantially larger than the final benchmark size. For the French and Chinese settings, we cap the pool size at 1,500 entities each. For the Azerbaijani setting, only 434 Wikidata entities satisfy all constraints, and we therefore use the full eligible pool. The final candidate pool used for selection contains 3,434 entities in total.

For each retained entity, we additionally collect basic metadata from Wikidata, including the entity identifier, page titles, sitelinks, and occupation labels. These metadata are used only for filtering, identifying, and coarse occupation-aware sampling.

\subsection{Biography Retrieval and Cleaning}
For each candidate, we retrieve the English and non-English-language Wikipedia pages through the MediaWiki API. Since our goal is to compare biography content across languages, we apply a lightweight normalization procedure to both pages before computing any statistics.

Specifically, we remove markup artifacts and non-biographical content that would otherwise inflate page length without contributing useful biography evidence. This includes infobox templates, formatting noise, and trailing sections such as references, bibliography, notes, and external links. After cleaning, we retain the main readable biography prose together with its section structure when available.
This cleaning procedure is designed to improve comparability across languages rather than to produce perfectly parallel biographies. In particular, the cleaned texts may still differ substantially in discourse structure and content organization, but the most obvious length distortions caused by non-biographical material are removed.


\subsection{Length-based Tiering and Language Calibration}
\label{app:length_tier}
After cleaning, we compute biography lengths using the same multilingual tokenizer (\texttt{google/mt5-small}). For a non-English language $L$, we first define the raw non-English-to-English token ratio for a biography pair as
\begin{equation}
r = \frac{\#\text{tokens(non-EN)}}{\#\text{tokens(EN)}}.
\end{equation}

Because token counts produced by a shared multilingual tokenizer may still reflect language- and script-specific segmentation tendencies, raw ratios can conflate content richness with tokenization bias. To reduce this effect, we estimate a language-specific inflation factor for each non-English language using parallel multilingual data from FLORES+.\footnote{\url{https://huggingface.co/datasets/openlanguagedata/flores_plus}} Concretely, given aligned sentence pairs $(x_{en}, x_L)$, we compute
\begin{equation}
b_L = \mathbb{E}_{(x_{en}, x_L)} \left[ \frac{\#\text{tokens}(x_L)}{\#\text{tokens}(x_{en})} \right],
\end{equation}
where all token counts are obtained with the same tokenizer.

We then define the normalized biography-level ratio as
\begin{equation}
r^{\text{norm}} = \frac{r}{b_L}
= \frac{\#\text{tokens(non-EN)}}{\#\text{tokens(EN)}} \cdot \frac{1}{b_L}.
\end{equation}

Table~\ref{tab:claw4-inflation-factors} reports the resulting language-specific inflation factors. As expected, the estimated factors differ substantially across non-English languages, reflecting systematic tokenization differences even for semantically aligned text. Notably, the calibration factors are greater than 1 for French and Azerbaijani, but below 1 for Chinese, indicating that the shared tokenizer tends to allocate relatively more tokens to the former two languages and fewer to the latter when compared with English.

We use $r^{\text{norm}}$ as a coarse heuristic for non-English-side content richness. Importantly, it is not intended as a direct measure of factual quality or factual correctness; rather, it provides a more comparable cross-lingual signal for prioritizing cases in which the non-English biography is likely to contain additional textual evidence beyond the English page. Since $r^{\text{norm}}$ remains centered around 1.0 for content-equivalent text under the calibration corpus, we partition candidates into three tiers:
\begin{itemize}
    \item \textbf{Tier A}: $r^{\text{norm}} \geq 1.20$
    \item \textbf{Tier B}: $1.00 \leq r^{\text{norm}} < 1.20$
    \item \textbf{Tier C}: $r^{\text{norm}} < 1.00$
\end{itemize}

We adopt a tiered design rather than a hard filter because biography length remains only a proxy for content richness. Tiering lets us prioritize non-English-richer candidates while preserving enough flexibility for downstream sampling decisions on occupation coverage and country-specific composition.

\subsection{Sentence Segmentation and Sentence-level Statistics}
\label{app:sent_seg}
For sentence-level statistics, we segment biographies using a multilingual neural sentence segmentation model (SaT) implemented via \texttt{wtpsplit}\footnote{We adopt \texttt{sat-3l-sm} (\url{https://huggingface.co/segment-any-text/sat-3l-sm}) for a trade-off between speed and performance.}. This model provides consistent sentence boundary detection across languages without relying on language-specific punctuation rules.

We define the sentence-level ratio analogously as
\begin{equation}
r_s = \frac{\#\text{sentences(non-EN)}}{\#\text{sentences(EN)}}.
\end{equation}

We do not apply an analogous calibration to sentence-level ratios. Unlike token counts, sentence segmentation is inherently less stable across languages and is more sensitive to discourse structure, punctuation conventions, and editorial practices. In addition, large-scale paragraph-aligned multilingual corpora covering lower-resource languages are scarce, making reliable estimation of sentence-level calibration factors impractical. We therefore treat sentence counts and sentence ratios only as complementary structural signals rather than precise units of semantic comparison.

\begin{table}[t]
\centering
\small
\begin{tabular}{lc}
\toprule
\textbf{Pair} & \textbf{Inflation Factor $b_L$} \\
\midrule
EN$\rightarrow$ZH & 0.91 \\
EN$\rightarrow$FR & 1.41 \\
EN$\rightarrow$AZ & 1.36 \\
\bottomrule
\end{tabular}
\caption{Language-specific token inflation factors estimated from FLORES+ using the \texttt{google/mt5-small} tokenizer. Values above 1 indicate that the non-English language tends to produce more tokens than English for semantically aligned text; values below 1 indicate the opposite.}
\label{tab:claw4-inflation-factors}
\end{table}

\subsection{Tier-aware Selection}
In this step, we aim to select candidates for whom the non-English-language page is more likely to contain additional biography content beyond the English page, while also balancing the occupation- and country-wise distribution.

As described in~\ref{app:length_tier}, we assign all 3,434 candidates into three tiers by normalized ratio $r^{\text{norm}}$. For each country, we select 300 instances as the source pool for downstream benchmark construction. The selection procedure is tier-aware: we first prioritize candidates from Tiers A and B, and use Tier C as a fallback only when necessary to complete the country quota.

Within the prioritized Tier A/B subset, we apply coarse occupation-aware sampling. Specifically, we pre-define four biography-rich occupation groups---\textit{artist}, \textit{athlete}, \textit{scientist}, and \textit{politician}---and infer for each candidate a single occupation bucket from its Wikidata occupation labels using keyword matching. When possible, the required Tier A/B quota is distributed approximately evenly across these four groups. This balancing is approximate rather than strict: if one group does not contain enough eligible A/B candidates, the remaining quota is filled from the rest of the available Tier A/B pool. Candidates outside the four focus groups are assigned to an \textit{other} bucket. These candidates are not explicitly balanced during the initial A/B allocation, but they remain eligible for final selection when filling the remaining quota.

Algorithm~\ref{alg:claw4-selection} summarizes the country-wise sampling procedure. Table~\ref{tab:claw4-pool-final} reports the candidate-pool size changes.


\begin{algorithm}[t]
\caption{Country-wise sampling procedure for CLAW-4L}
\label{alg:claw4-selection}
\small
\begin{algorithmic}[1]
\Require country-specific candidate set $\mathcal{C}$, non-English size $N$, minimum A+B ratio $\rho$
\State Keep only candidates with both English and non-English biographies
\State Infer one coarse occupation bucket for each candidate: 
\textit{artist}, \textit{scientist}, \textit{athlete}, \textit{politician}, or \textit{other}
\State Partition candidates into Tier A, Tier B, and Tier C
\State Let $N_{AB} = \min(|A \cup B|, \lceil \rho N \rceil)$
\State Select $N_{AB}$ candidates from Tier A/B
\If{occupation-aware balancing is enabled}
    \State Distribute the A/B non-English as evenly as possible across \textit{artist}, \textit{scientist}, \textit{athlete}, and \textit{politician}
    \State Fill any unassigned quota from the remaining Tier A/B pool
\Else
    \State Select from Tier A first, then Tier B
\EndIf
\State If fewer than $N$ candidates have been selected, continue adding remaining Tier A/B candidates
\State If the quota is still not met, fill the remaining slots from Tier C
\State \Return final selected set
\end{algorithmic}
\end{algorithm}

\begin{table}[t]
\centering
\small
\begin{tabular}{lrr}
\toprule
\textbf{Country} & \textbf{Before} & \textbf{After} \\
\midrule
France & 1500 & 300 \\
China & 1500 & 300 \\
Azerbaijan & 434 & 300 \\
\midrule
Total & 3434 & 900 \\
\bottomrule
\end{tabular}
\caption{Candidate-pool sizes change before/after tier-aware selection.}
\label{tab:claw4-pool-final}
\end{table}

\subsection{Stratified Biography Sampling for CLAW-4L}
\label{app:claw4l-sampling}

To obtain the final biography-claim benchmark, CLAW-4L, we design a stratified sampling strategy to reduce the dataset to 300 instances (100 per country), while maintaining the same data distribution as in the remaining 900-instance candidate pool.

\paragraph{Matched attributes.}
Within each country, we aimed to preserve the empirical distribution of the following attributes:
(i) occupation bucket,
(ii) selection tier,
(iii) normalized non-English-to-English token ratio,
(iv) non-English-side token length,
(v) nationality-count bucket, and
(vi) non-English-side infobox presence.
Occupation buckets consist of \textit{scientist}, \textit{athlete}, \textit{politician}, \textit{artist}, and \textit{other}. Nationality multiplicity was bucketed as 1, 2, or 3+. Infobox availability was treated as a binary attribute indicating whether non-empty structured infobox information was present on the non-English side.

\paragraph{non-English distributions.}
For each country-specific 300-instance pool, we derived a 100-instance non-English distribution for every matched attribute. For categorical variables, we scaled empirical counts from 300 to 100 and converted the resulting fractional allocations into integers using the largest-remainder method. This preserves the original proportions as closely as possible under the fixed sample size constraint.

For continuous variables, we did not attempt to match only the mean or variance. Instead, we discretized values into country-specific quantile bins and matched the empirical histogram over bins. Specifically, we used quantile-based bins for normalized token ratio and non-English-side token count, then applied the same largest-remainder procedure to obtain integer non-English counts for each bin. This design preserves the shape of the source distribution more faithfully than moment matching alone.

\paragraph{Constrained subset optimization.}
Given the non-English distributions, we selected a subset of size 100 for each country using constrained local search. Occupation counts were enforced exactly as a hard constraint. We first generated an initial feasible subset by randomly sampling the required number of instances from each occupation bucket. We then refined this subset by repeatedly swapping within-occupation, replacing one selected instance with an unselected instance from the same occupation bucket. This guarantees that occupation counts remain fixed throughout optimization.

All remaining attributes were optimized jointly using a weighted mismatch objective:
\begin{equation}
\begin{aligned}
\mathcal{L}(S) =\;&
\lambda_{\text{tier}} d_{\text{tier}}(S)
+ \lambda_{\text{ratio}} d_{\text{ratio}}(S) \\
&+ \lambda_{\text{tok}} d_{\text{tok}}(S)
+ \lambda_{\text{nat}} d_{\text{nat}}(S) \\
&+ \lambda_{\text{info}} d_{\text{info}}(S),
\end{aligned}
\end{equation}
where \(S\) is a candidate subset of size 100. Here,
\(d_{\text{tier}}, d_{\text{ratio}}, d_{\text{tok}},\) and \(d_{\text{nat}}\) are normalized \(L_1\) distances between the sampled and non-English count distributions for selection tier, normalized token-ratio bins, non-English-token bins, and nationality-count bins, respectively, and \(d_{\text{info}}\) is the absolute difference in non-English-infobox presence rate. Occupation is excluded from the objective because it is enforced exactly.

In our implementation, we used
\(\lambda_{\text{tier}}=4.0\),
\(\lambda_{\text{ratio}}=2.0\),
\(\lambda_{\text{tok}}=1.5\),
\(\lambda_{\text{nat}}=1.0\), and
\(\lambda_{\text{info}}=0.8\),
reflecting the priority of preserving the original tier structure while also maintaining close agreement on length- and metadata-related attributes.

\paragraph{Search procedure.}
Because this subset-selection problem is combinatorial, we used a multi-start local search strategy. For each country, we initialized the search from multiple random occupation-matched subsets and ran a fixed number of local swap steps from each initialization. We retained the subset with the lowest objective value across all runs. In practice, this procedure consistently found samples that matched the source distributions exactly or near-exactly on all tracked dimensions.

\paragraph{Outcome.}
The resulting 300-instance CLAW-4L closely tracks the composition of the 900-instance pool. Across all three country settings, occupation and selection-tier distributions were exactly matched. The remaining distributions over length bins, nationality-count bins, and infobox availability were also matched exactly in most cases, with only very small residual deviations in a few settings. This sampled biography pool was then used as the source set for constructing CLAW-4L-CX, our sentence-level human annotation benchmark.

We summarize the occupation-bucket distribution of the CLAW-4L in Table~\ref{tab:human-val-country-occ} and compare token- and sentence-level statistics between the 900-instance pool and the sampled CLAW-4L in Table~\ref{tab:claw4-full-vs-sampled}.

\begin{table*}[t]
\centering
\small
\begin{tabular}{lrrrrrr}
\toprule
\textbf{Country} & \textbf{N} & \textbf{Artist} & \textbf{Athlete} & \textbf{Politician} & \textbf{Scientist} & \textbf{Other} \\
\midrule
France      & 100 & 32 (32.00\%) & 26 (26.00\%) & 26 (26.00\%) & 10 (10.00\%) & 6 (6.00\%) \\
China       & 100 & 30 (30.00\%) & 35 (35.00\%) & 19 (19.00\%) & 3 (3.00\%)  & 13 (13.00\%) \\
Azerbaijan  & 100 & 32 (32.00\%) & 29 (29.00\%) & 7 (7.00\%)   & 7 (7.00\%)  & 25 (25.00\%) \\
\midrule
Total       & 300 & 94 (31.33\%) & 90 (30.00\%) & 52 (17.33\%) & 20 (6.67\%) & 44 (14.67\%) \\
\bottomrule
\end{tabular}
\caption{Occupation-bucket distribution in CLAW-4L, reported by country and in total. We consider four major categories (\textit{artist}, \textit{athlete}, \textit{politician}, \textit{scientist}), with remaining biographies grouped as \textit{other}.}
\label{tab:human-val-country-occ}
\end{table*}

\begin{table*}[t]
\centering
\small
\setlength{\tabcolsep}{4pt}
\renewcommand{\arraystretch}{1.08}
\begin{tabular}{llcccccc}
\toprule
\textbf{Split} & \textbf{Pair} & \textbf{N} & \textbf{Tokens (EN / non-EN)} & \textbf{Mean $r_t$} & \textbf{Mean $r_t^{norm}$} & \textbf{Sent (EN / non-EN)} & \textbf{Mean $r_s$} \\
\midrule
\multirow{4}{*}{Candidate Pool}
& EN$\rightarrow$FR & 300 & 479.22 / 1673.88 & 5.88 & 4.16 & 16.02 / 42.53 & 4.47 \\
& EN$\rightarrow$ZH & 300 & 498.84 / 1090.05 & 2.75 & 3.00 & 16.92 / 28.59 & 2.24 \\
& EN$\rightarrow$AZ & 300 & 543.36 / 1713.26 & 4.43 & 3.26 & 18.43 / 47.05 & 3.74 \\
\cmidrule(lr){2-8}
& \textbf{EN$\rightarrow$X} & \textbf{900} & \textbf{507.14 / 1492.39} & \textbf{4.35} & \textbf{3.47} & \textbf{17.12 / 39.39} & \textbf{3.48} \\
\midrule
\multirow{4}{*}{CLAW-4L}
& EN$\rightarrow$FR & 100 & 465.44 / 1867.14 & 5.92 & 4.19 & 16.08 / 48.41 & 4.64 \\
& EN$\rightarrow$ZH & 100 & 489.23 / 1041.28 & 2.60 & 2.83 & 15.75 / 27.18 & 2.01 \\
& EN$\rightarrow$AZ & 100 & 544.17 / 1967.12 & 4.35 & 3.20 & 19.37 / 55.21 & 3.48 \\
\cmidrule(lr){2-8}
& \textbf{EN$\rightarrow$X} & \textbf{300} & \textbf{499.61 / 1625.18} & \textbf{4.29} & \textbf{3.41} & \textbf{17.07 / 43.60} & \textbf{3.38} \\
\bottomrule
\end{tabular}
\caption{Main token- and sentence-level statistics for the 900-instance candidate pool and CLAW-4L. Raw token ratios $r_t$ denote non-English-to-English token ratios, $r_t^{norm}$ denotes ratios normalized by language-specific inflation factors estimated from parallel data, and sentence ratios $r_s$ denote non-English-to-English sentence ratios computed from cleaned biographies using a multilingual neural sentence segmentation model (SaT).}
\label{tab:claw4-full-vs-sampled}
\end{table*}

\subsection{Draft Claim Generation for CLAW-4L-CX}
\label{app:claim_extract_claw_4l}

Starting from CLAW-4L, we extract claims from both the English and non-English biographies using GPT-5.1. 
The extracted claims are served only as sentence selection signals for CLAW-4L-CX and drafts for human revisions.



In addition to producing a decontextualized English claim, the prompt is also designed to output a lightweight structured representation for each extracted claim. Concretely, each claim is parsed into the following fields:
\[
\begin{aligned}
(&\texttt{subject},\texttt{predicate},\texttt{object},\texttt{time},\\
 &\texttt{location},\texttt{reason},\texttt{manner},\texttt{hedge})
\end{aligned}
\]
This structured output (infobox) serves as an auxiliary representation for downstream benchmark construction. In particular, it supports 
structured claim editing during human annotation, and controlled relation construction in CLAW-4L-RC such as contradiction and asymmetric enrichment.

\noindent An output example is shown below:
\begin{tcblisting}{
    enhanced,
    width=\linewidth,
    colback=gray!4,
    colframe=gray!50,
    boxrule=0.4pt,
    arc=1.5pt,
    left=4pt,
    right=4pt,
    top=4pt,
    bottom=4pt,
    listing only,
    breakable,
    listing options={
        breaklines=true,
        breakatwhitespace=false,
        basicstyle=\ttfamily\small
    }
}
{
  "subject": "Maya Angelou",
  "predicate": "traveled",
  "object": "California",
  "time": "1940",
  "location": "California",
  "reason": "to join her mother",
  "manner": "by train",
  "hedge": "according to some accounts",
  "claim": "According to some accounts, Maya Angelou traveled by train to California in 1940 to join her mother."
}
\end{tcblisting}

\begin{table*}[t]
\centering
\small
\begin{tabular}{llrrrrr}
\toprule
\textbf{Lang} & \textbf{Country} & \textbf{N} & \textbf{Total Claims} & \textbf{Mean Claims} & \textbf{Median Claims} & \textbf{Mean Sentence Coverage} \\
\midrule
\multirow{4}{*}{EN}
& \textbf{Overall}    & \textbf{300} & \textbf{9261}  & \textbf{30.87} & \textbf{20.00} & \textbf{0.966} \\
& Azerbaijan          & 100          & 3359           & 33.59          & 24.00          & 0.964 \\
& China               & 100          & 2926           & 29.26          & 19.50          & 0.967 \\
& France              & 100          & 2976           & 29.76          & 17.00          & 0.968 \\
\midrule
\multirow{4}{*}{non-EN}
& \textbf{Overall}    & \textbf{300} & \textbf{26798} & \textbf{89.33} & \textbf{43.00} & \textbf{0.960} \\
& Azerbaijan          & 100          & 9379           & 93.79          & 45.50          & 0.947 \\
& China               & 100          & 7237           & 72.37          & 37.00          & 0.979 \\
& France              & 100          & 10182          & 101.82         & 42.00          & 0.955 \\
\bottomrule
\end{tabular}
\caption{Claim extraction statistics for CLAW-4L. \textbf{Mean Sentence Coverage} averages, over biography instances, the fraction of source sentences that yield at least one extracted claim.}
\label{tab:claw4_sent_claim_stats}
\end{table*}

\begin{figure}
    \centering
    \includegraphics[width=1.0\linewidth]{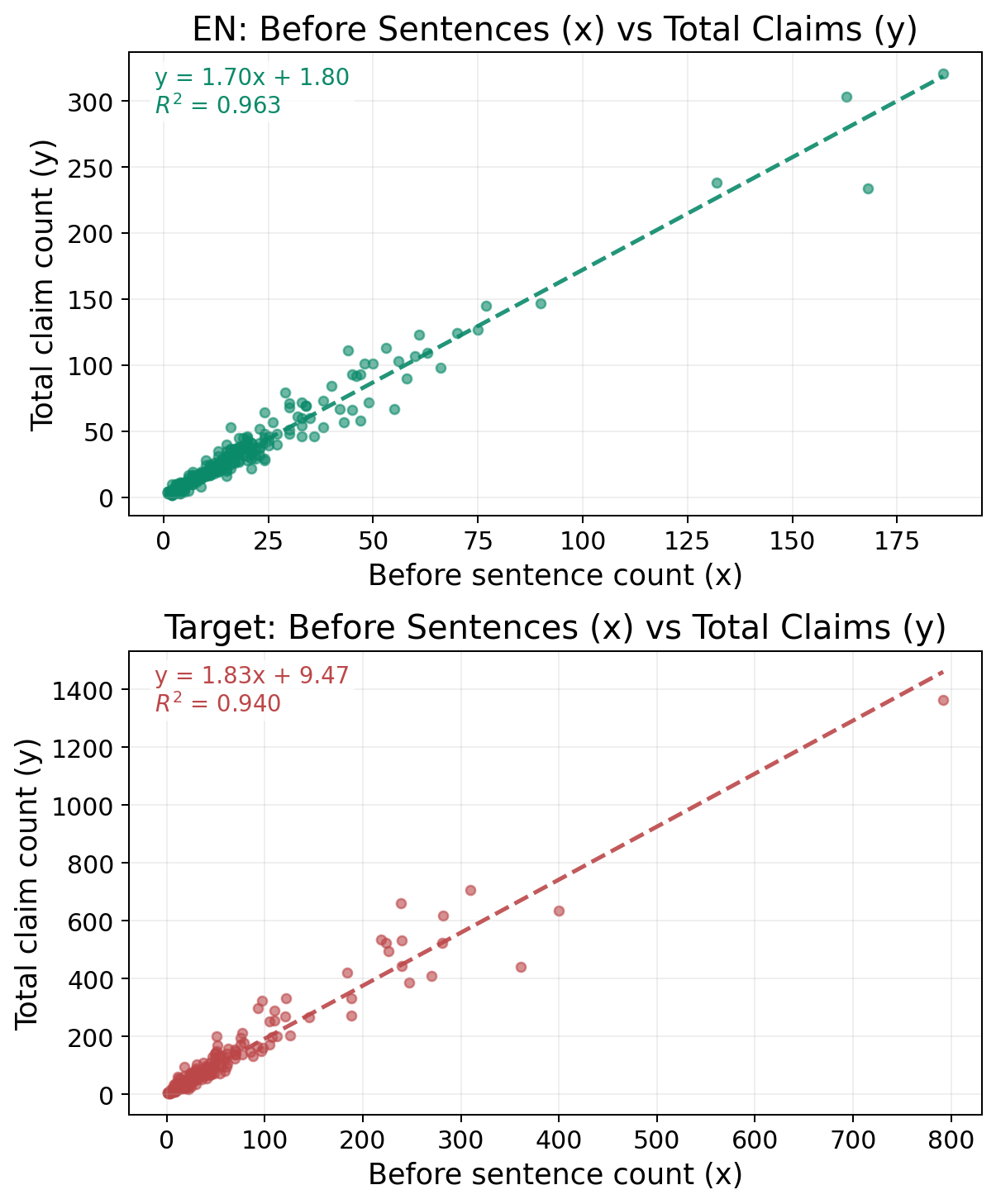}
    \caption{Relationship between source sentence count (x) and extracted claim count (y) per biography in CLAW-4L (EN on top, non-English language on bottom). Each point denotes one biography; dashed lines show least-squares fits (EN: $y = 1.70x + 1.88$, $R^2 = 0.961$; non-English: $y = 1.82x + 9.16$, $R^2 = 0.941$). Claim extraction generally yields more claims than source sentences, with mean claims-per-sentence ratios of 1.89 (EN) and 2.14 (non-English).}
    \label{fig:claim_vs_sent}
\end{figure}

\begin{figure}
    \centering
    \includegraphics[width=1.0\linewidth]{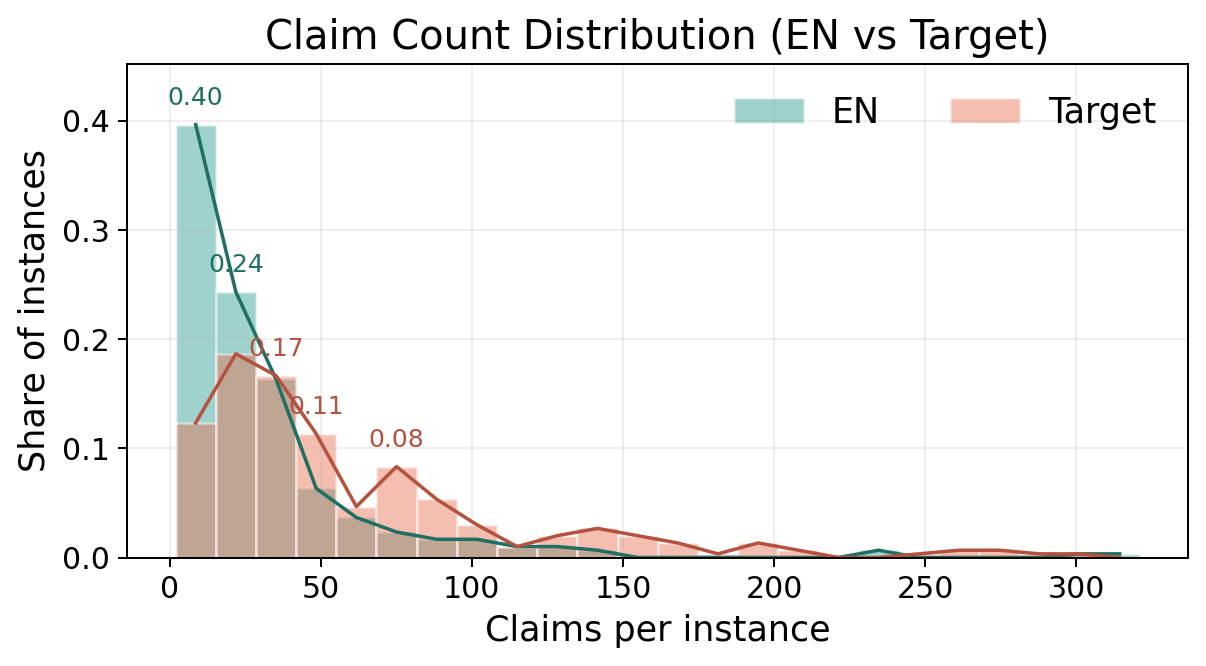}
    \caption{Distribution of extracted claim counts per biography in CLAW-4L, comparing English (EN) and non-English biographies in a single overlaid histogram (shared bins; y-axis shows instance share). EN is more concentrated at lower claim counts, while non-English biographies exhibit a heavier right tail, indicating more high-claim instances.}
    \label{fig:sampled_300_claim_count_distribution}
\end{figure}

\begin{figure}
    \centering
    \includegraphics[width=1.0\linewidth]{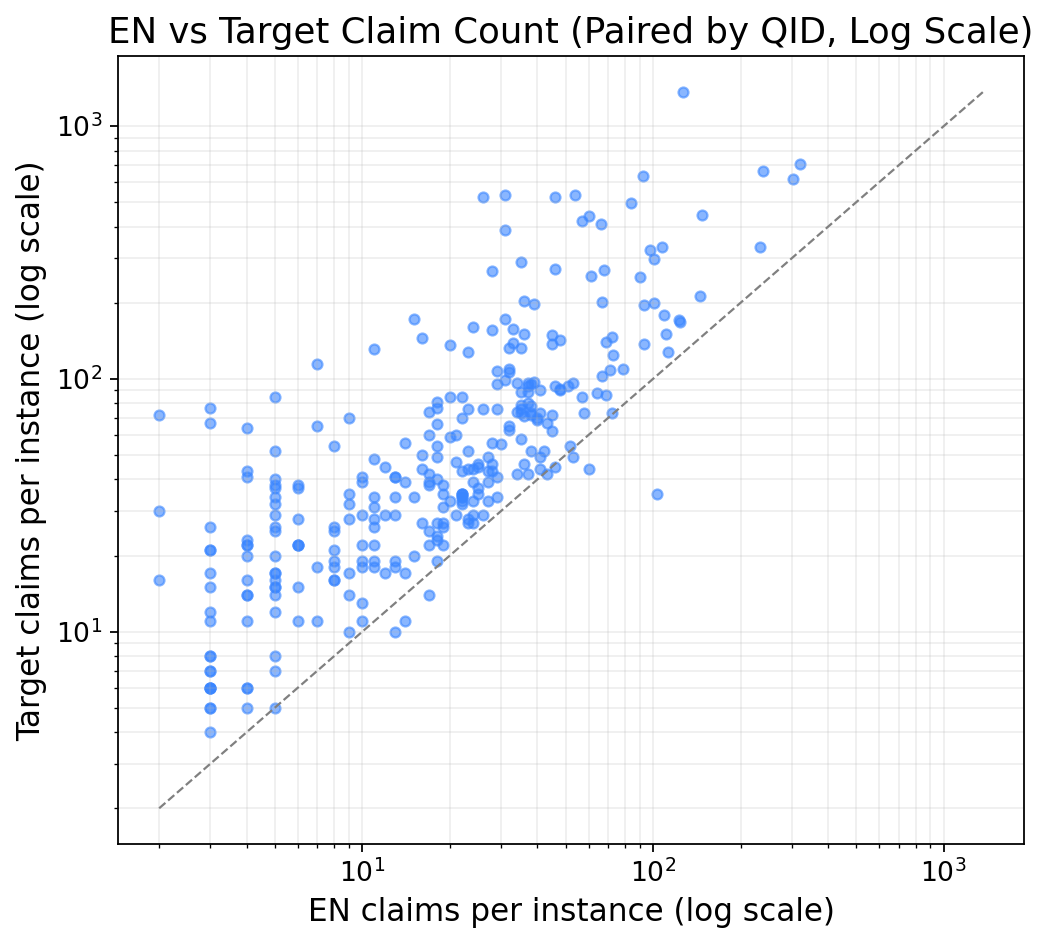}
    \caption{Log-scale comparison of extracted claim counts per biography between English and non-English texts in CLAW-4L. Each point is one biography; the diagonal dashed line indicates parity $(y = x)$. Most points lie above the diagonal, showing that the non-English biographies generally yield more extracted claims than their English counterparts, while the log scale improves visibility across the long-tailed count range.}
    \label{fig:en_non-English_claim_scatter}
\end{figure}

\subsection{Sentence Selection for CLAW-4L-CX}
\label{subsec:claw-4L-CX-strategy}
After claim extraction, we selected one sentence from each biography in CLAW-4L to construct the sentence-claim benchmark CLAW-4L-CX. Our selection procedure was designed to balance two goals: preserving the natural distribution of sentence-level claim complexity in CLAW-4L, while enriching the final selected set with sentences whose extracted claims are more faithful, informative, and easier to validate.

\paragraph{Candidate sentences and claim-count buckets.}
For each biography, we treated every source sentence that yielded at least one extracted claim as a candidate annotation sentence. Let \(S_b\) denote the set of candidate sentences for biography \(b\), and let \(C(s)\) denote the set of extracted claims associated with sentence \(s \in S_b\). We first grouped candidate sentences into four claim-count buckets according to \(|C(s)|\):
\[
\begin{aligned}
\mathcal{B}_1 &= \{s : |C(s)| = 1\},\\
\mathcal{B}_2 &= \{s : |C(s)| = 2\},\\
\mathcal{B}_3 &= \{s : |C(s)| = 3\},\\
\mathcal{B}_4 &= \{s : |C(s)| \ge 4\}.
\end{aligned}
\]
This four-way partition provided a simple but effective proxy for sentence-claim complexity, separating simple, medium, and claim-dense sentences while avoiding overly sparse high-count buckets.

\paragraph{Selection features.}
For each candidate sentence \(s\), we computed three sentence-level quality signals.

\textbf{(1) Claim coverage.}
To estimate how well the extracted claims collectively cover the factual content of the source sentence, we concatenated all claims in \(C(s)\) into a single text string \(\mathrm{concat}(C(s))\), and computed multilingual sentence similarity with LaBSE:
\begin{equation}
\begin{aligned}
\mathrm{cov}(s)
&=
\cos\Bigl(
\phi_{\mathrm{LaBSE}}(s), \\
&\qquad
\phi_{\mathrm{LaBSE}}\bigl(\mathrm{concat}(C(s))\bigr)
\Bigr).
\end{aligned}
\end{equation}
where \(\phi_{\mathrm{LaBSE}}(\cdot)\) denotes the sentence embedding function. Higher values indicate that the extracted claims more faithfully capture the factual core of the source sentence.

\textbf{(2) Fact density.}
We next estimated how much structured factual information a sentence contributes relative to its length. For a claim \(c \in C(s)\), let
\[
\begin{aligned}
F(c)=\{&
\texttt{subject},\texttt{predicate},\texttt{object},\texttt{time},\\
&\texttt{location},\texttt{reason},\texttt{manner},\texttt{hedge}
\}.
\end{aligned}
\]
We define an indicator \(\mathbf{1}[f(c)]\) that equals 1 if field \(f\) is non-empty, and 0 otherwise. For the \texttt{hedge} field, we set \(\mathbf{1}[\texttt{hedge}(c)] = 1\) only when the hedge value is not \texttt{"No"}. We exclude the free-form text fields \texttt{claim} and \texttt{source\_sent} from this computation. Sentence-level fact density is then defined as
\[
\mathrm{dens}(s)
=
\frac{
\sum_{c \in C(s)}
\sum_{f \in F(c)}
\mathbf{1}[f(c)]
}{
|s|_{\mathrm{tok}}
},
\]
where \(|s|_{\mathrm{tok}}\) is the token length of the source sentence. Higher values indicate that more structured factual information is packed into the sentence.

\textbf{(3) Rewrite cost.}
Finally, we quantified how much rewriting was required to transform the source sentence into its decontextualized claims. For each claim \(c \in C(s)\), we computed a normalized token-level edit distance
\[
\mathrm{ned}(s,c)
=
\frac{
\mathrm{ED}(s,c)
}{
\max(|s|_{\mathrm{tok}}, |c|_{\mathrm{tok}})
},
\]
where \(\mathrm{ED}(\cdot,\cdot)\) denotes token-level Levenshtein edit distance. We then define sentence-level rewrite cost as the mean normalized edit distance across claims:
\[
\mathrm{rew}(s)
=
\frac{1}{|C(s)|}
\sum_{c \in C(s)} \mathrm{ned}(s,c).
\]
Lower values indicate that the sentence can be converted into decontextualized claims with less rewriting, while larger values suggest stronger contextual dependence or more substantial reformulation.

\paragraph{Instance-level rank aggregation.}
Because these three features are measured on different scales, we did not combine them directly. Instead, for each biography \(b\), we converted the candidate sentences in \(S_b\) into within-biography normalized rank scores in \([0,1]\). Let \(n_b = |S_b|\). For each metric \(m \in \{\mathrm{cov}, \mathrm{dens}, \mathrm{rew}\}\), we first assign each candidate sentence \(s \in S_b\) a rank \(\mathrm{rank}_m(s) \in \{1,\dots,n_b\}\), where larger ranks always indicate better candidates. Thus, for claim coverage and fact density, higher raw values correspond to larger ranks, whereas for rewrite cost, lower raw values correspond to larger ranks. We then normalize these ranks as
\[
R_m(s)=
\frac{\mathrm{rank}_m(s)-1}{n_b-1}.
\]
This yields \(R_m(s) \in [0,1]\), with 1 assigned to the best-ranked sentence and 0 to the worst-ranked sentence for that metric within the same biography. If a biography yields only one candidate sentence, we assign that sentence a normalized rank score of 1 for all metrics by definition. Let these normalized rank scores be denoted
\[
R_{\mathrm{cov}}(s),\quad
R_{\mathrm{dens}}(s),\quad
R_{\mathrm{rew}}(s).
\]
We then defined the overall sentence quality score as an unweighted rank aggregation:
\[
\mathrm{score}(s)
=
\frac{
R_{\mathrm{cov}}(s)
+
R_{\mathrm{dens}}(s)
+
R_{\mathrm{rew}}(s)
}{3}.
\]
We used equal weights because no sentence-level development annotations were available for tuning feature importance, and the three signals were intended to play complementary roles: coverage captures faithfulness, density captures informational richness, and rewrite cost captures annotation difficulty.

\paragraph{Global complexity balancing.}
Selecting the highest-scoring sentence independently for each biography would tend to over-favor simpler one-claim sentences. To preserve realism while maintaining sufficient complexity diversity, we therefore imposed global constraints on the distribution of claim-count buckets in the final selected set.

We began by computing the empirical distribution of candidate sentences across the four buckets \(\mathcal{B}_1,\ldots,\mathcal{B}_4\) within each language side (English and non-English). We then applied a hard-coded bucket balance strategy: all claim buckets (1-, 2-, 3-, 4+-claim) share the same proportion (0.25) in the final CLAW-4L-CX. This design ensures that the final sentence set contained enough medium- and high-complexity examples to evaluate extraction quality beyond the easiest cases.

\paragraph{Constrained final selection.}
Let \(x_{b,s} \in \{0,1\}\) indicate whether candidate sentence \(s \in S_b\) is selected for biography \(b\). We selected the final sentence set by maximizing the total sentence quality subject to two types of constraints:
\[
\max_{x}
\sum_b \sum_{s \in S_b} \mathrm{score}(s)\,x_{b,s}
\]
subject to
\[
\sum_{s \in S_b} x_{b,s} = 1
\qquad \forall b,
\]
so that exactly one sentence is selected from each biography, and
\[
\sum_b \sum_{s \in S_b:\, g(s)=k} x_{b,s} = T_k
\qquad \forall k \in \{1,2,3,4\},
\]
where \(g(s)\) is the claim-count bucket of sentence \(s\), and \(T_k\) is the target number of selected sentences assigned to bucket \(k\) after the mild rebalance step described above.

In practice, this constrained selection procedure yielded a sentence set that remained close to the natural complexity distribution of the sampled biography pool while avoiding over-representation of trivial one-claim sentences. Sentence selection was performed independently for English and non-English biographies; consequently, the final EN and non-English sentences in CLAW-4L-CX are not sentence-aligned within biography pairs. The resulting 300 English and 300 non-English-language sentences constitute CLAW-4L-CX, our sentence-level human annotation benchmark.

\subsection{CLAW-4L-CX Statistics}

Table~\ref{tab:claw4_sent_bucket_distribution} summarizes the claim-count distribution of the final selected sentences in CLAW-4L-CX. By construction, the benchmark contains exactly one selected sentence from each sampled biography, yielding 300 English and 300 non-English-language sentences overall. Importantly, the final sentence set does not collapse onto only the simplest one-claim cases. On the English side, the selected sentences contain, on average, 2.63 claims, with 25 instances per country across the 1-, 2-, 3-, and 4+-claim buckets. On the non-English side, the corresponding mean is 2.72 claims, with 25 instances per claim bucket and per country. This confirms that the final benchmark preserves a meaningful range of sentence-level complexity, including both relatively simple and claim-dense cases.

Table~\ref{tab:claw4_sent_rank_shift} shows that this diversity is preserved without sacrificing sentence quality. Compared with CLAW-4L, the final selected sentences in CLAW-4L-CX exhibit consistent positive shifts across all three normalized ranking dimensions. On the English side, the selected set improves by +0.4107 in coverage rank, +0.2095 in fact-density rank, and +0.1965 in rewrite-rank, yielding an overall score shift of +0.2722. On the non-English side, the corresponding gains are +0.3285, +0.2221, and +0.4030, with an overall improvement of +0.3179. These shifts show that the selected sentences are not merely representative of the claim-count distribution; they are also systematically better candidates for human annotation. Specifically, higher coverage indicates that the extracted claims more faithfully capture the factual content of the source sentence, higher fact density indicates that the sentence contains richer, structured factual content, and higher rewrite-rank (equivalently, lower rewrite cost) indicates that the selected sentences require less aggressive reformulation into decontextualized claims, making them easier to verify and refine during annotation.

Taken together, these results suggest that CLAW-4L-CX achieves the intended trade-off in dataset construction. It preserves sentence-level diversity by covering a broad range of claim-count buckets while simultaneously enriching for higher-quality claim-bearing sentences. This makes it a suitable sentence-claim benchmark for evaluating claim extraction quality against human annotation.

\subsection{CLAW-4L-CX Annotators and Interface}
\label{app:claw4-sent-annotation}

Annotators were volunteer researchers recruited through the authors' academic and professional networks and were not financially compensated. Participation was voluntary, and annotators were informed that the annotations would be used for research.

\begin{figure*}
    \centering
    \includegraphics[width=1.0\linewidth]{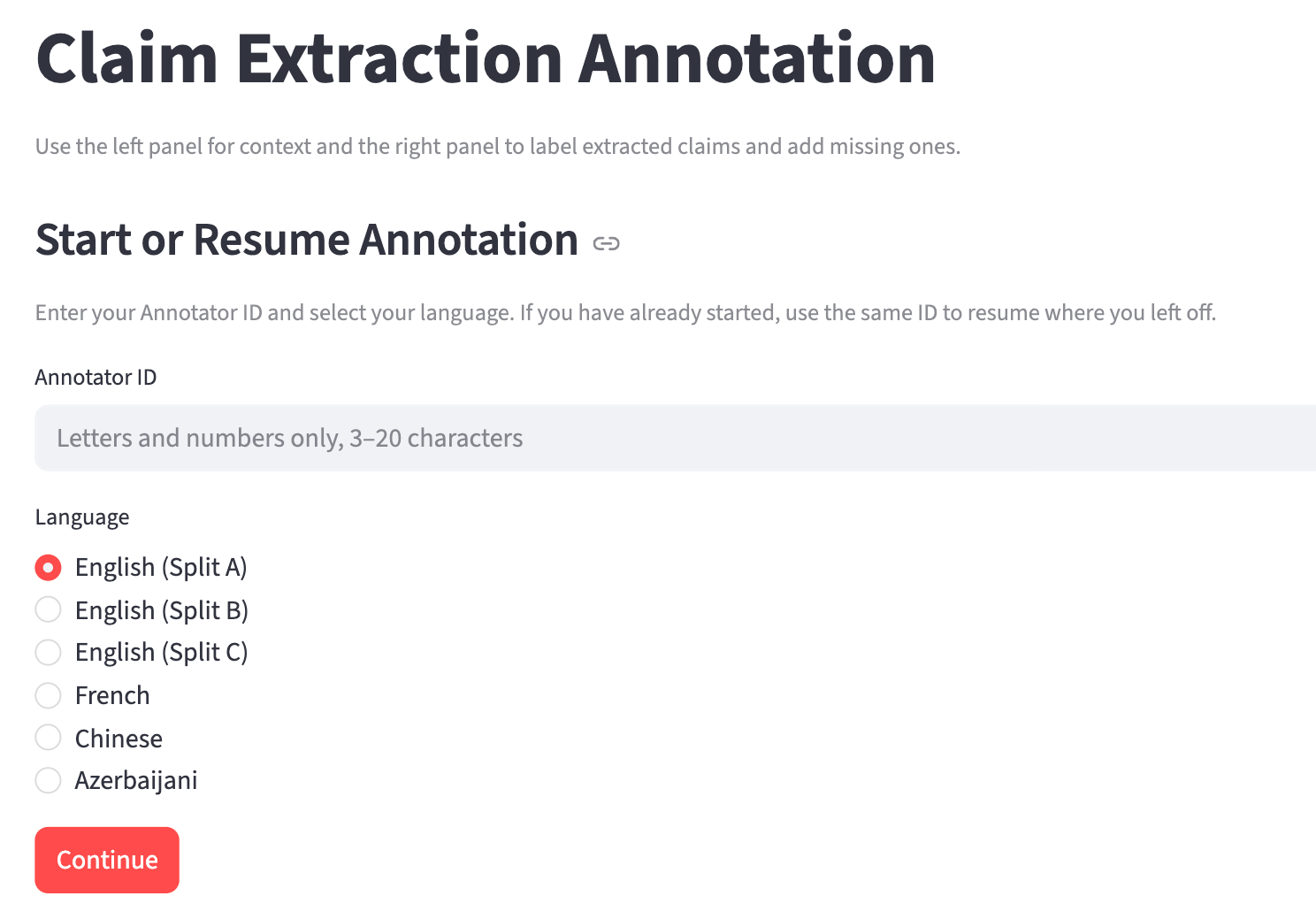}
    \caption{
    Human annotation interface for CLAW-4L-CX (1).
    }
    \label{fig:interface_1}
\end{figure*}

\begin{figure*}
    \centering
    \includegraphics[width=1.0\linewidth]{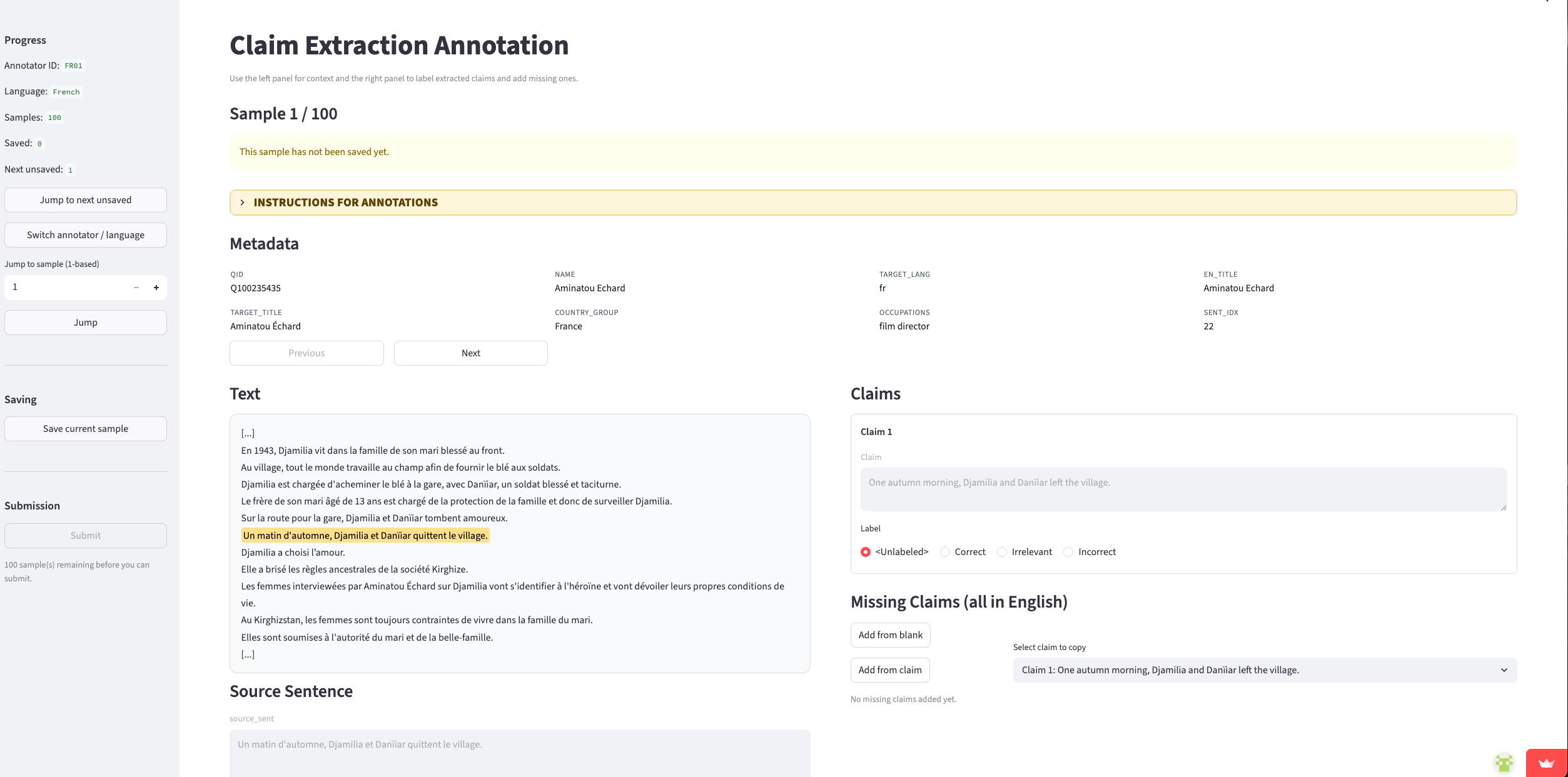}
    \caption{
    Human annotation interface for CLAW-4L-CX (2).
    }
    \label{fig:interface_2}
\end{figure*}

We provide our annotation interfaces in Figure~\ref{fig:interface_1} and Figure~\ref{fig:interface_2}.

\subsection{Post-hoc Inter-Annotator Agreement for CLAW-4L-CX}
\label{app:sent_iaa}

After selecting the claim-pair classifier described in Section~\ref{sec:claim_classifier}, we use it for a post-hoc agreement analysis of the \textsc{CLAW-4L-CX} annotations. For each language, we compare the reference claim sets produced by different annotators for the same sentence. Given two annotator claim sets \(\mathcal{C}^{(a)}\) and \(\mathcal{C}^{(b)}\), we score all cross-set claim pairs with the classifier. For each criterion \(k \in \{\textit{aligned}, \textit{exact-aligned}\}\), let \(E_k^{ab}\) be the set of claim pairs satisfying that criterion, where \textit{aligned} counts exact and partial alignments and \textit{exact-aligned} counts only exact matches. We compute existence-based set precision and recall:
\[
\begin{aligned}
P_k^{ab}
&=
\frac{|\{i : \exists j,\ (i,j)\in E_k^{ab}\}|}
{|\mathcal{C}^{(a)}|},\\
R_k^{ab}
&=
\frac{|\{j : \exists i,\ (i,j)\in E_k^{ab}\}|}
{|\mathcal{C}^{(b)}|}.
\end{aligned}
\]
We then compute \(F_{1,k}^{ab}=2P_k^{ab}R_k^{ab}/(P_k^{ab}+R_k^{ab})\). For each language, we report the average precision, recall, and F1 over all annotator pairs for all multiply annotated sentences in that language. The results are shown in Table~\ref{tab:iaa-by-language}.

\begin{table}[!htbp]
\centering
\small
\begin{tabular}{lcccccc}
\toprule
\multirow{2}{*}{Lang} & \multicolumn{3}{c}{Aligned} & \multicolumn{3}{c}{Exact-Aligned} \\
\cmidrule(lr){2-4} \cmidrule(lr){5-7}
& P & R & F1 & P & R & F1 \\
\midrule
AZ & 91.4 & 91.0 & 91.1 & 74.5 & 73.9 & 73.9 \\
EN & 98.9 & 100.0 & 99.3 & 81.2 & 80.7 & 80.3 \\
FR & 98.1 & 98.5 & 98.0 & 87.6 & 85.8 & 85.7 \\
ZH & 98.6 & 98.7 & 98.6 & 81.7 & 82.9 & 81.9 \\
\bottomrule
\end{tabular}
\caption{Post-hoc inter-annotator agreement for \textsc{CLAW-4L-CX} reference claim annotations. We compare claim sets produced by different annotators for the same sentence using the selected claim-pair classifier. A claim is counted as matched if it has at least one aligned or exact-aligned counterpart in the other annotator's set. Scores are reported as percentages.}
\label{tab:iaa-by-language}
\end{table}

\begin{table*}[t]
\centering
\small
\begin{tabular}{llrrrrrr}
\toprule
\textbf{Lang} & \textbf{Country} & \textbf{N} & \textbf{Mean \(n_{\text{claims}}\)} & \textbf{B1} & \textbf{B2} & \textbf{B3} & \textbf{B4} \\
\midrule
\multirow{4}{*}{EN}
& Azerbaijan & 100 & 2.67 & 25 & 25 & 25 & 25 \\
& China      & 100 & 2.61 & 25 & 25 & 25 & 25  \\
& France     & 100 & 2.60 & 25 & 25 & 25 & 25 \\
\cmidrule(lr){2-8}
& \textbf{Global} & \textbf{300} & \textbf{2.63} & \textbf{75} & \textbf{75} & \textbf{75} & \textbf{75} \\
\midrule
\multirow{4}{*}{non-EN}
& Azerbaijan & 100 & 2.68 & 25 & 25 & 25 & 25  \\
& China      & 100 & 2.84 & 25 & 25 & 25 & 25 \\
& France     & 100 & 2.63 & 25 & 25 & 25 & 25  \\
\cmidrule(lr){2-8}
& \textbf{Global} & \textbf{300} & \textbf{2.72} & \textbf{75} & \textbf{75} & \textbf{75} & \textbf{75} \\
\bottomrule
\end{tabular}
\caption{Claim-count distribution of the final selected sentences in CLAW-4L-CX. Within each language side, rows report country-level statistics followed by a global summary. Here, B1, B2, B3, and B4 denote sentences yielding 1, 2, 3, and 4 or more extracted claims, respectively.}
\label{tab:claw4_sent_bucket_distribution}
\end{table*}

\begin{table*}[t]
\centering
\small
\setlength{\tabcolsep}{6pt}
\renewcommand{\arraystretch}{1.10}
\begin{tabular}{llccccc}
\toprule
\textbf{Lang} & \textbf{Country} & \textbf{\(N_{\text{before}} \rightarrow N_{\text{after}}\)} & \textbf{\(\Delta R_{\mathrm{cov}} \uparrow\)} & \textbf{\(\Delta R_{\mathrm{dens}} \uparrow\)} & \textbf{\(\Delta R_{\mathrm{rew}} \uparrow\)} & \textbf{\(\Delta \mathrm{Score} \uparrow\)} \\
\midrule
\multirow{4}{*}{EN}
& Azerbaijan & 1837 \(\rightarrow\) 100 & 0.4004 & 0.1347 & 0.2813 & 0.2721 \\
& China      & 1502 \(\rightarrow\) 100 & 0.4127 & 0.2447 & 0.1714 & 0.2763 \\
& France     & 1523 \(\rightarrow\) 100 & 0.4190 & 0.2490 & 0.1367 & 0.2682 \\
\cmidrule(lr){2-7}
& \textbf{Global} & \textbf{4862 \(\rightarrow\) 300} & \textbf{0.4107} & \textbf{0.2095} & \textbf{0.1965} & \textbf{0.2722} \\
\midrule
\multirow{4}{*}{non-EN}
& Azerbaijan & 5141 \(\rightarrow\) 100 & 0.3177 & 0.2201 & 0.4190 & 0.3189 \\
& China      & 2635 \(\rightarrow\) 100 & 0.2522 & 0.2316 & 0.3657 & 0.2832 \\
& France     & 4511 \(\rightarrow\) 100 & 0.4154 & 0.2144 & 0.4241 & 0.3513 \\
\cmidrule(lr){2-7}
& \textbf{Global} & \textbf{12287 \(\rightarrow\) 300} & \textbf{0.3285} & \textbf{0.2221} & \textbf{0.4030} & \textbf{0.3179} \\
\bottomrule
\end{tabular}
\caption{Shifts in normalized rank-based sentence selection statistics from CLAW-4L to the final selected sentences in CLAW-4L-CX. Here, \(N_{\text{before}}\) and \(N_{\text{after}}\) denote the numbers of candidate and selected sentences, respectively. We report \(\Delta = \text{after} - \text{before}\) for normalized coverage rank, density rank, rewrite-rank, and the final aggregated score, where larger positive values indicate a better trade-off between informativeness and editable difficulty of the selected set. 
}
\label{tab:claw4_sent_rank_shift}
\end{table*}

\subsection{Construction of CLAW-4L-RC}
\label{app:claw4l_rc}

To evaluate claim-pair relations across English and non-English biographies, we construct \textbf{CLAW-4L-RC}, a 600-instance benchmark built from the same GPT-5.1 claim extraction outputs used in CLAW-4L-CX. Each instance consists of one claim from the English biography, one claim from the paired non-English biography, their structured claim fields, and a manually validated relation label. Because both sides are normalized into English claim verbalizations at the extraction stage, the claim-pair matching stage operates over English claim pairs, even though the underlying evidence comes from cross-lingual biography pairs.

\paragraph{Aligned seed pairs.}
We begin by mining candidate aligned pairs from the full claim sets extracted from CLAW-4L. For each paired biography, we compute LaBSE cosine similarity between every English-side claim and every non-English-side claim using their claim texts, and use the highest-scoring pairs as alignment seeds. This screening step is used only to reduce the manual search space over the large Cartesian product of candidate claim pairs within each biography pair.

To ensure that the resulting positives are not dominated by trivial or repetitive templates, we apply diversity-aware filtering before final validation. In particular, we balance the final aligned subset across the three country groups, yielding 33 Chinese, 33 Azerbaijani, and 34 French pairs, and additionally constrain the distribution of claim lengths as well as over-represented life-event patterns such as birth- and death-related claims. The resulting 100 candidate pairs are then manually validated. During validation, annotators verify both the claim texts and their structured fields, so the final aligned positives are supported at both the textual and structured levels.

\paragraph{Relation schema.}
Each claim pair is labeled in two stages. First, we assign an alignment label from:
\begin{itemize}
    \item \textbf{Aligned}: the two claims describe the same factual aspect and can both be true;
    \item \textbf{Contradicted}: the two claims describe the same factual aspect but assert incompatible values or states;
    \item \textbf{Not Relevant}: the two claims describe different factual aspects, even if they concern the same person.
\end{itemize}

If and only if the pair is labeled \textit{Aligned}, we further assign an enrichment label from:
\begin{itemize}
    \item \textbf{A$=$B}: neither claim contains meaningful extra atomic factual content beyond the other;
    \item \textbf{A$>$B}: Claim A contains the content of Claim B and adds additional non-conflicting detail;
    \item \textbf{B$>$A}: Claim B contains the content of Claim A and adds additional non-conflicting detail;
    \item \textbf{A$\leftrightarrow$B}: the two claims are aligned and non-contradictory, but each contains non-conflicting factual detail missing from the other.
\end{itemize}

We provide Table~\ref{tab:claim_relation_example} to illustrate the examples of different claim pair relations.

\begin{table*}[!ht]
\centering
\small
\renewcommand{\arraystretch}{1.3} 
\setlength{\tabcolsep}{5.4pt}
\begin{tabular}{l c p{6.2cm} p{6.2cm}}
\hline
\textbf{Category} & \textbf{Rel.} & \textbf{Claim A} & \textbf{Claim B} \\
\hline
\multirow{4}{*}{Alig.} 
& $A{=}B$ & Libby Lee Ha-yun attended Ying Wa Girls' School \textit{during her high school years}. & Libby Lee attended Ying Wa Girls' School \textit{during her senior secondary school years}. \\
\cmidrule(lr){3-4}
& $A{>}B$ & Libby Lee Ha-yun attended Ying Wa Girls' School \textcolor{orange}{in Mid-Levels} during her high school years. & Libby Lee attended Ying Wa Girls' School during her senior secondary school years. \\
\cmidrule(lr){3-4}
& $B{>}A$ & Libby Lee attended Ying Wa Girls' School during her senior secondary school years. & Libby Lee Ha-yun attended Ying Wa Girls' School \textcolor{orange}{in Mid-Levels} during her high school years. \\
\cmidrule(lr){3-4}
& $A{\leftrightarrow}B$ & Libby Lee Ha-yun attended Ying Wa Girls' School \textcolor{orange}{in Mid-Levels during her high school years}. & Libby Lee attended Ying Wa Girls' School \textcolor{orange}{in 1991}. \\
\hline
Contr. & $A{\perp}B$ & Libby Lee Ha-yun attended \textcolor{red}{Ying Wa} Girls' School during her high school years. & Libby Lee attended \textcolor{red}{Diocesan} Girls' School during her senior secondary school years. \\
\hline
Not Rel. & $A{\dashv}B$ & \textbf{Libby Lee Ha-yun} attended Ying Wa Girls' School during her high school years. & In 1957, \textbf{Huang Yifan} died in London, England, at the age of 61. \\
\hline
\end{tabular}
\caption{Examples of claim relations. \textcolor{orange}{Orange text} highlights the additional details that appear only on one side of the claim. \textcolor{red}{Red text} highlights the contradictory key words between two claims.}
\label{tab:claim_relation_example}
\end{table*}

\paragraph{Construction of non-positive relations.}
Starting from the manually validated aligned pairs, we construct the remaining relation types through controlled structured editing and GPT-5.1 rewriting.

For \textbf{Contradicted} pairs, we preserve the same underlying claim anchor while changing one field value to an incompatible alternative. Candidate replacement values are drawn from field-specific value pools built from the extracted claims in the sampled biography pool. In practice, we focus on fields whose modification naturally induces contradiction, such as \textit{time}, \textit{object}, and, in some cases, \textit{location}.

For \textbf{Not Relevant} pairs, we pair claims that do not describe the same factual aspect. These include both easier negatives, obtained from claims drawn from different biography instances, and harder negatives, obtained from claims within the same biography that share the subject but refer to different events, roles, works, or attributes. We additionally create a subset of such examples through controlled field perturbation followed by GPT-5.1 rewriting.

For aligned pairs with non-symmetric detail, we construct one-sided and mutual enrichment cases by varying the amount of non-conflicting information on the two sides. We create \textbf{A$>$B} examples by deleting minor fields such as \textit{reason}, \textit{manner}, or \textit{hedge}, or by coarsening values in fields such as \textit{time} and \textit{location} and obtain \textbf{B$>$A} by swapping claim A and claim B in \textbf{A$>$B}. We create \textbf{A$\leftrightarrow$B} examples by ensuring that each side retains at least one non-conflicting atomic detail that is absent from the other. After structured modification, GPT-5.1 is used to rewrite the affected claim text so that it remains fluent and consistent with the edited fields.

\paragraph{Human validation and fallback correction.}
All non-positive pairs produced by rule-based perturbation and GPT-5.1 rewriting are manually validated against their expected labels. Annotators check both the claim texts and the structured fields. When a generated example does not satisfy the intended relation label, we do not rely on automatic re-classification; instead, we manually correct or rewrite the pair until it matches the intended relation. As a result, all final instances in CLAW-4L-RC are human-validated at the pair level.

\subsection{CLAW-4L-RC Statistics}

Table~\ref{tab:claw4l-rc-label-dist} reports the label distribution and semantic similarity scores of claim pairs in \textsc{CLAW-4L-RC}. The dataset contains 600 claim pairs in total: 400 aligned pairs, 100 contradicted pairs, and 100 not-relevant pairs. The aligned category is further divided into four balanced sub-labels, each containing 100 instances.

To provide an auxiliary validation of the constructed pairs, we compute sentence-level semantic similarity using two English embedding models: \textit{BGE-Large-En-v1.5}\footnote{\url{https://huggingface.co/BAAI/bge-large-en-v1.5}} and \textit{All-MPNet-Base-v2}\footnote{\url{https://huggingface.co/sentence-transformers/all-mpnet-base-v2}}. For the \textit{Aligned} row, we report the macro-average over the four aligned sub-labels. For the \textit{Total} row, we report a label-size-weighted average over all categories, so that the aggregate score reflects the full 600-pair distribution rather than over-emphasizing the aligned category.

The similarity patterns are consistent across the two embedding models. Exact aligned pairs obtain the highest scores, followed by one-directional partial alignment and then mutual partial alignment. Contradicted pairs remain relatively close in embedding space, whereas not-relevant pairs receive substantially lower similarity scores. This pattern is consistent with our annotation design: aligned pairs should be semantically close, not-relevant pairs should be clearly distant, and contradicted pairs may still share substantial topical or lexical content despite containing incompatible information. These results provide additional evidence for the quality and internal consistency of the manually constructed \textsc{CLAW-4L-RC} dataset.

\begin{table}[!htbp]
\centering
\small
\setlength{\tabcolsep}{4pt}
\begin{tabular}{lrrr}
\toprule
\textbf{Label} & \textbf{\#} & \textbf{BGE} & \textbf{MPNet} \\
\midrule
Aligned -- Exact ($A=B$) & 100 & 96.87 & 96.04 \\
Aligned -- Partial ($A > B$) & 100 & 91.15 & 89.79 \\
Aligned -- Partial ($B > A$) & 100 & 91.15 & 89.79 \\
Aligned -- Partial ($A \leftrightarrow B$) & 100 & 85.47 & 84.27 \\
\cmidrule(lr){1-4}
\textit{Aligned} & 400 & 91.16 & 89.97 \\
\midrule
Contradicted ($A \perp B$) & 100 & 88.47 & 88.47 \\
Not Relevant ($A \dashv B$) & 100 & 51.31 & 35.86 \\
\midrule
\textbf{Weighted Total} & \textbf{600} & 84.07 & 80.70 \\
\bottomrule
\end{tabular}
\caption{Label distribution and semantic similarity in \textsc{CLAW-4L-RC}. BGE and MPNet denote cosine similarity scores computed with BGE-Large-En-v1.5 and All-MPNet-Base-v2 sentence embeddings, respectively; scores are reported on a 0-100 scale.}
\label{tab:claw4l-rc-label-dist}
\end{table}

\section{LLM Classifier Selection on CLAW-4L-RC}
\label{app:llm_classifier}

\subsection{Models}
To select a reliable claim-pair relation classifier, we evaluate a range of instruction-tuned LLMs on \textsc{CLAW-4L-RC} and additionally compare against two strong NLI baselines, \textit{DeBERTa-v3-large-mnli}\footnote{\url{https://huggingface.co/MoritzLaurer/DeBERTa-v3-large-mnli-fever-anli-ling-wanli}} and \textit{MiniCheck-7B}\footnote{\url{https://huggingface.co/bespokelabs/Bespoke-MiniCheck-7B}}. We focus primarily on medium-sized open-source models in the 3B--14B range, balancing inference efficiency with expected verification accuracy. We also include GPT-5.1 as a strong closed-source reference model, representing an upper-bound classifier in our setting.

Specifically, we evaluate the following model families:
\begin{itemize}
    \item \textbf{Qwen3}: Qwen3-4B, Qwen3-8B, and Qwen3-14B;
    \item \textbf{Qwen3.5}: Qwen3.5-4B and Qwen3.5-9B;
    \item \textbf{Llama-3}: Llama-3.2-3B-IT and Llama-3.1-8B-IT;
    \item \textbf{Ministral-3}: Ministral-3-3B-IT, Ministral-3-8B-IT, and Ministral-3-14B-IT;
    \item \textbf{Gemma-3}: Gemma-3-4B-IT and Gemma-3-12B-IT;
    \item \textbf{GPT-5}: GPT-5.1.
\end{itemize}

For each model, we evaluate two input settings: \textit{w/o} infobox, where the classifier only receives the two claim texts, and \textit{w/} infobox, where the corresponding structured infobox (parsed from the claims by GPT-5.1; a structured infobox example is provided in Appendix~\ref{app:claim_extract_claw_4l}) is additionally provided as contextual evidence.

\subsection{Evaluation Setting and Aggregate Metrics}

\textsc{CLAW-4L-RC} is a claim-pair relation classification benchmark with six fine-grained labels. Four labels correspond to aligned relations: exact alignment ($A{=}B$), one-sided enrichment from $A$ to $B$ ($A{>}B$), one-sided enrichment from $B$ to $A$ ($B{>}A$), and mutual non-conflicting enrichment ($A{\leftrightarrow}B$). The remaining two labels capture non-aligned relations: contradiction ($A{\perp}B$) and non-relevance ($A{\dashv}B$).

We report accuracy for each fine-grained label and further compute three aggregate metrics. Let
\[
s_{=}, s_{A>B}, s_{B>A}, s_{\leftrightarrow}, s_{\perp}, s_{\dashv}
\]
denote the per-label accuracies for $A{=}B$, $A{>}B$, $B{>}A$, $A{\leftrightarrow}B$, $A{\perp}B$, and $A{\dashv}B$, respectively.

First, \textbf{Align-FG} measures fine-grained discrimination within the aligned category:
\[
\text{Align-FG}
=
\frac{
s_{=} + s_{A>B} + s_{B>A} + s_{\leftrightarrow}
}{4}.
\]
This metric evaluates whether a classifier can distinguish different types of aligned and partially aligned relations, rather than merely recognizing that two claims are broadly related.

Second, \textbf{ARC} measures top-level relation classification accuracy by first averaging the four aligned sub-labels into a single aligned score, and then macro-averaging over the three top-level relation types:
\[
\text{ARC}
=
\frac{
\text{Align-FG} + s_{\perp} + s_{\dashv}
}{3}.
\]
This metric reflects performance on the main relation judgment task: aligned, contradicted, or not relevant, while avoiding over-weighting the aligned category, which contains four fine-grained sub-labels.

Finally, \textbf{Overall} is the macro-average over all six fine-grained labels:
\[
\text{Overall}
=
\frac{
\!s_{=} + s_{A>B} + s_{B>A} + s_{\leftrightarrow} + s_{\perp} + s_{\dashv}\!
}{6}.
\]
Unlike ARC, this metric treats all fine-grained labels equally and therefore reflects the classifier's full six-way classification performance.

\subsection{NLI Baseline Adaptation}

We additionally compare against two strong NLI-style baselines, \textit{MiniCheck-7B} and \textit{DeBERTa-v3-large-mnli}. Since these models are not designed for our six-way claim relation setting, we adapt their outputs into the \textsc{CLAW-4L-RC} label space as far as possible.

\paragraph{MiniCheck-7B.}
MiniCheck-7B outputs a binary label in $\{\textit{supported}, \textit{unsupported}\}$. We therefore apply it in both directions for a claim pair $(A,B)$. If $A$ supports $B$ and $B$ supports $A$, we map the pair to exact alignment ($A{=}B$). If $A$ supports $B$ but $B$ does not support $A$, we map it to one-sided enrichment ($A{>}B$); symmetrically, if $B$ supports $A$ but $A$ does not support $B$, we map it to ($B{>}A$). However, MiniCheck-7B cannot distinguish mutual partial alignment ($A{\leftrightarrow}B$), contradiction ($A{\perp}B$), or non-relevance ($A{\dashv}B$), since all of these collapse into the same \textit{unsupported} outcome. Accordingly, we only report the label-wise accuracies that can be directly computed.

\paragraph{DeBERTa-v3-large-mnli.}
DeBERTa-v3-large-mnli outputs a three-way NLI label in $\{\textit{entailment}, \textit{neutral}, \textit{contradiction}\}$. As with MiniCheck, we evaluate each pair bidirectionally. If $A$ entails $B$ and $B$ entails $A$, we map the pair to exact alignment ($A{=}B$). If $A$ entails $B$ and $B$ is neutral with respect to $A$, we map it to $A{>}B$; symmetrically, if $B$ entails $A$ and $A$ is neutral with respect to $B$, we map it to $B{>}A$. If either direction yields contradiction, we map the pair to contradiction ($A{\perp}B$). However, this model still cannot distinguish mutual partial alignment ($A{\leftrightarrow}B$) from non-relevance ($A{\dashv}B$), since both may appear as neutral in one or both directions. For this reason, we report only the directly computable label-wise accuracies and omit aggregate scores.

\subsection{Results}

\begin{figure*}
    \centering
    \includegraphics[width=1.0\linewidth]{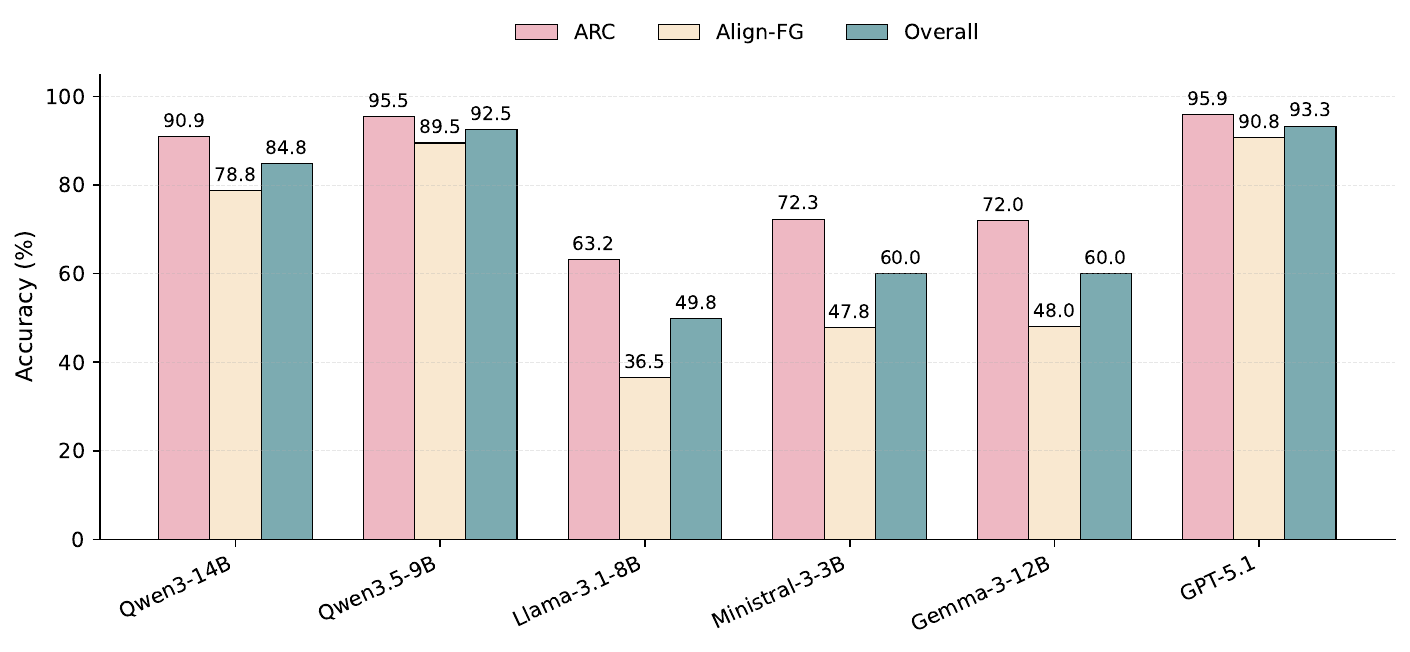}
    \caption{
    Comparison of the best-performing classifier from each model family on \textsc{CLAW-4L-RC} across three aggregate metrics (ARC, Align-FG, Overall). GPT-5.1 achieves the strongest performance, while Qwen3.5-9B is the closest open-source model across all metrics.
    }
    \label{fig:llm_classifier_res}
\end{figure*}

We provide the full results in Table~\ref{tab:infobox-ablation} and summarize them in Figure~\ref{fig:llm_classifier_res}. We highlight several key observations.

\paragraph{NLI baselines as partial classifiers.}
The NLI baselines provide a useful reference point but are inherently limited by label mismatch. MiniCheck-7B can only recover exact and one-sided support relations, while DeBERTa-v3-large-mnli additionally captures contradiction, but still cannot distinguish mutual partial alignment from non-relevance. Notably, DeBERTa-v3-large-mnli performs strongly on contradiction ($95.0$) and exact alignment ($90.0$), showing that supervised NLI training transfers well to coarse semantic consistency judgments. However, neither NLI baseline can model the finer enrichment distinctions central to \textsc{CLAW-4L-RC}, especially the mutual partial alignment case, which limits their usefulness as full classifiers in our setting.

\paragraph{Strong open-source classifier.}
Qwen3.5-9B emerges as the strongest open-source classifier, consistently achieving performance close to GPT-5.1 across all aggregate metrics. In particular, the gap between Qwen3.5-9B and GPT-5.1 is small (e.g., 95.5 vs. 95.9 in ARC), while both models substantially outperform other open-source families.

\paragraph{Fine-grained alignment is more challenging.}
We observe a clear positive correlation between ARC and Align-FG, indicating that models with strong top-level relation classification also perform well on fine-grained alignment distinctions. However, Align-FG scores are consistently lower than ARC across all models, confirming that fine-grained alignment (e.g., distinguishing $A{>}B$, $B{>}A$, and $A{\leftrightarrow}B$) is a more challenging task. Among these, the mutual enrichment case ($A{\leftrightarrow}B$) is typically the hardest, suggesting that modeling bidirectional, non-conflicting information requires more precise semantic reasoning.

\paragraph{Limited benefit of structured infoboxes.}
Providing GPT-5.1-parsed infoboxes does not consistently improve performance. For stronger models (e.g., Qwen3-14B, Qwen3.5-9B, GPT-5.1), infobox inputs often slightly degrade performance, suggesting that additional structured context may introduce noise or redundancy. In contrast, smaller models occasionally benefit from infobox inputs, indicating that external structure can partially compensate for weaker internal representations. Overall, these results suggest that the effectiveness of infobox augmentation depends on model capacity and input fidelity.

\paragraph{Scaling trends within model families.}
Within most model families, larger models tend to achieve better performance, consistent with expected scaling behavior. This trend holds for Qwen3 and Qwen3.5, where performance improves monotonically with model size. An exception is observed in the Ministral-3 family, where the 3B model outperforms larger variants.

These findings justify our choice of \textbf{Qwen3.5-9B} as the default open-source classifier in our experiments.

\begin{table*}[!hbpt]
\centering
\small
\setlength{\tabcolsep}{3.8pt}
\renewcommand{\arraystretch}{1.15}
\begin{tabular}{ll cccc cc ccc}
\toprule
\multirow{2}{*}{\textbf{Model}} 
& \multirow{2}{*}{\textbf{Info.}} 
& \multicolumn{4}{c}{\textbf{Aligned}} 
& \multicolumn{1}{c}{\textbf{Contr.}} 
& \multicolumn{1}{c}{\textbf{Not Rel.}} 
& \multicolumn{3}{c}{\textbf{Aggregate}} \\
\cmidrule(lr){3-6}
\cmidrule(lr){7-7}
\cmidrule(lr){8-8}
\cmidrule(lr){9-11}
& & $A{=}B$ & $A{>}B$ & $B{>}A$ & $A{\leftrightarrow}B$ 
& $A{\perp}B$ & $A{\dashv}B$ 
& \textbf{ARC} & \textbf{Align-FG} & \textbf{Overall} \\
\midrule

\multicolumn{11}{l}{\textit{\textbf{NLI}}} \\
\cmidrule(lr){1-11}
\multirow{1}{*}{MiniCheck-7B} 
& w/o & 85.0 & 82.0 & 82.0 & \xmark & \xmark & \xmark & \xmark & \xmark & \xmark \\
\cmidrule(lr){2-11}
\multirow{1}{*}{DeBERTa-v3-large} 
& w/o & 90.0 & 86.0 & 86.0 & \xmark & 95.0 & \xmark & \xmark & \xmark & \xmark \\

\midrule

\multicolumn{11}{l}{\textit{\textbf{Qwen3}}} \\
\cmidrule(lr){1-11}
\multirow{2}{*}{Qwen3-4B}
& w/o & 52.0 & 83.0 & 61.0 & 22.0 & 63.0 & 91.0 & 69.5 & 54.5 & 62.0 \\
& w/  & 37.0 & 81.0 & 75.0 & 37.0 & 66.0 & 97.0 & 73.5 & 57.5 & 65.5 \\
\cmidrule(lr){2-11}
 \multirow{2}{*}{Qwen3-8B}
& w/o & 71.0 & 79.0 & \textbf{95.0} & 27.0 & 83.0 & 98.0 & 83.0 & 68.0 &
75.5 \\
& w/  & 72.0 & 79.0 & 92.0 & 45.0 & 77.0 & \underline{99.0} & 82.7 & 72.0
& 77.3 \\
\cmidrule(lr){2-11}
\multirow{2}{*}{Qwen3-14B} 
& w/o & \underline{95.0} & 72.0 & 77.0 & 71.0 & 95.0 & \underline{99.0} & 90.9 & 78.8 & 84.8 \\
& w/  & 85.0 & 61.0 & 71.0 & 75.0 & 90.0 & \textbf{100.0} & 87.7 & 73.0 & 80.3 \\

\midrule
\multicolumn{11}{l}{\textit{\textbf{Qwen3.5}}} \\
\cmidrule(lr){1-11}
\multirow{2}{*}{Qwen3.5-4B} 
& w/o & 71.0 & 72.0 & \underline{93.0} & \textbf{88.0} & 93.0 & 90.0 & 88.0 & 81.0 & 84.5 \\
& w/  & 60.0 & 73.0 & 73.0 & \underline{85.0} & \underline{97.0} & 93.0 & 87.6 & 72.8 & 80.2 \\
\cmidrule(lr){2-11}
\multirow{2}{*}{Qwen3.5-9B} 
& w/o & 94.0 & \underline{88.0} & 91.0 & \underline{85.0} & \textbf{99.0} & 98.0 & \underline{95.5} & \underline{89.5} & \underline{92.5} \\
& w/  & 93.0 & \textbf{90.0} & 89.0 & 84.0 & \underline{97.0} & \textbf{100.0} & 95.3 & 89.0 & 92.2 \\

\midrule
\multicolumn{11}{l}{\textit{\textbf{Llama-3}}} \\
\cmidrule(lr){1-11}
\multirow{2}{*}{Llama-3.2-3B-IT} 
& w/o & 43.0 & 29.0 & 9.0 & 60.0 & 41.0 & 45.0 & 40.4 & 35.3 & 37.8 \\
& w/  & 29.0 & 11.0 & 5.0 & 81.0 & 7.0 & 33.0 & 23.8 & 31.5 & 27.7 \\
\cmidrule(lr){2-11}
\multirow{2}{*}{Llama-3.1-8B-IT} 
& w/o & 51.0 & 74.0 & 20.0 & 1.0 & 61.0 & 92.0 & 63.2 & 36.5 & 49.8 \\
& w/  & 39.0 & 69.0 & 18.0 & 2.0 & 75.0 & 97.0 & 68.0 & 32.0 & 50.0 \\

\midrule
\multicolumn{11}{l}{\textit{\textbf{Ministral-3}}} \\
\cmidrule(lr){1-11}
\multirow{2}{*}{Ministral-3-3B-IT} 
& w/o & 88.0 & 24.0 & 56.0 & 23.0 & 70.0 & \underline{99.0} & 72.3 & 47.8 & 60.0 \\
& w/  & 81.0 & 12.0 & 45.0 & 8.0 & 79.0 & 98.0 & 71.2 & 36.5 & 53.8 \\
\cmidrule(lr){2-11}
\multirow{2}{*}{Ministral-3-8B-IT} 
& w/o & 76.0 & 44.0 & 36.0 & 14.0 & 92.0 & 94.0 & 76.2 & 42.5 & 59.3 \\
& w/  & 57.0 & 39.0 & 31.0 & 12.0 & 87.0 & 94.0 & 71.9 & 34.8 & 53.3 \\
\cmidrule(lr){2-11}
\multirow{2}{*}{Ministral-3-14B-IT} 
& w/o & 75.0 & 2.0 & 1.0 & 1.0 & 88.0 & 89.0 & 65.6 & 19.8 & 42.7 \\
& w/  & 71.0 & 0.0 & 1.0 & 1.0 & 92.0 & 85.0 & 65.1 & 18.2 & 41.7 \\

\midrule
\multicolumn{11}{l}{\textit{\textbf{Gemma-3}}} \\
\cmidrule(lr){1-11}
\multirow{2}{*}{Gemma-3-4B-IT} 
& w/o & 87.0 & 37.0 & 38.0 & 0.0 & 40.0 & 71.0 & 50.5 & 40.5 & 45.5 \\
& w/  & 67.0 & 54.0 & 33.0 & 3.0 & 42.0 & 77.0 & 52.8 & 39.3 & 46.0 \\
\cmidrule(lr){2-11}
\multirow{2}{*}{Gemma-3-12B-IT} 
& w/o & 87.0 & 78.0 & 15.0 & 12.0 & 71.0 & 97.0 & 72.0 & 48.0 & 60.0 \\
& w/  & 79.0 & 73.0 & 22.0 & 10.0 & 70.0 & 97.0 & 71.0 & 46.0 & 58.5 \\

\midrule
\multicolumn{11}{l}{\textit{\textbf{GPT-5}}} \\
\cmidrule(lr){1-11}
\multirow{2}{*}{GPT-5.1} 
& w/o & \textbf{97.0} & \underline{88.0} & \underline{93.0} & \underline{85.0} & \underline{97.0} & \textbf{100.0} & \textbf{95.9} & \textbf{90.8} & \textbf{93.3} \\
& w/  & \underline{95.0} & 86.0 & 87.0 & 78.0 & 95.0 & \textbf{100.0} & 93.8 & 86.5 & 90.2 \\

\bottomrule
\end{tabular}
\caption{
Claim-pair relation classification results on \textsc{CLAW-4L-RC} across various classifiers including NLI models and LLMs.
\textit{w/o} denotes using only claim pairs as input, while \textit{w/} additionally provides GPT-5.1-parsed infoboxes. 
All metrics are accuracy (\%). 
\textbf{ARC} is the macro-average over the three top-level relation types, where the Aligned score is first averaged over its four sub-labels. 
\textbf{Align-FG} measures fine-grained accuracy within the Aligned category (average over its four sub-labels). 
\textbf{Overall} is the macro-average over all six fine-grained labels. 
\textbf{Best} and \underline{second-best} results in each column are marked in bold and underlined, respectively.
Entries marked with \xmark\ are not directly computable because the corresponding classifier does not provide enough label resolution to distinguish that relation type.}
\label{tab:infobox-ablation}
\end{table*}

\section{Cross-lingual Claim Extraction Methods}
\label{app:claim_extraction_method}

This section summarizes the English claim extraction frameworks adapted in our experiments and clarifies how they are unified for cross-lingual evaluation on \textsc{CLAW-4L-CX}. Rather than comparing the original backbone models or source tasks of prior work, we focus on the functional design choices that matter for our setting and then describe how each method is adapted to the cross-lingual sentence-to-claims setting.

\subsection{Core Operations in Claim Extraction}
\paragraph{Decomposition.}
Decomposition breaks a sentence or passage into smaller factual units so that each claim can be checked independently. For example, the sentence "Marie Curie won the Nobel Prize in Physics in 1903 and the Nobel Prize in Chemistry in 1911" can be decomposed into two claims, one for each award event.

\paragraph{Decontextualization.}
Decontextualization rewrites a claim so that it can be understood and verified without relying on the surrounding context. For example, "In 1903, she won the Nobel Prize in Physics" becomes "Marie Curie won the Nobel Prize in Physics in 1903."

\paragraph{Verifiable Selection.}
Verifiable selection determines whether a sentence contains objectively checkable factual content and filters out non-verifiable material such as opinions, speculation, or advice. For instance, "Solar power could transform the future of humanity" is not directly verifiable, whereas "Solar power provided 3.9\% of global electricity in 2021" is.

\paragraph{Disambiguation.}
Disambiguation handles cases with multiple plausible interpretations, typically by resolving ambiguity from context or abstaining when reliable resolution is not possible. For example, in "John met Paul after he won the award", the pronoun "he" may refer to either John or Paul. In such cases, a system should avoid producing a confident factual claim unless the referent can be resolved reliably.

\begin{table*}[!t]
\centering
\small
\setlength{\tabcolsep}{4.5pt}
\begin{tabular}{lcccccl}
\toprule
\textbf{Method (Year)} & \textbf{Uses Context} & \textbf{Decomp.} & \textbf{Decontext.} & \textbf{Verifiable Sel.} & \textbf{Disamb.} & \textbf{Execution} \\
\midrule
FactScore (2023)    & \xmark & \cmark & \xmark & \xmark & \xmark & Joint \\
DnDScore (2025)     & \cmark & \cmark & \cmark & \xmark & \cmark & In-prompt Seq. \\
VeriScore (2024)    & \cmark & \cmark & \cmark & \cmark & \xmark & Joint \\
FactCheck-GPT (2024) & \cmark & \cmark & \cmark & \cmark & \xmark & Multi-turn \\
Claimify (2025)     & \cmark & \cmark & \cmark & \cmark & \cmark & Multi-turn \\
\bottomrule
\end{tabular}
\caption{Functional comparison of the English claim extraction frameworks adapted in our experiments. A check mark (\cmark) indicates that the corresponding operation is explicitly specified in the method prompt or pipeline. \textit{Uses Context} indicates whether the method explicitly incorporates the surrounding sentence or paragraph context beyond the focal sentence. \textit{Execution} describes how the operations are organized: \textit{Joint} uses a single prompt without staged intermediate outputs, \textit{In-prompt Seq.} performs the steps sequentially within one prompt, and \textit{Multi-turn} splits them across multiple prompting rounds.}
\label{tab:xclaim_methods}
\end{table*}

\subsection{Adapted English Claim Extraction Frameworks}
Table~\ref{tab:xclaim_methods} summarizes the main functional differences among the compared extraction frameworks. We adapt five representative English claim extraction methods to our cross-lingual setting: FactScore, FactCheck-GPT, DnDScore, VeriScore, and Claimify. We prefix the adapted variants with \textit{X-}. In all cases, we modify the original prompts so that the system accepts either an English or a non-English-language sentence as input and always outputs atomic factual claims in English. The goal is to map both English and non-English biographies into a shared English claim space for downstream alignment and enrichment. We provide the X-Claimify prompts in Appendix~\ref{app:prompt}; prompts and runnable scripts for all adapted frameworks are included in the accompanying repository.

\paragraph{X-FactScore.}
X-FactScore is the simplest variant and focuses on direct decomposition. Given a sentence, it extracts one or more factual claims in English without explicitly requiring decontextualization, verifiable filtering, or ambiguity handling beyond what the LLM performs implicitly.

\paragraph{X-DnDScore.}
X-DnDScore extends direct decomposition by explicitly decontextualizing and handling ambiguity. In a single prompt, it first decomposes the input sentence into subclaims and then rewrites each subclaim into a stand-alone English proposition using only the provided paragraph context. Unlike simpler decomposition-based methods, it explicitly asks the model to identify ambiguities and avoid confident extraction when core references, entities, or relations cannot be resolved reliably.

\paragraph{X-VeriScore.}
X-VeriScore jointly performs decomposition, decontextualization, and verifiable selection in a single prompt. It extracts as many fine-grained verifiable facts as possible from the non-English sentence, while using the surrounding context only to resolve pronouns and definite descriptions. Each output claim must be self-contained, independently checkable, and written in English; if the sentence contains no verifiable factual content, the model is instructed to abstain by returning \textit{No verifiable claim.}

\paragraph{X-FactCheck-GPT.}
X-FactCheck-GPT uses a two-stage extraction design. It first decomposes the input sentence into standalone English atomic facts, using the provided context to resolve references and decontextualize the output. It then applies a separate checkworthy classification step to determine whether the sentence contains factual content worth keeping. This staged design is intended to reduce noisy or weakly factual outputs while preserving decontextualized atomic claims.

\paragraph{X-Claimify.}
X-Claimify is the most structured pipeline among the compared methods. It combines verifiable selection, decontextualization, disambiguation, and decomposition in a multi-turn prompting process. In our cross-lingual setting, it is adapted so that claims extracted from non-English biographies are directly normalized into English, making them comparable to claims extracted from the English side.


\subsection{Evaluation on \textsc{CLAW-4L-CX}}
\label{subsec:eval_claim_extract_quality}

We evaluate claim extraction quality on \textsc{CLAW-4L-CX}, where each sentence is independently annotated by three human references. Because generated claims may differ from the references in wording and in how contextual information is made explicit, exact string matching is inappropriate. We therefore perform classifier-based matching at the claim-set level using the claim-pair relation classifier (Qwen3.5-9B) described in Section~\ref{sec:claim_classifier}.

Let $\hat{\mathcal{C}}=\{\hat{c}_1,\ldots,\hat{c}_m\}$ denote the set of generated claims for a sentence, and let $\mathcal{C}^{(r)}=\{c^{(r)}_1,\ldots,c^{(r)}_n\}$ denote the reference claims provided by annotator $r \in \{1,2,3\}$. For each reference set $\mathcal{C}^{(r)}$, we apply the classifier to all generated--reference claim pairs and obtain relation labels.

We use degree-normalized claim-level matching rather than one-to-one maximum matching. For a matching criterion $k \in \{\text{align}, \text{exact}\}$, let $E_k^{(r)}$ denote the generated--reference claim pairs whose classifier-predicted relation satisfies criterion $k$. The \textit{align} criterion accepts both exact and partial alignment, while the \textit{exact} criterion accepts only exact alignment. For each generated claim \(i\) and reference claim \(j\), define
\[
\begin{aligned}
d_i^k
&= |\{j' : (i,j') \in E_k^{(r)}\}|,\\
\bar{d}_j^k
&= |\{i' : (i',j) \in E_k^{(r)}\}|.
\end{aligned}
\]
We then assign claim-level precision and recall credits:
\[
\begin{aligned}
p_i^k
&=
\begin{cases}
1/d_i^k, & d_i^k > 0,\\
0, & d_i^k = 0,
\end{cases}\\
r_j^k
&=
\begin{cases}
1/\bar{d}_j^k, & \bar{d}_j^k > 0,\\
0, & \bar{d}_j^k = 0.
\end{cases}
\end{aligned}
\]
This fan-out normalization reduces the precision credit of overly broad generated claims that align with multiple reference claims, and reduces the recall credit when multiple generated claims redundantly align with the same reference claim.

For each criterion $k$, we define precision, recall, and F1 as
\[
P_k^{(r)}
=
\frac{1}{|\hat{\mathcal{C}}|}
\sum_{i=1}^{m} p_i^k,
\qquad
R_k^{(r)}
=
\frac{1}{|\mathcal{C}^{(r)}|}
\sum_{j=1}^{n} r_j^k,
\]
\[
F_k^{(r)} =
\frac{2 P_k^{(r)} R_k^{(r)}}
{P_k^{(r)} + R_k^{(r)}}.
\]
Aligned-F1 measures whether the extractor recovers the relevant factual content even under partial alignment, whereas Exact-Aligned-F1 more directly reflects factual specificity and decomposition quality.

For each sentence, we compute these scores separately against each of the three human reference sets. We then select the reference set that yields the highest exact-alignment score,
\[
r^* = \arg\max_{r \in \{1,2,3\}} F_{\text{exact}}^{(r)},
\]
and report
\[
\begin{aligned}
\text{Aligned-F1} &= F_{\text{align}}^{(r^*)}, \\
\text{Exact-Aligned-F1} &= F_{\text{exact}}^{(r^*)} .
\end{aligned}
\]

Using F1 in both cases allows us to jointly capture precision and coverage. Intuitively, Aligned-F1 measures semantic content recovery, whereas Exact-Aligned-F1 more strongly rewards exact claim formulation.

\subsection{Two-stage Selection Protocol}

To separate the extraction-framework choice from the backbone choice, we adopt a two-stage evaluation protocol.

\paragraph{Stage 1: extraction framework selection.}
We first instantiate all compared extraction frameworks with GPT-5.1 and evaluate them on \textsc{CLAW-4L-CX}. Using a common strong backbone allows us to isolate differences in the extraction framework design rather than differences in the model capability. Based on this comparison, we select the best-performing extraction formulation.

\paragraph{Stage 2: open-source backbone selection.}
After fixing the best-performing extraction framework, we compare several open-source LLMs as practical backbones for cross-lingual claim extraction, including Qwen3.6-27B, Gemma-4-31B-it, and Mistral-3.2-24B-it. This second stage is motivated by cost, reproducibility, and scalability considerations. We use it to assess whether open-source models can approximate GPT-5.1 extraction quality, while keeping GPT-5.1 + X-Claimify as the reference extraction configuration for the main pipeline.

\subsection{Final Choice}

Table~\ref{tab:claim_extraction_alignment} shows that X-Claimify is the most reliable extraction framework on all language splits. It achieves the strongest Exact-Aligned-F1 in English, French, and Chinese, and also obtains the best or near-best Aligned-F1. This pattern is important for our setting because downstream alignment requires claims that both recover the source content and preserve a comparable event-level granularity.

The weaker performance of X-DnDScore is consistent with its operation order. X-DnDScore first decomposes a sentence into subclaims and then decontextualizes each subclaim separately. In biography text, this can copy the same contextual information into multiple claims, creating overlapping or redundant claim verbalizations. In contrast, methods that resolve context before final decomposition are better able to produce factually independent claims, which improves exact matching against human references. X-Claimify further combines verifiable selection, disambiguation, decontextualization, and decomposition in a staged pipeline, which makes it better suited to sentences with coreference, partial names, and event-level biographical details.

We therefore use GPT-5.1 + X-Claimify as the reference extraction configuration for constructing English and non-English-side claim sets and for identifying non-English enrichment claims in downstream enrichment experiments. Table~\ref{tab:claim_model_alignment} compares open-source backbones as lower-cost alternatives. Overall, GPT-5.1 remains the strongest and most consistent backbone across languages. An exception is the French split, where Gemma-4-31B-it slightly exceeds GPT-5.1 on Aligned-F1 (82.62 vs.\ 78.05), indicating stronger broad content recovery, but remains lower on Exact-Aligned-F1 (70.18 vs.\ 73.56), indicating weaker exact claim formulation. For pipeline consistency and stronger aggregate precision, we keep GPT-5.1 + X-Claimify as the reference extraction configuration for all languages.

\begin{table*}[!htbp]
\centering
\small
\setlength{\tabcolsep}{4.5pt}
\renewcommand{\arraystretch}{1.08}
\begin{tabular}{ll ccc ccc}
\toprule
\multirow{2}{*}{\textbf{Lang}}
& \multirow{2}{*}{\textbf{Method}}
& \multicolumn{3}{c}{\textbf{Aligned}}
& \multicolumn{3}{c}{\textbf{Exact-Aligned}} \\
\cmidrule(lr){3-5} \cmidrule(lr){6-8}
&
& \textbf{P} & \textbf{R} & \textbf{F1}
& \textbf{P} & \textbf{R} & \textbf{F1} \\
\midrule

\multirow{5}{*}{AZ}
& X-FactScore & \textbf{71.38} & 47.19 & 54.04 & 37.07 & 47.73 & 39.51 \\
& X-VeriScore & 68.79 & \underline{68.00} & \underline{66.30} & \underline{57.27} & \underline{56.31} & \underline{55.11} \\
& X-DnDScore & 64.72 & 36.87 & 44.76 & 19.09 & 32.95 & 22.87 \\
& X-FactCheck-GPT & 56.63 & 51.88 & 52.91 & 43.69 & 45.93 & 44.38 \\
& X-Claimify & \underline{69.92} & \textbf{72.93} & \textbf{70.52} & \textbf{67.77} & \textbf{66.83} & \textbf{66.80} \\
\midrule

\multirow{5}{*}{EN}
& X-FactScore & 76.25 & 44.81 & 52.89 & 23.63 & 38.21 & 27.83 \\
& X-VeriScore & \underline{77.51} & \underline{65.50} & \underline{68.82} & \underline{46.37} & \underline{49.48} & \underline{46.92} \\
& X-DnDScore & 69.28 & 32.98 & 41.91 & 9.98 & 17.77 & 11.61 \\
& X-FactCheck-GPT & 70.04 & 54.90 & 59.60 & 42.18 & 46.66 & 43.70 \\
& X-Claimify & \textbf{77.72} & \textbf{75.76} & \textbf{75.17} & \textbf{72.82} & \textbf{73.18} & \textbf{72.44} \\
\midrule

\multirow{5}{*}{FR}
& X-FactScore & \underline{77.93} & 49.54 & 57.89 & 31.12 & 44.67 & 35.23 \\
& X-VeriScore & 77.43 & \underline{66.76} & \underline{70.01} & \underline{52.03} & \underline{57.54} & \underline{52.91} \\
& X-DnDScore & 72.56 & 37.57 & 46.16 & 13.22 & 24.87 & 16.08 \\
& X-FactCheck-GPT & 70.71 & 58.10 & 62.13 & 48.75 & 52.25 & 49.87 \\
& X-Claimify & \textbf{82.48} & \textbf{76.98} & \textbf{78.05} & \textbf{73.84} & \textbf{74.30} & \textbf{73.56} \\
\midrule

\multirow{5}{*}{ZH}
& X-FactScore & 69.93 & 45.61 & 52.51 & 22.83 & 34.28 & 26.30 \\
& X-VeriScore & \textbf{76.70} & \underline{64.68} & \underline{68.12} & 41.99 & 45.63 & 42.94 \\
& X-DnDScore & 71.48 & 35.14 & 44.75 & 14.64 & 23.52 & 16.02 \\
& X-FactCheck-GPT & 63.86 & 50.84 & 54.70 & \underline{47.85} & \underline{53.53} & \underline{49.82} \\
& X-Claimify & \underline{76.08} & \textbf{69.32} & \textbf{71.12} & \textbf{70.52} & \textbf{74.93} & \textbf{71.91} \\

\bottomrule
\end{tabular}
\caption{Comparison of different claim extraction methods using GPT-5.1 against human reference claims on \textsc{CLAW-4L-CX}. Scores are percentages.}
\label{tab:claim_extraction_alignment}
\end{table*}

\begin{table*}[!htbp]
\centering
\small
\setlength{\tabcolsep}{4.0pt}
\renewcommand{\arraystretch}{1.08}
\begin{tabular}{ll ccc ccc c}
\toprule
\multirow{2}{*}{\textbf{Lang}}
& \multirow{2}{*}{\textbf{Method}}
& \multicolumn{3}{c}{\textbf{Aligned}}
& \multicolumn{3}{c}{\textbf{Exact-Aligned}}
& \multirow{2}{*}{\textbf{Avg-F1}} \\
\cmidrule(lr){3-5} \cmidrule(lr){6-8}
&
& \textbf{P} & \textbf{R} & \textbf{F1}
& \textbf{P} & \textbf{R} & \textbf{F1}
& \\
\midrule

\multirow{4}{*}{AZ}
& GPT-5.1 & \textbf{69.92} & 72.93 & \textbf{70.52} & \textbf{67.77} & \textbf{66.83} & \textbf{66.80} & \textbf{68.66} \\
& Mistral-Small-3.2-24B-IT & 64.00 & 70.29 & 64.90 & 59.59 & 52.42 & 54.35 & 59.62 \\
& Gemma-4-31B-IT & 66.25 & \textbf{78.45} & \underline{69.42} & 58.16 & 53.27 & 54.28 & 61.85 \\
& Qwen3.6-27B & \underline{67.30} & \underline{75.42} & 69.23 & \underline{63.90} & \underline{58.73} & \underline{59.88} & \underline{64.56} \\
\midrule

\multirow{4}{*}{EN}
& GPT-5.1 & \textbf{77.72} & \underline{75.76} & \underline{75.17} & \textbf{72.82} & \textbf{73.18} & \textbf{72.44} & \textbf{73.81}\\
& Mistral-Small-3.2-24B-IT & \underline{77.11} & 72.83 & 72.79 & 58.03 & 54.02 & 54.83 & 63.81 \\
& Gemma-4-31B-IT & 76.52 & \textbf{77.79} & \textbf{75.42} & \underline{64.61} & \underline{64.09} & \underline{63.83} & \underline{69.63} \\
& Qwen3.6-27B & 68.08 & 65.82 & 65.73 & 55.66 & 55.37 & 55.25 & 60.49 \\
\midrule

\multirow{4}{*}{FR}
& GPT-5.1 & 82.48 & 76.98 & 78.05 & \textbf{73.84} & \textbf{74.30} & \textbf{73.56} & \underline{75.81} \\
& Mistral-Small-3.2-24B-IT & 79.89 & 75.72 & 75.90 & 58.75 & 56.65 & 56.50 & 66.20 \\
& Gemma-4-31B-IT & \underline{82.88} & \textbf{84.84} & \textbf{82.62} & \underline{71.62} & \underline{69.60} & \underline{70.18} & \textbf{76.40} \\
& Qwen3.6-27B & \textbf{83.25} & \underline{81.26} & \underline{80.35} & 65.38 & 64.22 & 64.12 & 72.24 \\
\midrule

\multirow{4}{*}{ZH}
& GPT-5.1 & \textbf{76.08} & 69.32 & 71.12 & \textbf{70.52} & \textbf{74.93} & \textbf{71.91} & \textbf{71.52} \\
& Mistral-Small-3.2-24B-IT & \underline{74.87} & \textbf{73.63} & \textbf{72.33} & 46.32 & 45.78 & 45.74 & 59.04 \\
& Gemma-4-31B-IT & 73.15 & 69.64 & 70.46 & \underline{60.35} & \underline{61.47} & \underline{60.52} & \underline{65.49} \\
& Qwen3.6-27B & 74.06 & \underline{72.39} & \underline{71.99} & 56.70 & 57.21 & 56.31 & 64.15 \\

\bottomrule
\end{tabular}
\caption{Claim extraction quality of X-Claimify using different claim generation models against human reference claims on \textsc{CLAW-4L-CX}. Scores are percentages. Avg-F1 is the average of Aligned F1 and Exact-Aligned F1.}
\label{tab:claim_model_alignment}
\end{table*}

\section{Machine Translation Model Selection for Biography Enrichment}
\label{app:mt_selection_bio}

To support the \textit{Machine Translation and Generation Pipeline} setting in Section~\ref{subsec:generation_methods}, we compare five candidate non-English-to-English translation models: the three open-source generator LLMs used in our enrichment experiments (Qwen3.6-27B, Gemma-4-31B-it, and Mistral-3.2-24B-it), together with two multilingual machine translation models, M2M100-1.2B and NLLB-200-3.3B.

We evaluate these translation models on a small but challenging subset of \textsc{CLAW-4L}. Specifically, for each non-English language (French, Chinese, and Azerbaijani), we select the three biographies with the largest token counts, yielding 9 biographies in total. These long biographies are particularly informative for translation model selection because they better reflect the context length and information density challenges encountered in the full enrichment setting.

\subsection{Translation Protocol}

We translate each non-English biography into English with each of the five candidate models. As in the generation setting, we cannot always feed an entire long biography together with task instructions into the model at once, since increased context length substantially raises GPU memory requirements during inference. We therefore use the same incremental chunking strategy as in Section~\ref{subsec:generation_methods}. 

Since translation only requires the non-English biography as input, without the English biography, we use a larger per-chunk budget and cap the non-English-side input at 4,096 tokens. When a biography exceeds this limit, we split it incrementally using the same section-based procedure as in the generation setup, with newline-based and token-level fallback splitting for exceptionally long sections. The translated chunks are then concatenated in order to form the full English translation of the biography.

\subsection{Claim-based Evaluation of Translated Biographies}
\label{app:mt_eval_bio}

We evaluate translated biographies using a claim-based protocol rather than surface-form overlap. For each original non-English biography, we first extract non-English-to-English reference claims using X-Claimify instantiated with GPT-5.1. These claims serve as the semantic reference for the factual content expressed in the source biography. We then apply the same X-Claimify + GPT-5.1 extraction pipeline to each translated English biography produced by the candidate translation models.

Because biography-level claim sets can be large, direct all-pairs claim comparison is computationally expensive. We therefore adopt a coarse-to-fine matching strategy. First, for each translated claim, we retrieve candidate reference claims using \texttt{all-mpnet-base-v2}, keeping only pairs that satisfy a cosine similarity threshold of 0.7 and fall within the top-5 nearest neighbors. We use MPNet rather than BGE because Table~\ref{tab:claw4l-rc-label-dist} suggests that MPNet provides stronger separation between aligned and clearly unrelated claim pairs. The threshold of 0.7 is intentionally loose and is used only to prune highly irrelevant pairs before claim-level verification.

We then apply the degree-normalized claim-level scoring protocol described in Appendix~\ref{subsec:eval_claim_extract_quality} to the filtered candidate pairs. As in our extraction evaluation, we report both \textbf{Aligned-F1} and \textbf{Exact-Aligned-F1}. Aligned-F1 measures whether the translation preserves the relevant factual content even under slight differences in granularity, whereas Exact-Aligned-F1 more strictly evaluates whether the translated biography preserves claims in a form that matches the reference at the atomic level.

\begin{table*}[!htbp]
\centering
\small
\setlength{\tabcolsep}{4.6pt}
\renewcommand{\arraystretch}{1.15}
\begin{tabular}{ll ccc ccc}
\toprule
\multirow{2}{*}{\textbf{Lang}} & \multirow{2}{*}{\textbf{Model}} 
& \multicolumn{3}{c}{\textbf{Aligned}} 
& \multicolumn{3}{c}{\textbf{Exact-Aligned}} \\
\cmidrule(lr){3-5} \cmidrule(lr){6-8}
& & \textbf{P} & \textbf{R} & \textbf{F1} & \textbf{P} & \textbf{R} & \textbf{F1} \\
\midrule

\multirow{5}{*}{AZ $\rightarrow$ EN}
& Qwen3.6-27B&  92.27&  90.53&  91.39&  34.49&  32.92&  33.69\\
& \textbf{Gemma-4-31B-it}&  \textbf{92.66}&  \textbf{91.51}&  \textbf{92.08}&  \textbf{36.64}&  \textbf{33.70}&  \textbf{35.11}\\
& Mistral-3.2-24B-it&  91.08&  90.94&  91.01&  34.57&  32.23&  33.36\\
& M2M100-1.2B&  78.64&  64.86&  71.09&  27.28&  10.37&  15.03\\
& NLLB-200-3.3B&  88.08&  86.02&  87.04&  32.33&  29.89&  31.06\\
\midrule

\multirow{5}{*}{FR $\rightarrow$ EN}
& Qwen3.6-27B&  93.80&  90.32&  92.03&  45.55&  43.27&  44.38\\
& Gemma-4-31B-it&  \textbf{94.87}&  92.75&  \textbf{93.80}&  46.09&  44.89&  45.48\\
& \textbf{Mistral-3.2-24B-it}&  94.02&  \textbf{93.51}&  93.77&  \textbf{49.08}&  \textbf{48.89}&  \textbf{48.98}\\
& M2M100-1.2B&  85.69&  86.59&  86.14&  43.79&  41.27&  42.49\\
& NLLB-200-3.3B&  84.64&  84.69&  84.67&  43.30&  40.83&  42.03\\
\midrule

\multirow{5}{*}{ZH $\rightarrow$ EN}
& \textbf{Qwen3.6-27B}&  85.84&  \textbf{87.16}&  \textbf{86.50}&  33.36&  33.04&  33.20\\
& Gemma-4-31B-it&  \textbf{85.90}&  85.58&  85.74&  \textbf{36.57}&  \textbf{35.02}&  \textbf{35.78}\\
& Mistral-3.2-24B-it&  81.64&  80.11&  80.87&  30.34&  28.13&  29.20\\
& M2M100-1.2B&  66.44&  61.57&  63.91&  19.29&  16.32&  17.68\\
& NLLB-200-3.3B&  57.62&  52.77&  55.09&  16.24&  13.15&  14.53\\
\bottomrule
\end{tabular}
\caption{Claim-based evaluation of candidate non-English-to-English translation models for biography translation. We report precision (P), recall (R), and F1 under both coarse-grained \textbf{Aligned} matching and strict \textbf{Exact-Aligned} matching. Best values within each language pair are shown in bold.}
\label{tab:mt_selection_results}
\end{table*}

Results are shown in Table~\ref{tab:mt_selection_results}, and the final per-language choices are summarized in Table~\ref{tab:mt_selection_summary}. We make three main observations. First, \textbf{Exact-Aligned} scores are consistently much lower than coarse \textbf{Aligned} scores across all language pairs and models. This confirms that exact alignment is a substantially more demanding criterion for biography translation: while many translated claims remain semantically relevant to the source, exact claim preservation is highly sensitive to paraphrasing, granularity shifts, and small factual perturbations introduced during translation.

Second, the three open-source LLMs consistently outperform the two dedicated multilingual translation models across all three language pairs, under both \textbf{Aligned-F1} and \textbf{Exact-Aligned-F1}. This pattern likely reflects two factors. One is model scale: the open-source LLMs are substantially larger than M2M100-1.2B and NLLB-200-3.3B. The other is context handling: biography translation in our setting is long-context and discourse-dependent, whereas the multilingual MT models are more constrained in usable context length and therefore less able to exploit broader biography context during incremental translation.

Third, translation difficulty differs across languages. French-to-English translation is overall the easiest setting, with the highest scores across models, while Chinese-to-English is the most challenging. For final model selection, Azerbaijani shows a consistent winner under both coarse and strict matching, leading us to select Gemma-4-31B-it. French is slightly less clear-cut: Gemma-4-31B-it is marginally better on coarse \textbf{Aligned-F1} (93.80 vs.\ 93.77), but Mistral-3.2-24B-it shows a substantially stronger result on \textbf{Exact-Aligned-F1} (48.98 vs.\ 45.48). We therefore select Mistral-3.2-24B-it for French. Chinese is also less clear-cut: Qwen3.6-27B achieves the best coarse \textbf{Aligned-F1}, while Gemma-4-31B-it performs better under \textbf{Exact-Aligned-F1}. We ultimately select Qwen3.6-27B for Chinese, prioritizing stronger semantic claim preservation over stricter exact-form matching.

\begin{table}[t]
\centering
\small
\begin{tabular}{ll}
\toprule
\textbf{Language Pair} & \textbf{Selected MT Model} \\
\midrule
AZ $\rightarrow$ EN & Gemma-4-31B-it \\
FR $\rightarrow$ EN & Mistral-3.2-24B-it \\
ZH $\rightarrow$ EN & Qwen3.6-27B \\
\bottomrule
\end{tabular}
\caption{Final translation model selected for each non-English language.}
\label{tab:mt_selection_summary}
\end{table}

These language-specific translation choices are then used in the \textit{Machine Translation and Generation Pipeline} setting in the main experiments.

\section{Non-English-Side Enrichment Claim Selection}
\label{app:non-English_enrichment_claims}

After extracting English and non-English claims, we align each non-English claim against candidate English claims using the classifier in Section~\ref{sec:claim_classifier}. Because biography-level claim sets can be large, we avoid exhaustive all-pairs verification. For each non-English-side claim \(B \in \mathcal{C}^{X}\), we retrieve candidate English claims \(A \in \mathcal{C}^{en}\) using All-MPNet-Base-v2 cosine similarity, retaining only pairs whose similarity is at least 0.7 and that fall within the top-5 nearest English claims for \(B\). Pairs not retained by this retrieval step are treated as not relevant and are not sent to the LLM classifier.

\paragraph{Retrieval hyperparameters.}
Before top-\(k\) truncation, 67.43\%, 50.32\%, and 39.71\% of Non-English-side claims have at least three, five, and seven candidates above the 0.7 threshold, respectively. We therefore use top-5 as a middle operating point: it allows a broader candidate set than top-3 for claims with several plausible matches, while keeping the worst-case number of LLM verifier calls below that of top-7.

Let \(A\) denote an English claim and \(B\) denote a non-English-side claim. We group non-English-side claims into five status categories according to their best relation to the English side. Exact matches (\(A{=}B\)) and English-more-specific alignments (\(A{>}B\)) are treated as already covered by English and are not selected. Non-English-more-specific and mutually complementary alignments (\(B{>}A\) and \(A{\leftrightarrow}B\)) are treated as non-English-additive because the non-English side contains non-conflicting information not fully expressed in English. We also retain classifier-labeled conflicts (\(A{\perp}B\)) and unmatched non-English claims (\(A{\dashv}B\)) as enrichment candidates. Since both sides are human-written Wikipedia biographies, conflicts are not treated as arbitrary web noise and may require contextual reconciliation. For example, an English-side claim that a person received a master's degree from Stanford and a non-English-side claim that she received a Ph.D. from Stanford may be judged contradictory in isolation, although the full biography may need to resolve whether these refer to distinct degrees or an incomplete English account. Unmatched claims often correspond to genuinely non-English-only biographical information. The downstream generator receives the selected claims together with the original English biography and is responsible for integrating only compatible, supported additions.

\begin{table*}[!htbp]
\centering
\small
\begin{tabular}{lcccccccc}
\toprule
\multirow{2}{*}{\textbf{Direction}}
& \multirow{2}{*}{\textbf{EN}}
& \multirow{2}{*}{\textbf{X}}
& \textbf{Exact}
& \textbf{EN-More}
& \textbf{X-Add.}
& \textbf{Conflict}
& \textbf{Unmatched}
& \multirow{2}{*}{\textbf{Selected}} \\
\cmidrule(lr){4-4} \cmidrule(lr){5-5} \cmidrule(lr){6-6} \cmidrule(lr){7-7} \cmidrule(lr){8-8}
& & & \textbf{$A{=}B$} & \textbf{$A{>}B$} & \textbf{$B{>}A$ / $A{\leftrightarrow}B$} & \textbf{$A{\perp}B$} & \textbf{$A{\dashv}B$} & \\
\midrule
FR $\rightarrow$ EN & 29.8 & 101.8 & 5.2 & 7.2 & 23.5 & 9.7 & 56.2 & 89.4 \\
AZ $\rightarrow$ EN & 33.6 & 93.8 & 6.9 & 8.6 & 26.0 & 12.4 & 39.9 & 78.3 \\
ZH $\rightarrow$ EN & 29.3 & 72.4 & 3.7 & 5.5 & 20.6 & 8.4 & 34.1 & 63.2 \\
\midrule
Avg. & 30.9 & 89.3 & 5.3 & 7.1 & 23.4 & 10.2 & 43.4 & 77.0 \\
\bottomrule
\end{tabular}
\caption{Average counts of English claims, non-English claims, and non-English claim status categories per biography pair. Here \(A\) denotes an English claim and \(B\) denotes a non-English claim. \textbf{Exact} and \textbf{EN-More} are already covered by English and are not selected. \textbf{X-Add.} contains non-English claims with additional non-conflicting information. \textbf{Conflict} contains classifier-labeled contradictions between non-English claims and retrieved English candidates, while \textbf{Unmatched} contains non-English claims with no retrieved or classifier-relevant English-side counterpart. \textbf{Selected} is the set of Non-English-side enrichment claims used as structured evidence for generation.}
\label{tab:enrichment-claim-stats}
\end{table*}

Table~\ref{tab:enrichment-claim-stats} shows that English biographies cover only a small fraction of non-English claims. In the average row, exact matches and English-more-specific alignments account for only 12.4 of 89.3 non-English claims (13.9\%). In contrast, Non-English-additive alignments account for 23.4 claims (26.2\%), indicating that non-English biographies often elaborate on events already mentioned in English with additional factual detail. Unmatched claims form the largest category, with 43.4 claims (48.6\%), showing that nearly half of non-English claims have no relevant English-side counterpart. Classifier-labeled conflicts are smaller but non-negligible (10.2 claims, 11.4\%). Overall, 77.0 of 89.3 non-English claims (86.2\%) are selected as enrichment candidates, supporting our use of non-English biographies as complementary evidence rather than parallel paraphrases of English biographies.

\section{Enrichment Generation Evaluation}
\label{app:generation_eval_results}

\subsection{Evaluation Metrics}
\label{app:generation_eval_metrics}

We evaluate generated biographies in the same English claim space as Section~\ref{subsec:generation_eval}. Let \(\mathcal{C}^{en}\) denote the original English claims, \(\mathcal{C}^{add}\) the selected non-English enrichment claims, and \(\hat{\mathcal{C}}^{gen}\) the claims extracted from the generated biography. To determine support, we compare generated claims against the reference pool \(\mathcal{C}^{ref}=\mathcal{C}^{en}\cup\mathcal{C}^{X}\). Let \(A\) denote a reference claim and \(B\) a generated claim. We categorize each generated claim with contradiction priority: if any retained reference claim is labeled \(A{\perp}B\), then \(B\) is counted as contradicted. Otherwise, \(B\) is valid if some reference claim fully supports it, i.e., \(A{=}B\) or \(A{>}B\). Generated claims with neither contradiction nor full support are counted as unsupported, including generated-more-specific or mutually enriched relations (\(B{>}A\) or \(A{\leftrightarrow}B\)), not-relevant pairs, and claims with no retrieved reference counterpart. \textbf{Halluc.\%} is the sum of \textbf{Unsup.\%} and \textbf{Contr.\%}.

\noindent Claim growth is measured as
\[
\begin{aligned}
\#\mathrm{Gain}
&=
|\hat{\mathcal{C}}^{gen}|-|\mathcal{C}^{en}|,
\\
\mathrm{Coverage}
&=
\frac{\#\mathrm{Gain}}{|\mathcal{C}^{add}|}.
\end{aligned}
\]
We compute the estimated supported gain as
\[
\#\mathrm{Valid}
=
\mathrm{Valid}\times\#\mathrm{Gain},
\]
where \(\mathrm{Valid}\) is \textbf{Valid\%} converted to a proportion. This estimates how much of the net claim growth is supported by the reference pool, rather than counting all supported generated claims as new content.

The primary overall metric \textbf{Bal.} is defined in Section~\ref{subsec:generation_eval}.

\subsection{Full Results}
\label{app:generation_full_results}

Table~\ref{tab:generation_evidence_avg} summarizes the evidence-format averages used in the main analysis. Scores are averaged over all three non-English languages and all three generators for each evidence format. Table~\ref{tab:generation_full_results} reports the full generation results for all language, evidence-format, and generator combinations.

\begin{table}[!htbp]
\centering
\small
\begin{tabular}{lcc}
\toprule
\textbf{Evidence} & \textbf{\#Valid$\uparrow$} & \textbf{Halluc.\%$\downarrow$} \\
\midrule
Raw & 15.89 & 29.49 \\
Translation & 19.05 & 29.59 \\
Claims & \textbf{20.47} & \textbf{23.54} \\
\bottomrule
\end{tabular}
\caption{Average supported additions and hallucination rate across all non-English languages and generators for each evidence format.}
\label{tab:generation_evidence_avg}
\end{table}

\subsection{Qualitative Error Analysis}
\label{app:generation_error_analysis}

\paragraph{Case selection.}
We conduct a controlled manual analysis of outputs from Mistral-Small-3.2-24B-IT. For each language, we select the biography pairs with the lowest and highest numbers of verifier-supported output claims under the best-performing evidence format for that language (Claims for French and Chinese; Translation for Azerbaijani). We then inspect the Raw, Translation, and Claims outputs for the same entities. This yields 18 generated biographies: three languages, two cases per language, and three enrichment methods. Table~\ref{tab:qualitative_case_selection} lists the selected cases. The selection scores are used only to identify contrasting cases; the error types are based on manual inspection.

\begin{table}[!htbp]
\centering
\small
\setlength{\tabcolsep}{3.2pt}
\begin{tabular}{@{}lllr@{}}
\toprule
\textbf{Lang.} & \textbf{Case} & \textbf{Entity} & \textbf{Supported/Output} \\
\midrule
FR & Low  & Tanina Mammeri       & 3/24 \\
FR & High & Juliette Gréco       & 109/261 \\
ZH & Low  & Li Jiaqi             & 1/6 \\
ZH & High & Xie Bingying         & 98/207 \\
AZ & Low  & Aygün Kazimova       & 1/536 \\
AZ & High & Franghiz Ali-Zadeh   & 113/239 \\
\bottomrule
\end{tabular}
\caption{Cases selected for qualitative analysis. Scores are measured under Claims for FR and ZH and Translation for AZ; all three enrichment methods are inspected for every entity.}
\label{tab:qualitative_case_selection}
\end{table}

Table~\ref{tab:qualitative_error_examples} illustrates the principal error patterns. Comparisons are abridged only for space. Because the analysis deliberately samples high- and low-support cases, these examples characterize recurring failure modes rather than their prevalence in the full benchmark.

\begin{table}[!htbp]
\centering
\footnotesize
\setlength{\tabcolsep}{2.5pt}
\renewcommand{\arraystretch}{1.05}
\begin{tabular}{@{}>{\raggedright\arraybackslash}p{0.29\linewidth}p{0.65\linewidth}@{}}
\toprule
\textbf{Pattern and case} & \textbf{Evidence--output comparison} \\
\midrule
Missing event metadata \newline
\textit{Juliette Gréco; Raw}
& \textit{Evidence:} ``made her debut in the play \emph{Victor ou les Enfants au pouvoir} in November 1946.'' \newline
  \textit{Output:} ``began her career in the theater.'' \\
\midrule
Temporal error \newline
\textit{Juliette Gréco; Claims}
& \textit{Evidence:} ``was born on 7 February 1927.'' \newline
  \textit{Output:} ``was born on 6 February 1927.'' \\
\midrule
Education / location error \newline
\textit{Franghiz Ali-Zadeh; Raw}
& \textit{Evidence:} ``studied piano at the Azerbaijan State Conservatory under Ulfan Khalilov.'' \newline
  \textit{Output:} ``studied at the Moscow Conservatory.'' \\
\midrule
Over-specific enumeration \newline
\textit{Xie Bingying; Translation}
& \textit{Evidence:} a 2000 Huaxia Publishing House edition of \emph{Collected Works of Xie Bingying}. \newline
  \textit{Output:} three separate volume-specific claims (Vols.~1--3), each attributed to Anhui Literary Publishing House in August 1999. \\
\midrule
Repetitive generation \newline
\textit{Aygün Kazimova; Translation}
& The 6,384-word output repeatedly cycles through country variants of ``She is a winner of the `Golden Hit Parade' award in Russia [Ukraine, Belarus, \ldots].'' \\
\bottomrule
\end{tabular}
\caption{Representative omissions and hallucination patterns. Evidence is drawn from the reference pool; Output is the generated biography or its claim decomposition.}
\label{tab:qualitative_error_examples}
\end{table}

\subsection{Human Evaluation of Writing Quality}
\label{app:generation_human_eval}

\paragraph{Sampling and annotation.}
We evaluate outputs from Mistral-Small-3.2-24B-IT, the strongest generator in our automatic evaluation. For each language, we sort the outputs from its best-performing evidence format by word count---Claims for French and Chinese, and Translation for Azerbaijani---and select five approximately evenly spaced ranks (0, 25, 50, 74, and 99). We then collect the Raw, Translation, and Claims outputs for the same five entities. This produces 45 outputs: three languages, five biography pairs per language, and three methods. The five buckets are therefore length-based sampling strata defined using the seed method, from shortest (B1) to longest (B5), rather than post-hoc word-count bins for each individual output.

Three computer-science Ph.D. researchers who use English as a working language independently rate all 45 outputs on a 1--5 Likert scale. Readability/Fluency covers grammar, wording, sentence flow, and naturalness; Coherence/Integration covers abrupt insertions, disconnected facts, and local and global flow; and Wikipedia-Style Writing covers encyclopedic tone, section structure, and biographical style. Method labels were retained in the annotation interface for traceability, so the evaluation was not method-blind. We therefore treat it as a writing-quality sanity check rather than a definitive editorial assessment.

\begin{table}[!htbp]
\centering
\small
\setlength{\tabcolsep}{3.2pt}
\begin{tabular}{@{}lccc@{}}
\toprule
\textbf{Method} & \textbf{Read.} & \textbf{Coh.} & \textbf{Wiki.} \\
\midrule
Claims      & \textbf{3.933} & 3.867 & \textbf{3.822} \\
Raw         & 3.844 & \textbf{3.911} & 3.778 \\
Translation & 3.756 & 3.733 & 3.689 \\
\bottomrule
\end{tabular}
\caption{Human ratings of writing quality (1--5; higher is better), averaged over three annotators and 15 outputs per method. Read., Coh., and Wiki. denote Readability/Fluency, Coherence/Integration, and Wikipedia-Style Writing.}
\label{tab:generation_human_eval_overall}
\end{table}

\paragraph{Length-stratified results.}
Table~\ref{tab:generation_human_eval_length} reports the full method-by-length means. Writing quality generally declines as the sampled biographies become longer for all methods. Claims has the highest scores on all three dimensions in B5. This longest-bucket pattern is descriptive: each method--bucket cell contains only three outputs (one per language) and nine ratings, and the Translation B5 cell includes the 6,384-word repetitive Azerbaijani output described in Appendix~\ref{app:generation_error_analysis}.

\begin{table*}[!htbp]
\centering
\small
\setlength{\tabcolsep}{5.0pt}
\begin{tabular}{@{}llccccc@{}}
\toprule
\textbf{Dimension} & \textbf{Method} & \textbf{B1} & \textbf{B2} & \textbf{B3} & \textbf{B4} & \textbf{B5} \\
\midrule
\multirow{3}{*}{Readability}
& Claims      & 4.889 & 4.333 & \textbf{4.000} & 3.222 & \textbf{3.222} \\
& Raw         & 4.889 & 4.333 & 3.444 & 3.556 & 3.000 \\
& Translation & 4.889 & 4.111 & 3.667 & \textbf{3.667} & 2.444 \\
\midrule
\multirow{3}{*}{Coherence}
& Claims      & 4.778 & 4.222 & \textbf{4.000} & 3.444 & \textbf{2.889} \\
& Raw         & \textbf{4.889} & \textbf{4.444} & 3.889 & \textbf{3.667} & 2.667 \\
& Translation & 4.667 & 4.222 & 3.889 & \textbf{3.667} & 2.222 \\
\midrule
\multirow{3}{*}{Wikipedia style}
& Claims      & \textbf{4.889} & 4.222 & \textbf{3.889} & 3.444 & \textbf{2.667} \\
& Raw         & \textbf{4.889} & \textbf{4.444} & 3.556 & 3.667 & 2.333 \\
& Translation & 4.556 & 4.000 & 3.778 & \textbf{3.778} & 2.333 \\
\bottomrule
\end{tabular}
\caption{Writing-quality ratings by method and length-based sampling stratum. Each cell averages nine ratings: three outputs (one per language), each rated by three annotators. Bold marks the highest score within each dimension and bucket, including ties.}
\label{tab:generation_human_eval_length}
\end{table*}

\paragraph{Agreement and aggregate-score reliability.}
Table~\ref{tab:generation_human_eval_agreement} reports ordinal Krippendorff's $\alpha$, which ranges from 0.507 to 0.537 for the subjective discourse-level judgments. Because the reported output scores average three independent ratings, we also compute two-way random-effects, absolute-agreement intraclass correlations. ICC(2,1) measures the reliability of a single rating, whereas ICC(2,3) measures the reliability of the mean of three ratings; the latter ranges from 0.801 to 0.817. ICC treats the Likert scale as approximately interval and is therefore complementary to the ordinal agreement statistic.

\begin{table}[!htbp]
\centering
\small
\setlength{\tabcolsep}{3.0pt}
\begin{tabular}{@{}lccc@{}}
\toprule
\textbf{Dimension} & \textbf{$\alpha$} & \textbf{ICC(2,1)} & \textbf{ICC(2,3)} \\
\midrule
Readability     & 0.512 & 0.572 & 0.801 \\
Coherence       & 0.507 & 0.598 & 0.817 \\
Wikipedia style & 0.537 & 0.592 & 0.813 \\
\bottomrule
\end{tabular}
\caption{Inter-annotator agreement and reliability of single-rater and three-rater mean scores. Krippendorff's $\alpha$ uses the ordinal distance.}
\label{tab:generation_human_eval_agreement}
\end{table}

\section{Main Prompt}
\label{app:prompt}

Due to space constraints, this appendix reports the main prompts used in our pipeline. The complete prompt set, including variants for all adapted claim extraction frameworks and all enrichment generation methods, is provided in the accompanying public repository.

\subsection{Prompt for X-Claimify}
\begin{itemize}
    \item \textbf{Step 1 - Selection:} Figure~\ref{fig:prompt_selection_1} - Figure~\ref{fig:prompt_selection_3}.
    \item \textbf{Step 2 - Disambiguation:} Figure~\ref{fig:prompt_disambiguation_1} - Figure~\ref{fig:prompt_disambiguation_3}.
    \item \textbf{Step 3 - Decomposition:} Figure~\ref{fig:prompt_decomp_1} - Figure~\ref{fig:prompt_decomp_4}.
\end{itemize}

\subsection{Prompt for LLM classifier}
We provide the system prompt in Figure~\ref{fig:prompt_alignment_system_pt1} and Figure~\ref{fig:prompt_alignment_system_pt2}. The user prompt is in Figure~\ref{fig:prompt_alignment_user}.

\subsection{Prompt for "English + non-English" Generation}
We provide the system prompt in Figure~\ref{fig:prompt_sys_en_X_gen} and the user prompt in Figure~\ref{fig:prompt_user_en_X_gen}.

\subsection{Prompt for LLM-based Biography Translation}
We provide the system prompt and the user prompt in Figure~\ref{fig:prompt_bio_mt}. We keep the same prompts for all three open-source LLMs.

\begin{table*}[!htbp]
\centering
\resizebox{\textwidth}{!}{
\begin{tabular}{@{}llcccccccc}
\toprule
\multirow{2}{*}{\textbf{Lang}}
& \multirow{2}{*}{\textbf{Model}}
& \multicolumn{2}{c}{\textbf{Claim Growth}}
& \multicolumn{2}{c}{\textbf{Supported Additions}}
& \multicolumn{3}{c}{\textbf{Hallucination}}
& \multicolumn{1}{c}{\textbf{Overall}} \\
\cmidrule(lr){3-4} \cmidrule(lr){5-6} \cmidrule(lr){7-9} \cmidrule(lr){10-10}
&
& \textbf{\#Gain$\uparrow$} & \textbf{Coverage\%$\uparrow$}
& \textbf{Valid\%$\uparrow$} & \textbf{\#Valid$\uparrow$}
& \textbf{Unsup.\%$\downarrow$} & \textbf{Contr.\%$\downarrow$} & \textbf{Halluc.\%$\downarrow$}
& \cellcolor[rgb]{0.92,0.92,0.92}\textbf{Bal.$\uparrow$} \\
\midrule

\multicolumn{10}{l}{{\cellcolor[rgb]{0.957,0.957,0.957}}\textit{\textbf{Cross-Lingual Generation}}} \\
\multirow{3}{*}{AZ}
& Gemma-4-31B-IT & +17.66 & 22.55 & 69.99 & 12.36 & 29.36 & 0.65 & 30.01 & \cellcolor[rgb]{0.92,0.92,0.92}10.79 \\
& Qwen3.6-27B & +23.41 & 29.90 & 66.12 & 15.48 & 32.81 & 1.07 & 33.88 & \cellcolor[rgb]{0.92,0.92,0.92}11.26 \\
& Mistral-Small-3.2-24B-IT & +19.41 & 24.79 & 71.52 & 13.88 & 27.58 & 0.90 & 28.48 & \cellcolor[rgb]{0.92,0.92,0.92}20.54 \\
\cline{1-10}
\multirow{3}{*}{FR}
& Gemma-4-31B-IT & +21.94 & 24.54 & 72.43 & 15.89 & 27.08 & 0.49 & 27.57 & \cellcolor[rgb]{0.92,0.92,0.92}30.33 \\
& Qwen3.6-27B & +27.22 & 30.45 & 68.35 & 18.61 & 31.06 & 0.58 & 31.65 & \cellcolor[rgb]{0.92,0.92,0.92}28.77 \\
& Mistral-Small-3.2-24B-IT & +24.94 & 27.90 & 74.08 & 18.47 & 25.23 & 0.70 & 25.92 & \cellcolor[rgb]{0.92,0.92,0.92}44.24 \\
\cline{1-10}
\multirow{3}{*}{ZH}
& Gemma-4-31B-IT & +19.53 & 30.90 & 71.86 & 14.03 & 26.90 & 1.24 & 28.14 & \cellcolor[rgb]{0.92,0.92,0.92}22.03 \\
& Qwen3.6-27B & +24.79 & 39.22 & 67.77 & 16.80 & 30.59 & 1.64 & 32.23 & \cellcolor[rgb]{0.92,0.92,0.92}20.62 \\
& Mistral-Small-3.2-24B-IT & +24.19 & 38.28 & 72.44 & 17.52 & 26.22 & 1.34 & 27.56 & \cellcolor[rgb]{0.92,0.92,0.92}36.24 \\
\hline

\multicolumn{10}{l}{{\cellcolor[rgb]{0.957,0.957,0.957}}\textit{\textbf{Machine Translation and Generation Pipeline}}} \\
\multirow{3}{*}{AZ}
& Gemma-4-31B-IT & +19.43 & 24.81 & 69.27 & 13.46 & 29.80 & 0.93 & 30.73 & \cellcolor[rgb]{0.92,0.92,0.92}12.75 \\
& Qwen3.6-27B & +23.67 & 30.23 & 66.73 & 15.80 & 32.29 & 0.98 & 33.27 & \cellcolor[rgb]{0.92,0.92,0.92}14.11 \\
& Mistral-Small-3.2-24B-IT & +32.47 & 41.47 & 72.10 & 23.41 & 26.66 & 1.24 & 27.90 & \cellcolor[rgb]{0.92,0.92,0.92}56.54 \\
\cline{1-10}
\multirow{3}{*}{FR}
& Gemma-4-31B-IT & +24.48 & 27.38 & 71.63 & 17.53 & 27.84 & 0.53 & 28.37 & \cellcolor[rgb]{0.92,0.92,0.92}34.02 \\
& Qwen3.6-27B & +28.62 & 32.01 & 69.40 & 19.86 & 29.99 & 0.60 & 30.60 & \cellcolor[rgb]{0.92,0.92,0.92}36.20 \\
& Mistral-Small-3.2-24B-IT & +33.66 & 37.65 & 75.48 & 25.41 & 23.82 & 0.70 & 24.52 & \cellcolor[rgb]{0.92,0.92,0.92}73.18 \\
\cline{1-10}
\multirow{3}{*}{ZH}
& Gemma-4-31B-IT & +22.13 & 35.02 & 69.12 & 15.30 & 29.33 & 1.55 & 30.88 & \cellcolor[rgb]{0.92,0.92,0.92}18.97 \\
& Qwen3.6-27B & +27.33 & 43.24 & 68.32 & 18.67 & 29.57 & 2.11 & 31.68 & \cellcolor[rgb]{0.92,0.92,0.92}28.90 \\
& Mistral-Small-3.2-24B-IT & +30.73 & 48.62 & 71.65 & 22.02 & 26.63 & 1.72 & 28.35 & \cellcolor[rgb]{0.92,0.92,0.92}50.27 \\
\hline

\multicolumn{10}{l}{{\cellcolor[rgb]{0.957,0.957,0.957}}\textit{\textbf{Claim-Based Enrichment (Ours)}}} \\
\multirow{3}{*}{AZ}
& Gemma-4-31B-IT & +22.19 & 28.34 & 71.67 & 15.90 & 27.91 & 0.43 & 28.33 & \cellcolor[rgb]{0.92,0.92,0.92}28.25 \\
& Qwen3.6-27B & +28.32 & 36.17 & 69.61 & 19.71 & 29.79 & 0.60 & 30.39 & \cellcolor[rgb]{0.92,0.92,0.92}36.25 \\
& Mistral-Small-3.2-24B-IT & +22.30 & 28.48 & 77.94 & 17.38 & 21.51 & 0.55 & 22.06 & \cellcolor[rgb]{0.92,0.92,0.92}51.07 \\
\cline{1-10}
\multirow{3}{*}{FR}
& Gemma-4-31B-IT & +25.37 & 28.38 & 74.78 & 18.97 & 24.91 & 0.31 & 25.22 & \cellcolor[rgb]{0.92,0.92,0.92}48.00 \\
& Qwen3.6-27B & +32.89 & 36.79 & 73.86 & 24.29 & 25.88 & 0.26 & 26.14 & \cellcolor[rgb]{0.92,0.92,0.92}64.62 \\
& Mistral-Small-3.2-24B-IT & +32.65 & 36.52 & 80.31 & 26.22 & 19.36 & 0.33 & 19.69 & \cellcolor[rgb]{0.92,0.92,0.92}89.57 \\
\cline{1-10}
\multirow{3}{*}{ZH}
& Gemma-4-31B-IT & +23.79 & 37.64 & 78.48 & 18.67 & 21.15 & 0.37 & 21.52 & \cellcolor[rgb]{0.92,0.92,0.92}57.23 \\
& Qwen3.6-27B & +28.42 & 44.97 & 77.47 & 22.02 & 22.16 & 0.37 & 22.53 & \cellcolor[rgb]{0.92,0.92,0.92}66.50 \\
& Mistral-Small-3.2-24B-IT & +25.07 & 39.67 & 84.05 & 21.07 & 15.65 & 0.30 & 15.95 & \cellcolor[rgb]{0.92,0.92,0.92}81.42 \\
\bottomrule
\end{tabular}
}
\vspace{2pt}
\caption{Full generation results on \textsc{CLAW-4L}. Counts are averages per biography and rates are percentages.
\#Gain is the increase in English claims after enrichment, Coverage\% normalizes \#Gain by the number
of selected non-English-side enrichment claims, and \#Valid estimates the number of gained claims supported
by the reference pool. Unsup.\% aggregates unsupported, not-relevant, and unknown claims; Halluc.\% is
the sum of Unsup.\% and Contr.\%. Bal. is an overall trade-off score between supported additions and
supported-claim rate.}
\label{tab:generation_full_results}
\end{table*}

\begin{figure*}[t]
\centering
\begin{tcblisting}{
    enhanced,
    width=\textwidth,
    colback=gray!4,
    colframe=gray!50,
    boxrule=0.4pt,
    arc=1.5pt,
    title=\textbf{System Prompt for X-Claimify - Selection},
    colbacktitle=gray!15,
    coltitle=black,
    fonttitle=\small\bfseries,
    attach boxed title to top left={yshift=-2mm,xshift=2mm},
    boxed title style={
        boxrule=0pt,
        arc=1pt,
    },
    left=4pt,
    right=4pt,
    top=6pt,
    bottom=4pt,
    listing only,
    breakable=false,
    listing options={
        breaklines=true,
        breakatwhitespace=false,
        basicstyle=\ttfamily\small
    }
}
You are an assistant to a fact-checker. You will be given an excerpt from a biography. If it contains "[...]", this means that you are NOT seeing all sentences in the biography. You will also be given a particular sentence of interest from the biography. The excerpt and sentence may be written in any language. Your task is to determine whether this particular sentence contains at least one specific and verifiable proposition, and if so, to return a complete English sentence that contains only verifiable information.

Note the following rules:
- It does NOT matter whether the proposition is true or false.
- It does NOT matter whether the proposition is important to the biography.
- It does NOT matter whether the proposition contains ambiguous terms, e.g., a pronoun without a clear antecedent. Assume that the fact-checker has the necessary information to resolve all ambiguities.
- You will NOT consider whether a sentence contains a citation when determining if it has a specific and verifiable proposition.
- Preserve all verifiable information in the sentence. Only remove content that is subjective, evaluative, speculative, interpretive, or otherwise not specifically verifiable.
- If the sentence contains a specific and verifiable proposition, the final rewritten sentence must be in English.
- Do NOT simplify the sentence more than necessary.

You must consider the preceding and following sentences when determining if the sentence has a specific and verifiable proposition. For example:
- if preceding sentence = "Marie Curie moved to Paris in 1891." and sentence = "There she began studying at the Sorbonne." then sentence contains a specific and verifiable proposition.
- if preceding sentence = "Frida Kahlo became known for her self-portraits." and sentence = "These works often drew on her personal experiences." then sentence contains a specific and verifiable proposition.
- if preceding sentence = "Angela Merkel served as Chancellor of Germany." and sentence = "She later led the government during the financial crisis." then sentence contains a specific and verifiable proposition.
- if preceding sentence = "Ada Lovelace worked with Charles Babbage." and sentence = "Lovelace described the machine's potential in her notes." then sentence contains a specific and verifiable proposition.
- if sentence = "Her life and career included many achievements across different fields" and the following sentences expand on this point (e.g., give specific achievements, positions, or dates), then sentence is an introduction and does NOT contain a specific and verifiable proposition.
- if sentence = "In summary, Marie Curie left a lasting impact on science and society" and the preceding sentences provide details on these topics, then sentence is a conclusion and does NOT contain a specific and verifiable proposition.

Here are some examples of sentences that do NOT contain any specific and verifiable propositions:
- She is widely regarded as an inspiring figure
- Her achievements were truly remarkable
- Wangari Maathai played an important role in history
- Her story shows the power of perseverance
- This implies that she was a courageous person

Here are some examples of sentences that likely contain a specific and verifiable proposition and how they can be rewritten to only include verifiable information:
- Her groundbreaking work in chemistry changed the world -> "She worked in chemistry"
- She became one of the most influential politicians of her time -> "She was a politician"
- Frida Kahlo's distinctive artistic style made her one of Mexico's most celebrated painters -> "Frida Kahlo was a painter"
\end{tcblisting}
\caption{Prompt for X-Claimify - Selection (1)}
\label{fig:prompt_selection_1}
\end{figure*}

\begin{figure*}[t]
\centering
\begin{tcblisting}{
    enhanced,
    width=\textwidth,
    colback=gray!4,
    colframe=gray!50,
    boxrule=0.4pt,
    arc=1.5pt,
    title=\textbf{System Prompt for X-Claimify - Selection (Continued)},
    colbacktitle=gray!15,
    coltitle=black,
    fonttitle=\small\bfseries,
    attach boxed title to top left={yshift=-2mm,xshift=2mm},
    boxed title style={
        boxrule=0pt,
        arc=1pt,
    },
    left=4pt,
    right=4pt,
    top=6pt,
    bottom=4pt,
    listing only,
    breakable=false,
    listing options={
        breaklines=true,
        breakatwhitespace=false,
        basicstyle=\ttfamily\small
    }
}
- Wangari Maathai's environmental activism was crucial for Kenya -> "Wangari Maathai was an environmental activist"
- She later moved to Paris, where she began exhibiting her work -> "She later moved to Paris, where she began exhibiting her work"

Your output must adhere to the following format exactly. Only replace what's inside the <insert> tags; do NOT remove the step headers.
Sentence:
<insert>

4-step stream of consciousness thought process (1. reflect on criteria at a high-level -> 2. provide an objective description of the excerpt, the sentence, and its surrounding sentences -> 3. consider all possible perspectives on whether the sentence explicitly or implicitly contains a specific and verifiable proposition, or if it just contains an introduction for the following sentence(s), a conclusion for the preceding sentence(s), broad or generic statements, opinions, interpretations, speculation, etc. -> 4. only if it contains a specific and verifiable proposition: reflect on whether any changes are needed to ensure that the entire sentence only contains verifiable information and that the final rewritten sentence is in English):
<insert>

Final submission:
<insert 'Contains a specific and verifiable proposition' or 'Does NOT contain a specific and verifiable proposition'>

Sentence with only verifiable information (in English):
<insert English sentence, or 'None' if the sentence does NOT contain a specific and verifiable proposition>

\end{tcblisting}
\caption{Prompt for X-Claimify - Selection (2)}
\label{fig:prompt_selection_2}
\end{figure*}

\begin{figure*}[t]
\centering
\begin{tcblisting}{
    enhanced,
    width=\textwidth,
    colback=gray!4,
    colframe=gray!50,
    boxrule=0.4pt,
    arc=1.5pt,
    title=\textbf{User Prompt for X-Claimify - Selection},
    colbacktitle=gray!15,
    coltitle=black,
    fonttitle=\small\bfseries,
    attach boxed title to top left={yshift=-2mm,xshift=2mm},
    boxed title style={
        boxrule=0pt,
        arc=1pt,
    },
    left=4pt,
    right=4pt,
    top=6pt,
    bottom=4pt,
    listing only,
    breakable=false,
    listing options={
        breaklines=true,
        breakatwhitespace=false,
        basicstyle=\ttfamily\small
    }
}
Biography excerpt: 
{excerpt}

Sentence: 
{sentence}

\end{tcblisting}
\caption{Prompt for X-Claimify - Selection (3)}
\label{fig:prompt_selection_3}
\end{figure*}

\begin{figure*}[t]
\centering
\begin{tcblisting}{
    enhanced,
    width=\textwidth,
    colback=gray!4,
    colframe=gray!50,
    boxrule=0.4pt,
    arc=1.5pt,
    title=\textbf{System Prompt for X-Claimify - Disambiguation},
    colbacktitle=gray!15,
    coltitle=black,
    fonttitle=\small\bfseries,
    attach boxed title to top left={yshift=-2mm,xshift=2mm},
    boxed title style={
        boxrule=0pt,
        arc=1pt,
    },
    left=4pt,
    right=4pt,
    top=6pt,
    bottom=4pt,
    listing only,
    breakable=false,
    listing options={
        breaklines=true,
        breakatwhitespace=false,
        basicstyle=\ttfamily\small
    }
}
You are an assistant to a fact-checker. You will be given a biography subject, an excerpt from a biography, and a particular sentence from the biography. If the excerpt contains "[...]", this means that you are NOT seeing all sentences in the biography. The text before and after this sentence will be referred to as "the context". Your task is to "decontextualize" the sentence, which means:

1. determine whether it is possible to resolve partial names and undefined acronyms/abbreviations in the sentence using the biography subject and the context; if it is possible, you will make the necessary changes to the sentence
2. determine whether the sentence in isolation contains linguistic ambiguity that has a clear resolution using the biography subject and the context; if it does, you will make the necessary changes to the sentence

Note the following rules:
- "Linguistic ambiguity" refers to the presence of multiple possible meanings in a sentence. Vagueness and generality are NOT linguistic ambiguity. Linguistic ambiguity includes referential and structural ambiguity. Temporal ambiguity is a type of referential ambiguity.
- If a name is only partially given in the sentence, but the full name is provided in the biography subject or the context, the DecontextualizedSentence must always use the full name.
- The same rule applies to definitions for acronyms and abbreviations. However, the lack of a full name or a definition in the biography subject or the context does NOT count as linguistic ambiguity; in this case, you will just leave the name, acronym, or abbreviation as is.
- Do NOT include any citations in the DecontextualizedSentence.
- Do NOT use any external knowledge beyond what is stated in the biography subject, context, and sentence.
- Do NOT judge whether the sentence is broad, generic, important, or worth fact-checking. Assume the sentence has already been selected. Your only goal is to resolve ambiguity where possible.
- If an expression is ambiguous but the biography subject and context do not provide a clear consensus resolution, leave it unchanged. Only write "Cannot be decontextualized" if producing a standalone sentence would require guessing between multiple plausible interpretations.

Here are some correct examples that you should pay attention to:

1. Biography subject = "Marie Curie", Context = "Marie Curie moved to Paris in 1891. There she began studying at the Sorbonne.", Sentence = "There she began studying at the Sorbonne."
- For referential ambiguity, "There" and "she" are unclear. A group of readers shown the biography subject and the context would likely reach consensus about the correct interpretation: "There" refers to Paris and "she" refers to Marie Curie.
- DecontextualizedSentence: In Paris, Marie Curie began studying at the Sorbonne.

2. Biography subject = "Frida Kahlo", Context = "[...] Frida Kahlo became known for her self-portraits. These works often drew on her personal experiences.", Sentence = "These works often drew on her personal experiences."
- For referential ambiguity, "These works" and "her" are unclear. A group of readers shown the biography subject and the context would likely reach consensus about the correct interpretation: "These works" refers to Frida Kahlo's self-portraits and "her" refers to Frida Kahlo.
- DecontextualizedSentence: Frida Kahlo's self-portraits often drew on her personal experiences.

\end{tcblisting}
\caption{Prompt for X-Claimify - Disambiguation (1)}
\label{fig:prompt_disambiguation_1}
\end{figure*}

\begin{figure*}[t]
\centering
\begin{tcblisting}{
    enhanced,
    width=\textwidth,
    colback=gray!4,
    colframe=gray!50,
    boxrule=0.4pt,
    arc=1.5pt,
    title=\textbf{System Prompt for X-Claimify - Disambiguation (Continued)},
    colbacktitle=gray!15,
    coltitle=black,
    fonttitle=\small\bfseries,
    attach boxed title to top left={yshift=-2mm,xshift=2mm},
    boxed title style={
        boxrule=0pt,
        arc=1pt,
    },
    left=4pt,
    right=4pt,
    top=6pt,
    bottom=4pt,
    listing only,
    breakable=false,
    listing options={
        breaklines=true,
        breakatwhitespace=false,
        basicstyle=\ttfamily\small
    }
}
4. Biography subject = "Angela Merkel", Context = "Angela Merkel served as Chancellor of Germany. She later led the government during the financial crisis.", Sentence = "She later led the government during the financial crisis."
- For referential ambiguity, "She" is unclear. A group of readers shown the biography subject and the context would likely reach consensus that "She" refers to Angela Merkel.
- "the government" and "the financial crisis" are not further specified in the biography subject or the context, so they remain unchanged.
- DecontextualizedSentence: Angela Merkel later led the government during the financial crisis.

5. Biography subject = "Jane Doe", Context = "Jane Doe collaborated with several researchers during her early academic career.", Sentence = "They later moved to Berlin."
- For referential ambiguity, "They" is unclear in isolation. A group of readers shown the biography subject and the context would likely fail to reach consensus about whether "They" refers to Jane Doe and one specific collaborator, Jane Doe and several collaborators, or only the collaborators.
- DecontextualizedSentence: Cannot be decontextualized

Your output must adhere to the following format exactly. Do NOT remove any section headers. Only replace the content inside the <insert> tags. Do NOT add any extra headers, commentary, or text outside these five sections.

Sentence:
<insert the original sentence exactly as given>

Incomplete Names, Acronyms, Abbreviations:
<insert step-by-step reasoning about whether the sentence contains any partial names, acronyms, or abbreviations that can be resolved using the biography subject or the context>

Linguistic Ambiguity in the Sentence:
<insert step-by-step reasoning about referential ambiguity and structural ambiguity. If the sentence could be interpreted in multiple ways, explicitly write: "The sentence could be interpreted as: ..." before judging whether readers would likely reach consensus>

Changes Needed to Decontextualize the Sentence:
<insert a list of all changes needed to make the sentence fully decontextualized; if no changes are needed, write "None"; if the sentence cannot be decontextualized, write "N/A">

DecontextualizedSentence:
<insert the final decontextualized sentence; if the sentence cannot be decontextualized, write exactly "Cannot be decontextualized">

\end{tcblisting}
\caption{Prompt for X-Claimify - Disambiguation (2)}
\label{fig:prompt_disambiguation_2}
\end{figure*}

\begin{figure*}[t]
\centering
\begin{tcblisting}{
    enhanced,
    width=\textwidth,
    colback=gray!4,
    colframe=gray!50,
    boxrule=0.4pt,
    arc=1.5pt,
    title=\textbf{User Prompt for X-Claimify - Disambiguation},
    colbacktitle=gray!15,
    coltitle=black,
    fonttitle=\small\bfseries,
    attach boxed title to top left={yshift=-2mm,xshift=2mm},
    boxed title style={
        boxrule=0pt,
        arc=1pt,
    },
    left=4pt,
    right=4pt,
    top=6pt,
    bottom=4pt,
    listing only,
    breakable=false,
    listing options={
        breaklines=true,
        breakatwhitespace=false,
        basicstyle=\ttfamily\small
    }
}
Biography subject:
{subject}

Excerpt:
{excerpt}

Sentence:
{sentence}

\end{tcblisting}
\caption{Prompt for X-Claimify - Disambiguation (3)}
\label{fig:prompt_disambiguation_3}
\end{figure*}

\begin{figure*}[t]
\centering
\begin{tcblisting}{
    enhanced,
    width=\textwidth,
    colback=gray!4,
    colframe=gray!50,
    boxrule=0.4pt,
    arc=1.5pt,
    title=\textbf{System Prompt for X-Claimify - Decomposition},
    colbacktitle=gray!15,
    coltitle=black,
    fonttitle=\small\bfseries,
    attach boxed title to top left={yshift=-2mm,xshift=2mm},
    boxed title style={
        boxrule=0pt,
        arc=1pt,
    },
    left=4pt,
    right=4pt,
    top=6pt,
    bottom=4pt,
    listing only,
    breakable=false,
    listing options={
        breaklines=true,
        breakatwhitespace=false,
        basicstyle=\ttfamily\small
    }
}
You are an assistant for a group of fact-checkers. You will be given a biography subject, an excerpt from a biography, and a particular sentence from the biography. If the excerpt contains "[...]", this means that you are NOT seeing all sentences in the biography. The text before and after this sentence will be referred to as "the context".

Your task is to identify all specific and verifiable propositions in the sentence, decompose them if necessary, and output them as a JSON list of dictionaries, where each dictionary corresponds to exactly one claim.

A proposition is "decontextualized" if:
(1) it is fully self-contained, meaning it can be understood in isolation without the biography subject, the context, or the other propositions, AND
(2) its meaning in isolation matches its meaning when interpreted alongside the biography subject, the context, and the other propositions.

The propositions should be the simplest possible discrete units of information, but they should NOT be decomposed into trivial subparts that are not independently worth fact-checking.

Note the following rules:
- Assume the sentence has already passed the selection and disambiguation stages. Do NOT decide whether the sentence is worth fact-checking. Only extract and decompose the specific and verifiable propositions that it contains.
- A sentence may contain zero, one, or multiple specific and verifiable propositions.
- Do NOT include subjective, evaluative, speculative, rhetorical, or interpretive content as propositions unless the sentence factually attributes that content to a source or speaker.
- Do NOT use any external knowledge beyond what is stated in the biography subject, the context, and the sentence.
- Do NOT include any citations.

Sentences like the following do NOT contain a specific and verifiable proposition:
- Marie Curie's legacy remains unmatched
- Wangari Maathai became a symbol of hope for many people
- Frida Kahlo's life story continues to inspire generations
- The exhibition marked a major cultural turning point
- Scholars often regard her as one of the most visionary figures of her time
- This suggests that she possessed extraordinary emotional strength

Additional extraction rules:
- Sometimes a sentence is partly evaluative or general, but still contains a specific and verifiable proposition. In such cases, extract only the verifiable part.
- If the sentence contains coordination, apposition, or multiple factual units, decompose it into as many specific and verifiable propositions as needed, but no further.
- Do NOT decompose a proposition into trivial subparts that are not independently useful for fact-checking. For example, do NOT split "Ada Lovelace described the machine's potential in her notes" into "Ada Lovelace had notes" and "The machine had potential".
- If a referential term, partial name, acronym, abbreviation, temporal expression, or location can be clarified using the biography subject or the context, clarify it in the final proposition.
- If a relative expression such as "there", "then", "later", or "that year" cannot be made standalone from the visible context, do NOT guess. Only omit it if removing it does NOT materially change the factual content being extracted.
- If the excerpt contains "[...]", do NOT assume hidden information unless it is strongly implied by the sentence and the visible context.

\end{tcblisting}
\caption{Prompt for X-Claimify - Decomposition (1)}
\label{fig:prompt_decomp_1}
\end{figure*}

\begin{figure*}[t]
\centering
\begin{tcblisting}{
    enhanced,
    width=\textwidth,
    colback=gray!4,
    colframe=gray!50,
    boxrule=0.4pt,
    arc=1.5pt,
    title=\textbf{System Prompt for X-Claimify - Decomposition (Continued)},
    colbacktitle=gray!15,
    coltitle=black,
    fonttitle=\small\bfseries,
    attach boxed title to top left={yshift=-2mm,xshift=2mm},
    boxed title style={
        boxrule=0pt,
        arc=1pt,
    },
    left=4pt,
    right=4pt,
    top=6pt,
    bottom=4pt,
    listing only,
    breakable=false,
    listing options={
        breaklines=true,
        breakatwhitespace=false,
        basicstyle=\ttfamily\small
    }
}
For each claim, output one dictionary with exactly the following keys:
- "subject": the entity that the claim is about; use a fully clarified string if possible, otherwise null
- "predicate": the main relation or event expressed by the claim; use a short verbal phrase if possible, otherwise null
- "object": the main object, complement, non-English, or content of the claim; fill this whenever possible as part of the core subject-predicate-object structure, even if it overlaps with a more specific semantic field such as "location" or "time"
- "time": the time expression if explicitly stated or clearly resolvable from the visible context; otherwise null
- "location": the location expression if explicitly stated or clearly resolvable from the visible context; otherwise null
- "reason": the reason or purpose expression only if explicitly stated in the claim; otherwise null
- "manner": the manner or means expression only if explicitly stated in the claim; otherwise null
- "hedge": the exact hedge word or phrase if the claim is hedged (for example: "reportedly", "may have", "according to some accounts"); otherwise "No"
- "claim": the final specific, verifiable, and fully decontextualized claim as a single string
- "subject", "predicate", and "object" form the core claim frame and should be filled whenever possible. Other fields provide optional semantic refinements.

Important formatting rules:
- Output must be valid JSON.
- Output must be a JSON list.
- Each item in the list must be a dictionary with exactly the keys listed above.
- All values must be either a string or null, except:
- "hedge", which must be either a string or "No"
- "claim", which must always be a single string
- If the sentence contains no specific and verifiable propositions, output [].
- Do NOT output any explanation, markdown, headers, or extra text outside the JSON.

Here are some correct examples:

Example 1:
Biography subject = "Marie Curie"
Context = "Marie Curie moved to Paris in 1891. There she began studying at the Sorbonne."
Sentence = "There she began studying at the Sorbonne."
Output =
[
{
  "subject": "Marie Curie",
  "predicate": "began studying",
  "object": "the Sorbonne",
  "time": null,
  "location": "Paris",
  "reason": null,
  "manner": null,
  "hedge": "No",
  "claim": "Marie Curie began studying at the Sorbonne in Paris."
}
]

\end{tcblisting}
\caption{Prompt for X-Claimify - Decomposition (2)}
\label{fig:prompt_decomp_2}
\end{figure*}

\begin{figure*}[t]
\centering
\begin{tcblisting}{
    enhanced,
    width=\textwidth,
    colback=gray!4,
    colframe=gray!50,
    boxrule=0.4pt,
    arc=1.5pt,
    title=\textbf{System Prompt for X-Claimify - Decomposition (Continued)},
    colbacktitle=gray!15,
    coltitle=black,
    fonttitle=\small\bfseries,
    attach boxed title to top left={yshift=-2mm,xshift=2mm},
    boxed title style={
        boxrule=0pt,
        arc=1pt,
    },
    left=4pt,
    right=4pt,
    top=6pt,
    bottom=4pt,
    listing only,
    breakable=false,
    listing options={
        breaklines=true,
        breakatwhitespace=false,
        basicstyle=\ttfamily\small
    }
}
Example 2:
Biography subject = "Wangari Maathai"
Context = "Wangari Maathai founded the Green Belt Movement in 1977 and later received the Nobel Peace Prize."
Sentence = "Wangari Maathai founded the Green Belt Movement in 1977 and later received the Nobel Peace Prize."
Output =
[
{
  "subject": "Wangari Maathai",
  "predicate": "founded",
  "object": "the Green Belt Movement",
  "time": "1977",
  "location": null,
  "reason": null,
  "manner": null,
  "hedge": "No",
  "claim": "Wangari Maathai founded the Green Belt Movement in 1977."
},
{
  "subject": "Wangari Maathai",
  "predicate": "received",
  "object": "the Nobel Peace Prize",
  "time": null,
  "location": null,
  "reason": null,
  "manner": null,
  "hedge": "No",
  "claim": "Wangari Maathai received the Nobel Peace Prize."
}
]

Example 3:
Biography subject = "Maya Angelou"
Context = "According to some accounts, Maya Angelou traveled by train to California in 1940 to join her mother."
Sentence = "According to some accounts, Maya Angelou traveled by train to California in 1940 to join her mother."
Output =
[
{
  "subject": "Maya Angelou",
  "predicate": "traveled",
  "object": "California",
  "time": "1940",
  "location": "California",
  "reason": "to join her mother",
  "manner": "by train",
  "hedge": "according to some accounts",
  "claim": "According to some accounts, Maya Angelou traveled by train to California in 1940 to join her mother."
}
]

\end{tcblisting}
\caption{Prompt for X-Claimify - Decomposition (3)}
\label{fig:prompt_decomp_3}
\end{figure*}

\begin{figure*}[t]
\centering
\begin{tcblisting}{
    enhanced,
    width=\textwidth,
    colback=gray!4,
    colframe=gray!50,
    boxrule=0.4pt,
    arc=1.5pt,
    title=\textbf{User Prompt for X-Claimify - Decomposition},
    colbacktitle=gray!15,
    coltitle=black,
    fonttitle=\small\bfseries,
    attach boxed title to top left={yshift=-2mm,xshift=2mm},
    boxed title style={
        boxrule=0pt,
        arc=1pt,
    },
    left=4pt,
    right=4pt,
    top=6pt,
    bottom=4pt,
    listing only,
    breakable=false,
    listing options={
        breaklines=true,
        breakatwhitespace=false,
        basicstyle=\ttfamily\small
    }
}
Biography subject:
{subject}

Excerpt:
{excerpt}

Sentence:
{sentence}

\end{tcblisting}
\caption{Prompt for X-Claimify - Decomposition (4)}
\label{fig:prompt_decomp_4}
\end{figure*}

\begin{figure*}[t]
\centering
\begin{tcblisting}{
    enhanced,
    width=\textwidth,
    colback=gray!4,
    colframe=gray!50,
    boxrule=0.4pt,
    arc=1.5pt,
    title=\textbf{System Prompt for Claim Infobox Extraction},
    colbacktitle=gray!15,
    coltitle=black,
    fonttitle=\small\bfseries,
    attach boxed title to top left={yshift=-2mm,xshift=2mm},
    boxed title style={
        boxrule=0pt,
        arc=1pt,
    },
    left=4pt,
    right=4pt,
    top=6pt,
    bottom=4pt,
    listing only,
    breakable=false,
    listing options={
        breaklines=true,
        breakatwhitespace=false,
        basicstyle=\ttfamily\small
    }
}
You are an assistant for a group of fact-checkers.

You will be given exactly one already-written claim. Your task is NOT to extract new claims from a sentence, NOT to split the claim into multiple propositions, and NOT to rewrite the dataset. Your task is only to fill the infobox fields that describe the given claim.

For the given claim, output exactly one JSON object with exactly these keys:
- "subject": the entity that the claim is about; use a fully clarified string if possible, otherwise null
- "predicate": the main relation or event expressed by the claim; use a short verbal phrase if possible, otherwise null
- "object": the main object, complement, non-English, or content of the claim; fill this whenever possible as part of the core subject-predicate-object structure, even if it overlaps with a more specific semantic field such as "location" or "time"
- "time": the time expression if explicitly stated in the claim; otherwise null
- "location": the location expression if explicitly stated in the claim; otherwise null
- "reason": the reason or purpose expression only if explicitly stated in the claim; otherwise null
- "manner": the manner or means expression only if explicitly stated in the claim; otherwise null
- "hedge": the exact hedge word or phrase if the claim is hedged, for example "reportedly", "may have", or "according to some accounts"; otherwise "No"
- "claim": copy the input claim exactly, except for trimming surrounding whitespace

Example:
Claim: According to some accounts, Maya Angelou traveled by train to California in 1940 to join her mother.
Output:
{
  "subject": "Maya Angelou",
  "predicate": "traveled",
  "object": "California",
  "time": "1940",
  "location": "California",
  "reason": "to join her mother",
  "manner": "by train",
  "hedge": "according to some accounts",
  "claim": "According to some accounts, Maya Angelou traveled by train to California in 1940 to join her mother."
}
\end{tcblisting}
\caption{Prompt for Extraction Infobox (1)}
\label{fig:prompt_infobox}
\end{figure*}

\begin{figure*}[t]
\centering
\begin{tcblisting}{
    enhanced,
    width=\textwidth,
    colback=gray!4,
    colframe=gray!50,
    boxrule=0.4pt,
    arc=1.5pt,
    title=\textbf{User Prompt for Claim Infobox Extraction},
    colbacktitle=gray!15,
    coltitle=black,
    fonttitle=\small\bfseries,
    attach boxed title to top left={yshift=-2mm,xshift=2mm},
    boxed title style={
        boxrule=0pt,
        arc=1pt,
    },
    left=4pt,
    right=4pt,
    top=6pt,
    bottom=4pt,
    listing only,
    breakable=false,
    listing options={
        breaklines=true,
        breakatwhitespace=false,
        basicstyle=\ttfamily\small
    }
}
Claim:
{claim}
\end{tcblisting}
\caption{Prompt for Extraction Infobox (2)}
\label{fig:user_prompt_infobox}
\end{figure*}

\begin{figure*}[t]
\centering
\begin{tcblisting}{
    enhanced,
    width=\textwidth,
    colback=gray!4,
    colframe=gray!50,
    boxrule=0.4pt,
    arc=1.5pt,
    title=\textbf{System Prompt for Claim Alignment - w/o infobox},
    colbacktitle=gray!15,
    coltitle=black,
    fonttitle=\small\bfseries,
    attach boxed title to top left={yshift=-2mm,xshift=2mm},
    boxed title style={boxrule=0pt, arc=1pt},
    left=4pt, right=4pt, top=6pt, bottom=4pt,
    listing only,
    breakable=false,
    listing options={
        breaklines=true,
        breakatwhitespace=false,
        basicstyle=\ttfamily\small
    }
}
You are an expert cross-lingual claim relation annotator.
First decide Alignment, then decide Enrichment only when Alignment is "Aligned".

Alignment choices:
- Aligned
- Contradicted
- Not Relevant

Enrichment choices (only for aligned pairs):
- None
- A > B
- B > A
- A <> B

Label definitions:
- Not Relevant: Claim A and Claim B are about different factual aspects, even if they refer to the same person/entity. A factual aspect can be a date, place, occupation, relationship, event, award, work, role, quantity, or other biographical fact.
- Contradicted: Claim A and Claim B are about the same factual aspect, but assert mutually exclusive values or states, so they cannot both be true.
- Aligned: Claim A and Claim B are about the same factual aspect and can both be true. This includes exact paraphrases, translation variants, precision differences, and non-conflicting added detail.
- Enrichment None: Use when both claims contain the same set of atomic facts with no additional details on either side.
- Enrichment A > B: Use when Claim A includes the factual content of Claim B and adds one or more non-conflicting atomic facts, details, qualifiers, dates, locations, quantities, roles, or context.
- Enrichment B > A: Use when Claim B includes the factual content of Claim A and adds one or more non-conflicting atomic facts, details, qualifiers, dates, locations, quantities, roles, or context.
- Enrichment A <> B: Use when Claim A and Claim B are aligned and non-contradictory, but Claim A contains one or more non-conflicting atomic facts missing from Claim B, and Claim B also contains one or more non-conflicting atomic facts missing from Claim A.

Decision process:
1) Decide Alignment first.
2) If Alignment is not "Aligned", Enrichment must be an empty string "".
3) If Alignment is "Aligned", Enrichment must be one of: "None", "A > B", "B > A", "A <> B".
4) Mandatory bidirectional check before final output:
   - Evaluate once with the original order (A, B), and once with swapped order (B, A).
   - Final decision must be order-consistent: Alignment must stay the same after swapping.
   - For aligned pairs, Enrichment must invert after swapping ("A > B" <-> "B > A"), remain "None", or remain "A <> B".
   - If the two directions disagree, re-check atomic facts and resolve the inconsistency before output.

\end{tcblisting}
\caption{Prompt for Alignment (1)}
\label{fig:prompt_alignment_system_pt1}
\end{figure*}

\begin{figure*}[t]
\centering
\begin{tcblisting}{
    enhanced,
    width=\textwidth,
    colback=gray!4,
    colframe=gray!50,
    boxrule=0.4pt,
    arc=1.5pt,
    title=\textbf{System Prompt for Claim Alignment - w/o infobox (Continued)},
    colbacktitle=gray!15,
    coltitle=black,
    fonttitle=\small\bfseries,
    attach boxed title to top left={yshift=-2mm,xshift=2mm},
    boxed title style={boxrule=0pt, arc=1pt},
    left=4pt, right=4pt, top=6pt, bottom=4pt,
    listing only,
    breakable=false,
    listing options={
        breaklines=true,
        breakatwhitespace=false,
        basicstyle=\ttfamily\small
    }
}
Boundary guidance:
- Missing information is not contradiction. Use "Contradicted" only when both claims explicitly assert incompatible values for the same factual aspect.
- Sentence tense alone does not affect the label, unless the claims explicitly assert incompatible temporal states or dates/periods.
- For the same named entities, minor surface-form, wording, paraphrase, or transliteration differences do not change facts and should not be treated as factual mismatch: capitalization, translation variants, punctuation, spacing, aliases.
- Different name forms of the same entity do not count as enrichment, including translations, transliterations, aliases, full names, middle names, patronymics, original-language names, or parenthesized glosses.

Output format requirements:
- Output exactly one JSON object and nothing else.
- Use exactly these keys and values: {"Why": "...", "Alignment": "...", "Enrichment": "..."}
- "Why" must be a concise explanation string. It must first repeat Claim A and Claim B in order (A first, then B), then include the Decision process: identify the key factual aspect being compared, explain the Alignment decision, and explain the Enrichment decision.
- "Alignment" must be one of: "Aligned", "Contradicted", "Not Relevant".
- "Enrichment" must be one of: "None", "A > B", "B > A", "A <> B", or "" (only when not aligned).
- No markdown and no extra text outside the JSON object.
\end{tcblisting}
\caption{Prompt for Alignment (2)}
\label{fig:prompt_alignment_system_pt2}
\end{figure*}

\begin{figure*}[t]
\centering
\begin{tcblisting}{
    enhanced,
    width=\textwidth,
    colback=gray!4,
    colframe=gray!50,
    boxrule=0.4pt,
    arc=1.5pt,
    title=\textbf{User Prompt for Claim Alignment - w/o infobox (Continued)},
    colbacktitle=gray!15,
    coltitle=black,
    fonttitle=\small\bfseries,
    attach boxed title to top left={yshift=-2mm,xshift=2mm},
    boxed title style={
        boxrule=0pt,
        arc=1pt,
    },
    left=4pt,
    right=4pt,
    top=6pt,
    bottom=4pt,
    listing only,
    breakable=false,
    listing options={
        breaklines=true,
        breakatwhitespace=false,
        basicstyle=\ttfamily\small
    }
}
Classify the relation between Claim A and Claim B.

Direction reminder:
- A > B means Claim A contains Claim B's factual content plus extra non-conflicting detail.
- B > A means Claim B contains Claim A's factual content plus extra non-conflicting detail.
- A <> B means both claims are aligned and non-contradictory, but each claim contains extra non-conflicting detail that the other claim does not contain.

Entity reference (for cross-lingual identity matching):
- Wikidata canonical name: {name}
- English Wikipedia title: {en_title}
- non-English-language Wikipedia title: {non-English_title}
- Use this only for identity matching, not as evidence of enrichment.

Claim A is the first claim below.
<claim_a>
{a_claim}
</claim_a>

Claim B is the second claim below.
<claim_b>
{b_claim}
</claim_b>
\end{tcblisting}
\caption{Prompt for Alignment (3)}
\label{fig:prompt_alignment_user}
\end{figure*}




\begin{figure*}[t]
\centering
\begin{tcblisting}{
    enhanced,
    width=\textwidth,
    colback=gray!4,
    colframe=gray!50,
    boxrule=0.4pt,
    arc=1.5pt,
    title=\textbf{System Prompt for Cross-Lingual Generation},
    colbacktitle=gray!15,
    coltitle=black,
    fonttitle=\small\bfseries,
    attach boxed title to top left={yshift=-2mm,xshift=2mm},
    boxed title style={
        boxrule=0pt,
        arc=1pt,
    },
    left=4pt,
    right=4pt,
    top=6pt,
    bottom=4pt,
    listing only,
    breakable=false,
    listing options={
        breaklines=true,
        breakatwhitespace=false,
        basicstyle=\ttfamily\small
    }
}
You are an expert Wikipedia biography editor.
Your task is to improve an existing English biography using additional facts from a non-English biography excerpt.
Follow these rules strictly:
- If there is a factual conflict between the current English biography and the non-English-language excerpt, prefer the current English biography.
- Add only information that is explicitly supported by the non-English-language excerpt.
- Do not use external knowledge or model prior knowledge.
- Only make necessary additions/edits relative to the current English biography.
- If a fact already exists in English and is consistent, keep it and do not rewrite it unnecessarily.
- Preserve all factual information already present in the current English biography by default.
- Do not remove or invalidate English-only facts based on the non-English-language excerpt.
- Do not hallucinate, speculate, or invent citations.
- Use neutral encyclopedic English in Wikipedia biography style.
- Integrate new facts naturally at appropriate positions in the biography.
- Avoid duplicated facts or repetitive phrasing.
- Preserve the existing section structure by default.
- Only add a new section when it is necessary for coherent organization of newly added information.
- Section headers must use the format: **SECTION_TITLE** followed by a newline.
- Return the full revised English biography text only (no JSON, no Markdown code fence, no explanation).
\end{tcblisting}
\caption{System Prompt for Cross-Lingual Generation}
\label{fig:prompt_sys_en_X_gen}
\end{figure*}

\begin{figure*}[t]
\centering
\begin{tcblisting}{
    enhanced,
    width=\textwidth,
    colback=gray!4,
    colframe=gray!50,
    boxrule=0.4pt,
    arc=1.5pt,
    title=\textbf{User Prompt for Cross-Lingual Generation},
    colbacktitle=gray!15,
    coltitle=black,
    fonttitle=\small\bfseries,
    attach boxed title to top left={yshift=-2mm,xshift=2mm},
    boxed title style={
        boxrule=0pt,
        arc=1pt,
    },
    left=4pt,
    right=4pt,
    top=6pt,
    bottom=4pt,
    listing only,
    breakable=false,
    listing options={
        breaklines=true,
        breakatwhitespace=false,
        basicstyle=\ttfamily\small
    }
}
Task:
Revise the current English biography by incorporating factual information from the non-English-language excerpt that is missing from the current English biography.

Metadata:
- qid: {{qid}}
- original_language: {{original_language}}
- non-English_language: {{non-English_language}}
- english_wikipedia_name: {{english_wikipedia_name}}
- original_language_wikipedia_name: {{original_language_wikipedia_name}}
- wikidata_name: {{wikidata_name}}
- nationality: {{nationality}}
- occupations: {{occupations}}

Requirements:
- Treat the current English biography as the primary reference when conflicts occur.
- Integrate new facts from the non-English excerpt into the English biography when missing.
- Added or revised facts must be supported by the non-English excerpt.
- Do not introduce facts from outside the provided non-English excerpt.
- If a fact already exists in English and is consistent, do not rewrite it unnecessarily.
- Keep chronology and factual consistency.
- Place newly added facts in contextually appropriate locations for fluent biography writing.
- Avoid repeated statements about the same fact.
- Keep the original section layout whenever possible; add new sections only when necessary.
- Do not output notes, bullets, or explanations.
- Output only the full revised English biography.

Current English biography:
<<<CURRENT_ENGLISH_BIO_START>>>
{{current_english_bio}}
<<<CURRENT_ENGLISH_BIO_END>>>

non-English biography excerpt:
<<<non-English_BIO_EXCERPT_START>>>
{{non-English_bio_excerpt}}
<<<non-English_BIO_EXCERPT_END>>>
\end{tcblisting}
\caption{User Prompt for Cross-Lingual Generation}
\label{fig:prompt_user_en_X_gen}
\end{figure*}

\begin{figure*}[t]
\centering
\begin{tcblisting}{
    enhanced,
    width=\textwidth,
    colback=gray!4,
    colframe=gray!50,
    boxrule=0.4pt,
    arc=1.5pt,
    title=\textbf{Prompt for LLM-based Biography Translation},
    colbacktitle=gray!15,
    coltitle=black,
    fonttitle=\small\bfseries,
    attach boxed title to top left={yshift=-2mm,xshift=2mm},
    boxed title style={
        boxrule=0pt,
        arc=1pt,
    },
    left=4pt,
    right=4pt,
    top=6pt,
    bottom=4pt,
    listing only,
    breakable=false,
    listing options={
        breaklines=true,
        breakatwhitespace=false,
        basicstyle=\ttfamily\small
    }
}
system_prompt: |
  You are a professional translator for Wikipedia biographies.
  Translate the source text into natural, faithful English.
  Keep all facts, dates, names, numerals, and structure.
  Do not omit any information.
  Do not add explanations, notes, or comments.
  Output only the translated English text.

user_prompt_template: |
  Task: Translate the following biography chunk into English.

  Metadata:
  - original_language: {{original_language}}
  - non-English_language: {{non-English_language}}
  - english_wikipedia_name: {{english_wikipedia_name}}
  - original_language_wikipedia_name: {{original_language_wikipedia_name}}
  - wikidata_name: {{wikidata_name}}
  - nationality: {{nationality}}
  - occupations: {{occupations}}
  - qid: {{qid}}
  - chunk_id: {{chunk_id}}
  - chunk_index: {{chunk_index}}
  - total_chunks_for_qid: {{total_chunks_for_qid}}

  Requirements:
  - Preserve all facts and details faithfully.
  - Keep section markers like **SECTION_NAME** if present.
  - Keep list structure and paragraph boundaries when possible.
  - Do not summarize and do not omit any content.
  - Output only translated English text for this chunk.

  Source text to translate:
  <<<SOURCE_CHUNK_START>>>
  {{source_chunk}}
  <<<SOURCE_CHUNK_END>>>

\end{tcblisting}
\caption{Prompt for LLM-based Biography Translation}
\label{fig:prompt_bio_mt}
\end{figure*}

\end{document}